# Bayesian inference for data-efficient, explainable, and safe robotic motion planning: A review

Chengmin Zhou, Chao Wang, Haseeb Hassan, Himat Shah, Bingding Huang, Pasi Fränti*

*First correspondence: Pasi Fränti. Affiliation: Machine Learning Group, School of Computing, University of Eastern Finland, FI-80100 Joensuu, Finland (email: franti@cs.uef.fi).
Second correspondence: Bingding Huang. Affiliation: College of Big Data and Internet, Shenzhen Technology University, 518118 Shenzhen, China (email: huangbingding@sztu.edu.cn).
First author: Chengmin Zhou. Chengmin Zhou is with Machine Learning Group, School of Computing, University of Eastern Finland, FI-80100 Joensuu, Finland, and College of Big Data and Internet, Shenzhen Technology University, 518118 Shenzhen, China simultaneously (email: zhou@cs.uef.fi).
Second author: Chao Wang. Affiliation: College of Blockchain industry, Chengdu University of Information Technology, 610225 Chengdu, China (email: wangchao@cuit.edu.cn).
Third author: Haseeb Hassan. Affiliation: College of Health Science and Environmental Engineering, Shenzhen Technology University, Shenzhen,518118, China (email: haseeb@sztu.edu.cn).
Fourth author: Himat Shah. Affiliation: Machine Learning Group, School of Computing, University of Eastern Finland, FI-80100 Joensuu, Finland (email: himat@cs.uef.fi).

Abstract: Bayesian inference has many advantages in robotic motion planning over four perspectives: The uncertainty quantification of the policy, safety (risk-aware) and optimum guarantees of robot's motions, data-efficiency in reinforcement learning (RL) training, and reducing the sim2real gap when the robot is applied to real-world tasks. However, the application of Bayesian inference in robotic motion planning is lagging behind the comprehensive theory of Bayesian inference. Further, there are no comprehensive reviews to summarize the progress of Bayesian inference to give researchers a systematic understanding in robotic motion planning. This paper first provides the probabilistic theories of Bayesian inference which are the preliminary of Bayesian inference for complex cases. Second, the Bayesian estimation is given to estimate the posterior of policies or unknown functions which are used to compute the policy. Third, the classical model-based Bayesian RL and model-free Bayesian RL algorithms for robotic motion planning are summarized, while these algorithms in complex cases are also analyzed. Fourth, the analysis of Bayesian inference in inverse RL is given to infer the reward functions in a data-efficient manner. Fifth, we systematically present the hybridization of Bayesian inference and RL which is a promising direction to improve the convergence of RL for better motion planning. Sixth, given the Bayesian inference, we present the interpretable and safe robotic motion plannings which are the hot research topic recently. Finally, all algorithms reviewed in this paper are summarized analytically as the knowledge graphs, and the future of Bayesian inference for robotic motion planning is also discussed, to pave the way for data-efficient, explainable, and safe robotic motion planning strategies for practical applications.

Key words: Bayesian inference, Bayesian reinforcement learning, Interpretability, Safety, Bayesian inverse reinforcement learning, Robotics

## 1. Introduction

Bayesian inference is widely used in various fields like the motion planning of autonomous car [1][2] and the structure design of protein [3][4] to provide safe, optimal, data-efficient, and interpretable predictions. Bayesian inference not only provides predictions as that of classical RL like deep Q learning (DQN) [5], but also quantifies the uncertainty of predictions which are critical in safe-related applications like the autonomous driving and surgical predictions [6]. Moreover, Bayesian inference provides predictions that are robust to the stochastic noise of the real world, therefore further secure the safety of predictions in the sim2real transplantation [7]. Classical RL and inverse RL can compute the optimal predictions given unlimited data. However, the data available in practice is limited, and this blocks the further applications of data-driven RL and inverse RL in real-world tasks. In contrast, Bayesian inference is both optimal and data-efficient by incorporating the domain knowledge to the parameters like the state transitions or policies to make the algorithms converge with less data. State-of-the-art model-free RL like the soft actor critic [8] is attractive to compute the optimal predictions by training neural networks which is however like a "black box". The predictions made by the trained but unknown neural networks are unexplainable or uninterpretable to humans because the transitions or dynamics of these RL are unknown.

In contrast, Bayesian inference makes the predictions by training and maintaining known transition matrixes. This makes the predictions explainable. Further, additional generative models (post-training explanations) like the semantic masks [9] and importance score [10] can be inferred by Bayesian inference. These generative models explicitly reveal how good the predictions are, therefore further improve the interpretability of predictions or actions made by policies.

However, using Bayesian inference alone to solve the complex real-world problems faces two critical problems: 1) The solutions derived from Bayesian inference is expensive to solve, due to the *intractable integral* derived. 2) Bayesian inference is competent to solve low-dimensional problems like the point estimations with Gaussian linear model and low-dimensional dataset, but slowly and expensively performed in *high-dimensional nonlinear non-Gaussian problems* due to the unlimited dimensions of the probability density function (PDF) in the transition function. Above two problems can be mitigated from three perspectives: 1) *Surrogate models* like linear Gaussian model [11] to reduce the complexity of transition functions. 2) *Transition (posterior) approximation* by sampling methods like the Monte-Carlo sampling [12][13] and variational inference [14][15]. 3) The *combinations of Bayesian inference and RL*, resulting in the *classical model-based Bayesian RL* [16][17][18] and *model-free Bayesian RL* [16]. The combinations of Bayesian inference and RL make the best of efficient RL architecture to fast find the converged transition functions. Further, the Bayesian inference can be *hybridized* with RL from various ways like incorporating Bayesian surprise [19] and Bayesian curiosity [20] to the RL objectives to accelerate the policy exploration of RL policies.

To conclude, given the support of surrogate models, approximation methods and the combinations of Bayesian inference with RL, the solutions based on Bayesian inference are expected to be safe, optimal, data-efficient, and interpretable. This makes the Bayesian inference competitive to solve the real-world motion planning problems especially in safe-critical applications. However, there are less works that systematically and comprehensively summarize the contributions of Bayesian inference for robotic motion planning. This review is expected to fill in this gap to provide systematic and comprehensive overview of Bayesian inference for robotic motion planning, therefore paving the way for better motion planning algorithms in academia and industry.

This review is organized as follows: 1) **Section 2** gives the probability theories which ground the understanding following motion planning algorithms based on Bayesian inference. 2) **Section 3** gives the Bayesian estimation. 3) **Sections 4-6** gives the classical model-based Bayesian RL and model-free Bayesian RL, as well as the Bayesian RL in complex cases including the unknown reward case, partial observation case, multi-agent case, and multi-task case. 4) **Section 7** gives the Bayesian inference in inverse RL. 5) **Section 8** gives the hybridization of Bayesian inference and RL for better convergence. 6) **Sections 9-10** give the Bayesian inference for interpretable and safe motion planning respectively. 7) **Section 11** gives the conclusion and discussions of Bayesian inference for future motion plannings. The framework of Bayesian inference for robotic motion planning is simplified as Figure 1.

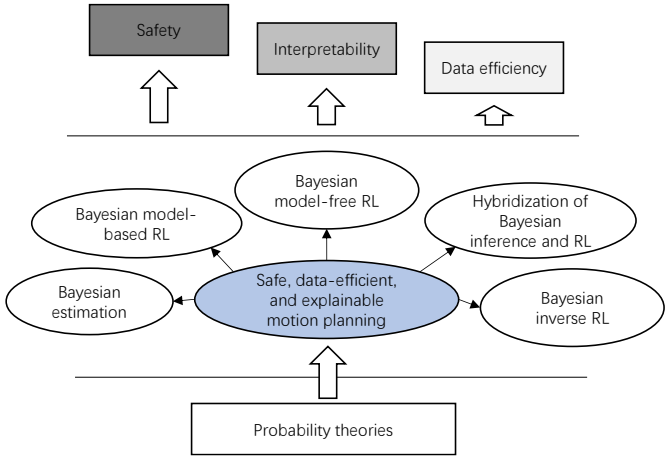

Figure 1. The framework of Bayesian inference for robotic motion planning. This review focuses on applying Bayesian inference to intelligent robotic motion planning from five perspectives: Bayesian estimation, model-based Bayesian RL, model-free Bayesian RL, hybridization of Bayesian inference and RL, and Bayesian inverse RL. The direct consequence is the improvement in the data efficiency. The Bayesian inference also provides the uncertainty quantification in the posterior estimation which makes the motion planning a step closer to interpretability and safety. The interpretability and safety can be further improved by adding extra processes which will be elaborated in Sections 9-10.

## 2. Probability theories

This section elaborates the prime Bayes theories to ground the understanding of following advance motion planning algorithms based on Bayesian inference.

***Probability distribution function.*** The non-negative probability distribution function $p(x)$ satisfies $\int_a^b p(x)dx = 1, x \in [a, b]$ where $x$ is a *random variable* over the interval $[a, b]$.

***Two common probability moments.*** The moments related to this review refer to the first and second moments, known as the *mean* and *covariance*. The mean is defined by the expectation of random variable $x$ via

$$\mu = \mathbb{E}[x] = \int x p(x) dx \quad (1)$$

In matrix case, the mean is defined by the expectation of the expectation of a general matrix function via $\mu = \mathbb{E}[F(x)] = \int F(x)p(x)dx$. The covariance is defined by

$$\Sigma = \mathbb{E}[(x-\mu)(x-\mu)^\intercal] \quad (2)$$

Normally, it is expensive to compute the mean and covariance via computing the integrals. Drawing samples from the probability density function to approximate the true mean and covariance is widely used. The sample mean is defined by

$$\mu = \frac{1}{N}\sum_{i=1}^N x_i \quad (3)$$

where $N$ denotes the number of samples. The sample covariance is defined by

$$\Sigma = \frac{1}{N-1}\sum_{i=1}^N (x_i - \mu)(x_i - \mu)^\intercal \quad (4)$$

where the denominator is $N-1$, instead of the $N$, to acquire unbiased estimate of the true covariance given the *Bessel's correction* [21].

***Statistically independent, uncorrelated.*** If two random variables $x$ and $y$ are *statistically independent*, the joint density function $p(x, y)$ factorized by $p(x,y) = p(x|y)p(y) = p(y|x)p(x) = p(y, x)$ is simplified as

$$p(x, y) = p(x)p(y), p(x|y) = p(x) \quad (5)$$

If two random variables $x$ and $y$ are *uncorrelated*, their expectation is factorized as

$$\mathbb{E}[xy^\intercal] = \mathbb{E}[x]\mathbb{E}[y]^\intercal \quad (6)$$

***Normalized product.*** If $p_1(x)$ and $p_2(x)$ are two probability density functions of $x$, the *normalized product (fused probability density function)* $p(x)$ is written as

$$p(x) = \eta p_1(x)p_2(x), \eta = (\int p_1(x)p_2(x)dx)^{-1} \quad (7)$$

where $\eta$ is the *normalization constant* that makes $\int_a^b p(x)dx = 1, x \in [a, b]$. In the Bayesian context, the computing of the normalized product $p(x|y_1, y_2)$ is written as

$$p(x|y_1, y_2) = \eta p(x|y_1)p(x|y_2), \eta = \frac{p(y_1)p(y_2)}{p(y_1, y_2)p(x)} \quad (8)$$

***Shannon information, mutual information, Fisher information, and Cramer-Rao lower bound to quantify the uncertainty.*** The uncertainty can be measured by the moments, while the uncertainty can be also measured by the *Shannon*

*information* and *mutual information*. The Shannon information $H(x)$ represents *how certain we are after we estimate a probability density function for random variables*, and it is defined by

$$H(x) = -\mathbb{E}[\ln p(x)] = -\int p(x)\ln p(x)\,dx \quad (9)$$

The mutual information $I(x, y)$ represents *how much knowing one of the variables reduces uncertainty about the other*, and it is defined by

$$I(x, y) = \mathbb{E}\left[\ln \frac{p(x,y)}{p(x)p(y)}\right] = \iint p(x, y) \ln \frac{p(x,y)}{p(x)p(y)} dxdy \quad (10)$$

The relationship between the Shannon information and mutual information is written as

$$I(x, y) = H(x) + H(y) - H(x, y) \quad (11)$$

The *fisher information* is defined by

$$I(x|\theta) = \mathbb{E}\left[\left(\frac{\partial \ln p(x|\theta)}{\partial \theta}\right)^\top \left(\frac{\partial \ln p(x|\theta)}{\partial \theta}\right)\right] \quad (12)$$

where the deterministic parameter $\theta$ decides the random variable $x$. Normally, we can just sample limited variables as the measurement $x_{meas}$ to acquire the *unbiased estimate* of the deterministic parameter $\hat{\theta}$. The fisher information is to bound the covariance of the unbiased estimate via

$$cov(\hat{\theta}|x_{meas}) = \mathbb{E}[(\hat{\theta} - \theta)(\hat{\theta} - \theta)^\top] \geq I^{-1}(x|\theta) \quad (13)$$

This is known as the *Cramer-Rao lower bound* [22] which *sets a limit to how certain we are in the estimation of the unbiased estimate $\hat{\theta}$ under the measurements $x_{meas}$*.

**Bayes' theorem (rule).** Given the joint density function $p(x, y)$ which is factorized by $p(x, y) = p(x|y)p(y) = p(y|x)p(x) = p(y, x)$, the Bayes' theorem (rule) [23] is written by

$$p(x|y) = \frac{p(y|x)p(x)}{p(y)} \quad (14)$$

where in motion planning case, $y$ denotes the *observation (measurement)*. The $x$ denotes the *state or parameter*. The $p(x)$ denotes the *prior distribution function* which capsulates all priori information. The denominator $p(y)$ denotes the *known evidence* for the normalization. The *likelihood distribution function* $p(y|x)$ is also written as $\mathcal{L}(x; y)$. The $p(x|y)$ denotes the *posterior distribution function* which we are expected to acquire.

Obviously, to compute the posterior $p(x|y)$, it is necessary to compute the likelihood $p(y|x)$, the prior $p(x)$, and known evidence $p(y)$. Normally, the prior $p(x)$ is known. Given the joint density function $p(x, y)$, the known evidence $p(y)$ is computed by the *marginalization* via

$$p(y) = p(y)\int p(x|y)\,dx = \int p(x|y)\,p(y)dx = \int p(y|x)\,p(x)dx \quad (15)$$

where $\int p(x|y)\,dx = 1$. Hence, the focus is on the computing of the likelihood $p(y|x)$ to acquire the posterior $p(x|y)$.

**Gaussian distribution, Dirichlet distribution, and Beta distribution**. It is usual to assume that the prior and posteriori follow some specific distributions to facilitate the computing of the posteriori moments. Common distributions include the *Gaussian* distribution, *Dirichlet* distribution and *Beta* distribution [24]. The probability density function of Gaussian distribution is defined by

$$p(x|\mu, \sigma^2) = \frac{1}{\sigma\sqrt{2\pi}} \exp\left(-\frac{(x-\mu)^2}{2\sigma^2}\right) = \mathcal{N}(\mu, \sigma^2) \quad (16)$$

where the parameters $\mu$ and $\sigma^2$ (also written as $\Sigma$) denote the mean and variance of $p(x)$. In the case of a pair of random variable $(x, y)$, it's *joint distribution* $p(x, y)$ is written as

$$p(x, y) = \mathcal{N}\left(\begin{bmatrix}\mu_x \\ \mu_y\end{bmatrix}, \begin{bmatrix}\Sigma_{xx} & \Sigma_{xy} \\ \Sigma_{yx} & \Sigma_{yy}\end{bmatrix}\right) \quad (17)$$

where $\mathbb{E}[xx^\top] = \Sigma = \Sigma_{xx}$ and $\Sigma_{yx} = \Sigma_{xy}^\top$. Given the joint distribution and $p(x, y) = p(x|y)p(y)$, the posterior distribution or the likelihood of $x$ given the value of $y$ (observations) is written as

$$p(x|y) = \mathcal{N}(\mu_x + \Sigma_{xy}\Sigma_{yy}^{-1}(y - \mu_y), \Sigma_{xx} - \Sigma_{xy}\Sigma_{yy}^{-1}\Sigma_{yx}) \quad (18)$$

The Dirichlet distribution describes the probability of $K$ outcomes, instead of describing two outcomes in the Beta distribution. There are $K$ positive parameters $\boldsymbol{\alpha} = \alpha_1, \alpha_2, \ldots, \alpha_K$, while there are two positive parameters ($\alpha$ and $\beta$) in the Beta distribution. The probability density function of Dirichlet distribution is defined by

$$p(\boldsymbol{\theta}; \boldsymbol{\alpha}) = \frac{1}{B(\boldsymbol{\alpha})}\prod_{i=1}^{K}\theta_i^{\alpha_i - 1}, \sum_{i=1}^{K}\theta_i = 1 \quad (19)$$

The $B(\boldsymbol{\alpha})$ is the *multivariate Beta function* defined by

$$B(\boldsymbol{\alpha}) = \frac{\prod_{i=1}^{K}\Gamma(\alpha_i)}{\Gamma(\sum_{i=1}^{K}\alpha_i)} \quad (20)$$

The moments of Dirichlet distribution are

$$\text{mean of }\theta_i: \langle\theta_i\rangle = \frac{\alpha_i}{\sum_{i=k}^{K}\alpha_k} \quad (21)$$

$$\text{variance of }\theta_i: \frac{\langle\theta_i\rangle(1 - \langle\theta_i\rangle)}{1 + \sum_{k=1}^{K}\alpha_k} \quad (22)$$

$$\text{covariance of }\theta_i \text{ and }\theta_j: -\frac{\langle\theta_i\rangle\langle\theta_j\rangle}{1 + \sum_{k=1}^{K}\alpha_k}, \text{where } i \neq j \quad (23)$$

The probability density function of Beta distribution is defined by

$$p(\theta; \alpha, \beta) = \frac{\theta^{\alpha-1}(1-\theta)^{\beta-1}}{B(\alpha, \beta)} \quad (24)$$

The *Beta function* $B(\alpha, \beta)$ is defined by

$$B(\alpha, \beta) = \frac{\Gamma(\alpha)\Gamma(\beta)}{\Gamma(\alpha+\beta)} = \int_0^1 t^{\alpha-1}(1-t)^{\beta-1}dt, \Gamma(z) = \int_0^\infty t^{z-1}e^{-t}dt \quad (25)$$

where $\Gamma$ is the *gamma function*. The moments of Beta distribution are

$$\text{mean}: \frac{\alpha}{\alpha+\beta} \quad (26)$$

$$\text{variance}: \frac{\alpha\beta}{(\alpha+\beta)^2(\alpha+\beta+1)} \quad (27)$$

## 3. Bayesian estimation

This section addresses the Bayesian inference for posterior estimation (simplified as Bayesian estimation). First, we address the Bayesian learning which grounds the Bayesian estimation. However, the posterior derived from Bayesian estimation is intractable or computationally expensive. This problem is expected to be solved by the posterior approximations which include the linearization, sampling, and variational inference. Finally, we give the Bayesian optimization to find the (local) optimal posterior with as little data as possible.

**3.1 Bayesian learning with Gaussian conjugate distribution**

This section describes the Bayesian learning with Gaussian conjugate distribution [16]. First, the basic concepts of Gaussian process (GP) and Gaussian process regression (GPR) are given. Then, kernel-based and non-parametric Bayesian learning Bayesian learning is given. Finally, Bayesian learning is extended to the neural network case.

**3.1.1 Gaussian process (GP) and Gaussian process regression (GPR)**

The variable set $X \in \mathbb{R}^N$ is *Gaussian random variables* if $X \sim \mathcal{N}(\mu, \Sigma)$. GP is an *indexed set of Gaussian random variables* which can be represented by function $F(x), x \in \chi$ where $\chi = \{x_1, x_2..x_T\}$ is the index set. Hence, $X = (F(x_1), F(x_2)..F(x_T))$ which is a vector-valued Gaussian random variables. GP Function $F$ is unknown, but it can be fully specified by its mean $\bar{f}(x) = \mathbb{E}[F(x)]$ and its covariance $k(x, x') = \mathbf{Cov}[F(x), F(x')]$. Hence, GP can be simply written as $F(\cdot) \sim \mathcal{N}\left(\bar{f}(\cdot), k(\cdot, \cdot)\right)$ where the covariance $k(\cdot, \cdot)$ is also called *kernel function* which encodes the correlations (prior information) between the members of GP.

GPR [25] is the regression approach to infer GP Function $F$. GP Function $F$ and associated indexed random variable $F(x)$ are unknown, and they cannot be inferred *directly*. However, it is possible to collect associated indexed observation $Y(x)$. $F(x)$ and $Y(x)$ have some sorts of relationships which might be represented by simple linear statistical model

$$Y(x) = HF(x) + N(x) \quad (28)$$

where $H$ is the state transformation model for linear transformation, $N$ the Gaussian noise. Hence, GP Function $F$ can be inferred *indirectly* from the observations $Y(x)$ and defined linear statistical model $H$ by Bayesian inference to find the *updated mean and covariance* (posterior) which specify updated GP Function $F$.

**3.1.2 Kernel-based and non-parametric Bayesian learning**

The Bayesian learning or Bayesian inference in this case can be simplified as follows:

1) Preparations: First, the *prior* (initial probability density function of GP Function $F$) should be selected. Second, the (linear) *model* that describe the relationship between $F(x)$ and $Y(x)$ should be selected. Third, the *samples* of the observation $Y(x)$ should be collected.

2) Inference: The Bayes rule is applied to infer posterior conditioned on the samples.

Specifically, Samples are collected and represented as $D_T = \{(x_t, y_t)\}_{t=1}^T$. The model is selected by

$$Y_T = HF_T + N_T \quad (29)$$

$Y_T = (y_1, y_2..y_T)^\top$, $F_T = (F(x_1), F(x_2)..F(x_T))^\top$, and $N_T \sim (0, \Sigma)$ where $\Sigma$ is the covariance of measurement noise. Hence, the prior is selected via $F_T \sim \mathcal{N}(\bar{f}, K)$ where $\bar{f} = (\bar{f}(x_1), \bar{f}(x_2)..\bar{f}(x_T))^\top$ and $K_{i,j} = k(x_i, x_j)$. It is easy to see $Y_T \sim \mathcal{N}(H\bar{f}, HKH^\top + \Sigma)$ because $F_T$ and $N_T$ are defined as the independent Gaussian. Selected linear model can be transformed into

$$\begin{pmatrix} F(x) \\ F_T \\ Y_T \end{pmatrix} = \begin{bmatrix} 1 & 0 & 0 \\ 0 & I & 0 \\ 0 & H & I \end{bmatrix} \begin{pmatrix} F(x) \\ F_T \\ N_T \end{pmatrix} \quad (30)$$

where $I$ is the identity matrix.

*Joint distribution* is therefore written as

$$\begin{pmatrix} F(x) \\ F_T \\ Y_T \end{pmatrix} = \mathcal{N}\left\{ \begin{pmatrix} \bar{f}(x) \\ \bar{f} \\ H\bar{f} \end{pmatrix}, \begin{bmatrix} k(x,x) & k(x)^\top & k(x)^\top H^\top \\ k(x) & K & KH^\top \\ Hk(x) & HK & HKH^\top + \Sigma \end{bmatrix} \right\}$$
(31)

where $k(x) = (k(x_1, x), k(x_2, x)..k(x_T, x))^\top$. According to the Gauss-Markov theorem, the posterior distribution $p(F(x)|Y_T)$ or $p(F(x)|D_T)$ is the Gaussian distribution. By factoring joint distribution $p(F(x), Y_T) = p(Y_T)p(F(x)|Y_T)$, the posterior mean and covariance of $F(x)$ conditioned on $D_T$ are

$$\mathbb{E}[F(x)|D_T] = \bar{f}(x) + k(x)^\top \alpha \quad (32)$$

$$\mathbf{Cov}[F(x), F(x')|D_T] = k(x, x') - k(x)^\top C k(x') \quad (33)$$

where $\alpha$ and $C$ denote *sufficient statistics of the posterior moments* and

$$\alpha = H^\top (HKH^\top + \Sigma)^\top (y_T - H\bar{f}) \quad (34)$$

$$C = H^\top (HKH^\top + \Sigma)^\top H \quad (35)$$

where $y_T = (y_1, y_2..y_T)^\top$ is one realization of $Y_T$.

**3.1.3 Parametric Bayesian learning**

In parametric case [16], the GP function is assumed to consist of the feature vector and weight vector, written as $F(\cdot) = \phi(\cdot)^\top W$ where $\phi(\cdot) = (\varphi_1, \varphi_2, ..., \varphi_n)^\top$ is the feature vector, and $W = (W_1, W_2, ..., W_n)^\top$ is the weight vector. Hence, the linear model that describe the relationship between $F(x)$ and $Y(x)$ is written as

$$Y(x) = H \phi(\cdot)^\top W + N(x) \quad (36)$$

The randomness of GP function is from the weight vector $W$. Assuming the prior $W \sim \mathcal{N}(\bar{w}, S_w)$ where $\bar{w}$ and $S_w$ denote the mean and variance of the prior. The posterior moments of $W$ are computed as

$$\mathbb{E}[W|\mathcal{D}_T] = \bar{w} + S_w \Phi H^\top (H\Phi^\top S_w \Phi H^\top + \Sigma)^{-1}(y_T - H\Phi^\top \bar{w})$$
(37)

$$Cov[W|\mathcal{D}_T] = S_w - S_w \Phi H^\top (H\Phi^\top S_w \Phi H^\top + \Sigma)^{-1} H\Phi^\top S_w \quad (38)$$

where $\Phi = [\phi(x_1), \phi(x_2), ..., \phi(x_T)]$ is the $n \times T$ feature matrix. Due to $F(\cdot) = \phi(\cdot)^\top W$, the posterior moments of $F$ can be computed accordingly by

$$\mathbb{E}[F(x)|\mathcal{D}_T] = \phi(x)^\top \bar{w} + \phi(x)^\top S_w \Phi H^\top (H\Phi^\top S_w \Phi H^\top + \Sigma)^{-1}(y_T - H\Phi^\top \bar{w}) \quad (39)$$

$$Cov[F(x), F(x')|\mathcal{D}_T] = \boldsymbol{\phi}(x)^\mathsf{T}\boldsymbol{S}_w - \boldsymbol{\phi}(x)^\mathsf{T}\boldsymbol{S}_w\boldsymbol{\Phi}\mathbf{H}^\mathsf{T}(\mathbf{H}\boldsymbol{\Phi}^\mathsf{T}\boldsymbol{S}_w\boldsymbol{\Phi}\mathbf{H}^\mathsf{T} + \boldsymbol{\Sigma})^{-1}\mathbf{H}\boldsymbol{\Phi}^\mathsf{T}\boldsymbol{S}_w\boldsymbol{\phi}(x') \quad (40)$$

To conclude, Bayesian learning with Gaussian conjugate distribution is kernel-based and widely used in various applications. The fundamental problems of this method are in the inefficiency of computation and data when computing the posterior. Posterior approximation is the classical way to address this problem. For instance, [26] uses a GP-based surrogate model parameterized by $\xi$ to approximate the standard Bayesian GP posterior by

$$\min_{\xi} D_{KL}(p(y|x_{t+1})||q_\xi(y|x_{t+1})) \quad (41)$$

to increase computational and sample efficiency. This is achieved via a neural network model with a Bayesian last layer.

## 3.2 Bayesian estimations to find the maximum posterior moments

The *objective* of this motion planning problem is to infer the maximum posterior moments of state $x_k$ given the input $v_k$ and observation $y_k$ in the motion and observation models. In this section, Bayesian inference are given first to compute the maximum posterior in linear-Gaussian case. Then, the Bayes filter and Bayesian inference is analyzed for maximizing the posterior in non-linear non-Gaussian case [22].

### 3.2.1 Bayesian inference for linear-Gaussian estimations

In the linear-Gaussian posterior estimations, we assume that the robot follows the following *motion* and *observation models* respectively

$$\begin{cases} x_k = A_{k-1}x_{k-1} + v_k + w_k, k = 1..K \\ y_k = C_k x_k + n_k, k = 0..K \end{cases} \quad (42)$$

where $x_k$ denotes the system states. The $A_{k-1}$ denotes the state transition matrix. The $v_k$ denotes the input. The $w_k$ denotes the process noise. The $y_k$ denotes the observation or measurement. The $C_k$ denotes the observation matrix. The $n_k$ denotes the measurement noise. We also assume that the initial state $x_0 \sim \mathcal{N}(\check{x}_0, \check{P}_0)$ where $\check{x}_0$ and $\check{P}_0$ denotes the prior's moment estimations. The process noise $w_k \sim \mathcal{N}(0, Q_k)$. The measurement noise $n_k \sim \mathcal{N}(0, R_k)$.

There are two situations about the dataset. First, we can access a batch of data with or without the time-sequential dependency among the data (*batch discrete-time case*). Second, we can recursively access the time-sequential data (*recursive discrete-time case*).

***Bayesian inference to compute posterior (batch discrete-time case).*** From the Bayesian inference perspective [22], the prior is written as

$$p(x|v) = \mathcal{N}(\check{x}, \check{P}) = \mathcal{N}(Av, AQA^\mathsf{T}) \quad (43)$$

The estimated prior mean is $\check{x} = \mathbb{E}[x] = \mathbb{E}[A(v+w)] = Av$ where $A$ is lower-triangular, and $x = A(v+w)$ is the *lifted matrix form* of motion model that considers the batch data in the entire trajectory. The lifted covariance is $\check{P} = \mathbb{E}[(x - \mathbb{E}[x])(x - \mathbb{E}[x])^\mathsf{T}] = AQA^\mathsf{T}$ where $Q = \mathbb{E}[ww^T]$. The joint density of lifted state and observation is written as

$$p(x,y|v) = \mathcal{N}\left(\begin{bmatrix} \check{x} \\ C\check{x} \end{bmatrix}, \begin{bmatrix} \check{P} & \check{P}C^\mathsf{T} \\ C\check{P} & C\check{P}C^\mathsf{T} + R \end{bmatrix}\right) \quad (44)$$

where $R = \mathbb{E}[nn^\mathsf{T}]$. Joint density is factored by $p(x,y|v) = p(x|y,v)p(y|v)$ where $p(y|v)$ is the known Gaussian distribution. Hence, the full Bayesian posterior $p(x|y,v)$ is computed via the factoring of joint Gaussian distribution [22]

$$p(x|y,v) = \mathcal{N}\left(\left(\check{P}^{-1} + C^\mathsf{T}R^{-1}C\right)^{-1}\left(\check{P}^{-1}\check{x} + C^\mathsf{T}R^{-1}y\right), \left(\check{P}^{-1} + C^\mathsf{T}R^{-1}C\right)^{-1}\right) \quad (45)$$

The above process repeats until the convergence of the posterior moments.

***Kalman filter via Bayesian inference to compute posterior (recursive discrete-time case).*** The batch solution discussed above cannot be used online, due to the fact that the batch solution uses the future data to infer the past state. However, future data is inaccessible when inferring the states online. The Kalman filter [27] can solve the problem of online state inference in a recursive way, and it can be derived via the Bayesian inference [22].

The Gaussian prior estimation at $k-1$ is written as

$$p(x_{k-1}|\check{x}_0, v_{1:k-1}, y_{0:k-1}) = \mathcal{N}(\hat{x}_{k-1}, \hat{P}_{k-1}) \quad (46)$$

where $\check{x}$ denotes the prior estimation and $\hat{x}$ denotes the posterior estimation. According to the linearity of the motion model, the prior estimation at $k$ (*prediction step*) is written as

$$p(x_k|\check{x}_0, v_{1:k}, y_{0:k-1}) = \mathcal{N}(\check{x}_k, \check{P}_k) \quad (47)$$

where the prior moments are acquired by

$$\begin{cases} \check{x}_k = A_{k-1}\hat{x}_{k-1} + v_k \\ \check{P}_k = A_{k-1}\hat{P}_{k-1}A_{k-1}^\mathsf{T} + Q_k \end{cases} \quad (48)$$

The joint density of state and observation at time $k$ is written as

$$p(x_k, y_k|\check{x}_0, v_{1:k}, y_{0:k-1}) = \mathcal{N}\left(\begin{bmatrix} \mu_x \\ \mu_y \end{bmatrix}, \begin{bmatrix} \Sigma_{xx} & \Sigma_{xy} \\ \Sigma_{yx} & \Sigma_{yy} \end{bmatrix}\right) = \mathcal{N}\left(\begin{bmatrix} \check{x}_k \\ C_k\check{x}_k \end{bmatrix}, \begin{bmatrix} \check{P}_k & \check{P}_k C_k^\mathsf{T} \\ C_k\check{P}_k & C_k\check{P}_k C_k^\mathsf{T} + R_k \end{bmatrix}\right) \quad (49)$$

By factoring the joint density, the posterior $p(x_k|\check{x}_0, v_{1:k}, y_{0:k})$ at time $k$ (*correction step*) is written as

$$p(x_k|\check{x}_0, v_{1:k}, y_{0:k}) = \mathcal{N}\left(\mu_x + \Sigma_{xy}\Sigma_{yy}^{-1}(y_k - \mu_y), \Sigma_{xx} - \Sigma_{xy}\Sigma_{yy}^{-1}\Sigma_{yx}\right) \quad (50)$$

where posterior mean $\hat{x}_k = \mu_x + \Sigma_{xy}\Sigma_{yy}^{-1}(y_k - \mu_y)$, and posterior covariance $\hat{P}_k = \Sigma_{xx} - \Sigma_{xy}\Sigma_{yy}^{-1}\Sigma_{yx}$. After the prediction-correction steps, we acquire the predictor, Kalman gain, and corrector written as

$$\text{Predictor:} \begin{cases} \check{x}_k = A_{k-1}\hat{x}_{k-1} + v_k \\ \check{P}_k = A_{k-1}\hat{P}_{k-1}A_{k-1}^\mathsf{T} + Q_k \end{cases} \quad (51)$$

$$\text{Kalman gain: } K_k = \check{P}_k C_k^\mathsf{T}(C_k\check{P}_k C_k^\mathsf{T} + R_k)^{-1} \quad (52)$$

$$\text{Corrector:} \begin{cases} \hat{x}_k = \check{x}_k + K_k(y_k - C_k\check{x}_k) \\ \hat{P}_k = (1 - K_k C_k)\check{P}_k \end{cases} \quad (53)$$

where the *Kalman gain* $K_k$ weights the innovation's contribution to the moment estimations, and $y_k - C_k\check{x}_k$ is the *innovation* which is the difference between the actual and expected observations.

### 3.2.2 Bayes filter and Bayes inference for non-linear non-Gaussian estimations

***Bayes filter to compute the posterior (recursive discrete-time case).*** In the case where the motion and observation models are non-linear, and the probability density function is non-Gaussian, the motion and observation models are rewritten as

$$\begin{cases} x_k = f(x_{k-1}, v_k, w_k), k = 1..K \\ y_k = g(x_k, n_k), k = 0..K \end{cases} \quad (54)$$

To recursively infer the states, we assume the system have the *Markov property*: A stochastic process has the Markov property, if the conditional probability density functions (PDFs) of future states depend only upon current state, and not on any other past states.

Bayes filter [28] seeks computing $p(x_k|\check{x}_0, v_{1:k}, y_{0:k})$ which is also known as the *belief* for $x_k$. The posterior belief is factored by Bayes rule and Markov property via

$$p(x_k|\check{x}_0, v_{1:k}, y_{0:k}) = \eta p(y_k|x_k) p(x_k|\check{x}_0, v_{1:k}, y_{0:k-1}) = \eta p(y_k|x_k) \int p(x_k|x_{k-1}, v_k) p(x_{k-1}|\check{x}_0, v_{1:k-1}, y_{0:k-1}) dx_{k-1} \quad (55)$$

where

$$\begin{cases} p(x_k|\check{x}_0, v_{1:k}, y_{0:k-1}) = \int p(x_k, x_{k-1}|\check{x}_0, v_{1:k}, y_{0:k-1}) dx_{k-1} \\ = \int p(x_k|x_{k-1}, \check{x}_0, v_{1:k}, y_{0:k-1}) p(x_{k-1}|\check{x}_0, v_{1:k}, y_{0:k-1}) dx_{k-1} \\ p(x_k|x_{k-1}, \check{x}_0, v_{1:k}, y_{0:k-1}) = p(x_k|x_{k-1}, v_k) \\ p(x_{k-1}|\check{x}_0, v_{1:k}, y_{0:k-1}) = p(x_{k-1}|\check{x}_0, v_{1:k-1}, y_{0:k-1}) \end{cases} \quad (56)$$

The $\eta$ preserves the axiom of total probability. The $p(y_k|x_k)$ is the observation correction based on observation model $g(\cdot)$. The $p(x_k|x_{k-1}, v_k)$ is the motion prediction based on motion model $f(\cdot)$. The $p(x_{k-1}|\check{x}_0, v_{1:k-1}, y_{0:k-1})$ is the prior belief.

In the prediction step, the prior belief $p(x_{k-1}|\check{x}_0, v_{1:k-1}, y_{0:k-1})$ propagates forward to acquire the *predicted belief* $p(x_k|\check{x}_0, v_{1:k}, y_{0:k-1})$, given the $v_k$ and motion model $f(\cdot)$. In the correction step, the predicted belief is updated, given the observation $y_k$ and the observation model $g(\cdot)$. This prediction-correction steps results in the posterior belief $p(x_k|\check{x}_0, v_{1:k}, y_{0:k})$.

However, Bayes filter in non-linear non-Gaussian case is intractable due to the fact that: 1) Infinite-dimension *space of probability density function* which requires infinite memory to represent the posterior belief. 2) The *computing of integral* requires infinite computing resources. To overcome the problem in the computing of Bayes filter, first, the probability density function should be *approximated via sampling* and constrained by surrogate models like Gaussian. Second, motion and observation models should be simplified via *linearization* for instance. Third, the integral should be *approximated via sampling* to facilitate the computing of the integral.

Hence, the linearization and approximation are crucial for the computing of the posterior belief of Bayes filter. The linearization results in the EKF [29] and IEKF [22]. The approximation via the *deterministic* Monte-Carlo sampling results in the sigmapoint Kalman filter (SPKF) and iterated SPKF (ISPKF) [22]. The approximation via *sampling the importance sampling method* based on Monte-Carlo sampling results in the particle filter (PF) and its variants [30]. We will elaborate them in section Posterior approximations: Linearization, sampling, and variational inference.

***Bayesian inference to compute the posterior (Batch discrete-time case).*** Bayes filter is based on the Markov property and cannot be extended to the batch discrete-time case. However, Bayesian inference can be used to the batch discrete-time case for non-linear non-Gaussian estimation. It shares the same steps as that in linear-Gaussian estimation, but the *linearization* of motion and observation models should be applied when switching the Bayesian inference from linear-Gaussian case to non-linear non-Gaussian case, as that in EKF.

Moreover, Bayesian inference can be extended from the discrete-time estimation to the continuous-time estimation [31]. The mechanism in the continuous-time estimation is almost the same as that in the discrete-time estimation because the Bayesian inference in discrete-time case can be seen as a special case of that in continuous-time case.

Other statistical methods to find the maximum posterior moments include the maximum a posterior (MAP) via Gauss-Newton [22], Maximum Likelihood via Gauss-Newton [32][22], and sliding-window filters (SWFs) [33]. The basic idea of MAP in linear-Gaussian case is to maximize the posterior distribution $p(x|v, y)$ to find the best state estimate (posterior mean) $\hat{x}$ by applying the Bayesian rules

$$\hat{x} = \arg\max_x p(x|v, y) = \arg\max_x \frac{p(y|x,v)p(x|v)}{p(y|v)} = \arg\max_x p(y|x)p(x|v) \quad (57)$$

where $p(y|x) = \prod_{k=0}^{K} p(y_k|x_k)$ and $p(x|v) = p(x_0|\check{x}_0) \prod_{k=1}^{K} p(x_k|x_{k-1}, v_k)$. The maximum likelihood only use the observations to find the maximum posterior without the prior. We assume the observation model is $y_k = g_k(x) + n_k$ where the observation noise $n_k \sim \mathcal{N}(0, R_k)$ and $k$ here is an arbitrary data index, instead of the time index. The objective function is defined as

$$J(x) = \frac{1}{2} \sum_k (y_k - g_k(x))^\top R_k^{-1} (y_k - g_k(x)) = -\log p(y|x) + C \quad (58)$$

where $C$ is a constant. The posterior mean is acquired by minimizing the objective function (maximizing the likelihood) via

$$\hat{x} = \arg\min_x J(x) = \arg\max_x \log p(y|x) \quad (59)$$

The MAP and maximum likelihood can be further solved by Gauss-Newton [22]. The Gauss-Newton iterates in entire trajectory, while IEKF iterates in one timestep. The SWFs finds the trade-off between Gauss-Newton and IEFK in terms of the iteration by iterating over many time-steps within a sliding window.

### 3.3 Posterior approximations: Linearization, sampling, and variational inference

#### 3.3.1 Linearization and extended Kalman filter

The extended Kalman Filter (EKF) can be derived from the Bayes filter by linearizing its motion and observation models, and constraining its probability density function via Gaussian. The linearization of models here is not the linearization of true state in linear-Gaussian estimation, but the *linearity in the current estimated state mean*. The benefits are two folds: 1) Nonlinear motion and observation models can be used in the estimation. 2) The noise (the Jacobians embedded in the covariance of the noise) can be applied to the nonlinear estimation. The disadvantage is that linearization brings high *inaccuracy* in posterior approximation especially when the models are largely non-linear.

During the linearization of EKF, the posterior belief, and noise are constrained to be Gaussian via

$$p(x_k|\check{x}_0, v_{1:k}, y_{0:k}) = \mathcal{N}(\hat{x}_k, \hat{P}_k) \quad (60)$$

$$w_k \sim \mathcal{N}(0, Q_k) \quad (61)$$

$$n_k \sim \mathcal{N}(0, R_k) \quad (62)$$

where $\hat{x}_k$ and $\hat{P}_k$ denote the posterior mean and covariance. The $w_k$ is the process noise and the $n_k$ is the measurement noise.

To linearize the current estimated state mean, the nonlinear models are rewritten as

$$\begin{cases} x_k = f(x_{k-1}, v_k, w_k) \approx \check{x}_k + F_{k-1}(x_{k-1} - \hat{x}_{k-1}) + w'_k \\ y_k = g(x_k, n_k) \approx \check{y}_k + G_k(x_k - \check{x}_k) + n'_k \end{cases} \quad (63)$$

where the current estimated state mean $\check{x}_k = f(\hat{x}_{k-1}, v_k, 0)$, $F_{k-1} = \frac{\partial f(x_{k-1}, v_k, w_k)}{\partial x_{k-1}}|_{\hat{x}_{k-1}, v_k, 0}$, and $w'_k = w_k \frac{\partial f(x_{k-1}, v_k, w_k)}{\partial w_k}|_{\hat{x}_{k-1}, v_k, 0}$. The current estimated observation $\check{y}_k = g(\check{x}_k, 0)$, $G_k = \frac{\partial g(x_k, n_k)}{\partial x_k}|_{\check{x}_k, 0}$, and $n'_k = n_k \frac{\partial g(x_k, n_k)}{\partial n_k}|_{\check{x}_k, 0}$.

Given the above linearization, the statistic properties of current state $x_k$ and observation $y_k$ are

$$\mathbb{E}[x_k] \approx \check{x}_k + F_{k-1}(x_{k-1} - \hat{x}_{k-1}) + \mathbb{E}[w'_k], \mathbb{E}[w'_k] = 0 \quad (64)$$

$$\mathbb{E}[(x_k - \mathbb{E}[x_k])(x_k - \mathbb{E}[x_k])^\mathsf{T}] \approx \mathbb{E}[w'_k {w'_k}^\mathsf{T}] = Q'_k \quad (65)$$

$$p(x_k|x_{k-1}, v_k) \approx \mathcal{N}(\check{x}_k + F_{k-1}(x_{k-1} - \hat{x}_{k-1}), Q'_k) \quad (66)$$

$$\mathbb{E}[y_k] \approx \check{y}_k + G_k(x_k - \check{x}_k) + \mathbb{E}[n'_k], \mathbb{E}[n'_k] = 0 \quad (67)$$

$$\mathbb{E}[(y_k - \mathbb{E}[y_k])(y_k - \mathbb{E}[y_k])^\mathsf{T}] \approx \mathbb{E}[n'_k {n'_k}^\mathsf{T}] = R'_k \quad (68)$$

$$p(y_k|x_k) \approx \mathcal{N}(\check{y}_k + G_k(x_k - \check{x}_k), R'_k) \quad (69)$$

Given the statistic properties, we recall the Bayes filter, we know that $\int p(x_k|x_{k-1}, v_k) p(x_{k-1}|\check{x}_0, v_{1:k-1}, y_{0:k-1}) dx_{k-1}$ is Gaussian and

$$\int p(x_k|x_{k-1}, v_k) p(x_{k-1}|\check{x}_0, v_{1:k-1}, y_{0:k-1}) dx_{k-1} = \mathcal{N}(\check{x}_k, F_{k-1}\hat{P}_{k-1}F_{k-1}^\mathsf{T} + Q'_k) \quad (70)$$

because the integral is still Gaussian after passing a Gaussian through a (stochastic) nonlinearity [22]. We also know that $\eta p(y_k|x_k) \int p(x_k|x_{k-1}, v_k) p(x_{k-1}|\check{x}_0, v_{1:k-1}, y_{0:k-1}) dx_{k-1}$ is also Gaussian and

$$\eta p(y_k|x_k) \int p(x_k|x_{k-1}, v_k) p(x_{k-1}|\check{x}_0, v_{1:k-1}, y_{0:k-1}) dx_{k-1} = \mathcal{N}(\check{x}_k + K_k(y_k - \check{y}_k), (1 - K_k G_k)(F_{k-1}\hat{P}_{k-1}F_{k-1}^\mathsf{T} + Q'_k)) \quad (71)$$

according to the normalized product of Gaussian probability density function. Finally, we arrive the predictor, Kalman gain, and corrector of EKF written as

$$\text{Predictor:} \begin{cases} \check{x}_k = f(\hat{x}_{k-1}, v_k, 0) \\ \check{P}_k = F_{k-1}\hat{P}_{k-1}F_{k-1}^\mathsf{T} + Q'_k \end{cases} \quad (72)$$

$$\text{Kalman gain: } K_k = \check{P}_k G_k^\mathsf{T} (G_k \check{P}_k G_k^\mathsf{T} + R'_k)^{-1} \quad (73)$$

$$\text{Corrector:} \begin{cases} \hat{x}_k = \check{x}_k + K_k(y_k - \check{y}_k) \\ \hat{P}_k = (1 - K_k G_k)\check{P}_k \end{cases} \quad (74)$$

where $y_k - \check{y}_k$ is the innovation. The performance of EKF can be further improved by iterative EKF (IEKF). This is achieved by replacing the estimated mean $\check{x}_k$ with the *arbitrary* operating point $x_{op,k}$.

However, the estimated mean $\check{x}_k$ or *arbitrary* operating point $x_{op,k}$ is just assumed to be the real prior mean, but they are not in reality. Moreover, the new probability density function may not be Gaussian after passing the probability density function through the non-linearity, but we assume it is Gaussian in linearization. These factors bring inaccuracy to the posterior approximation.

### 3.3.2 Approximation via sampling

*MC sampling, sigmapoint Kalman filter, and particle filter.* Monte-Carlo sampling or Monte-Carlo method follows a particular pattern: 1) Define the domain of inputs and generate inputs randomly from a probability distribution over the domain. 2) Perform a deterministic computation like counting on the inputs. 3) Aggregate the results for final approximation.

When passing a probability density function through a non-linearity, a large number of samples are first generated from the input density. Second, each sample is transformed through non-linearity. Third, the output density is built from the transformed samples. Overall, Monte-Carlo sampling suits any probability density function, any non-linear function without the need to know the mathematical form. More samples denote a higher accuracy but lower computation speed. The means of input and output are the same in linear case, while the means may change in non-linear case.

Monte-Carlo sampling grounds the *sigmapoint Kalman filter* (SPKF) where the samples are generated deterministically. The *prediction steps* of SPKF are as follows:

1) Convert the Gaussian representation $\{\mu_z, \Sigma_{zz}\}$ stacked with the prior and motion noise

$$\mu_z = \begin{bmatrix} \hat{x}_{k-1} \\ 0 \end{bmatrix}, \Sigma_{zz} = \begin{bmatrix} \hat{P}_{k-1} & 0 \\ 0 & Q_k \end{bmatrix} \quad (75)$$

to a sigmapoint representation $z_i$ via

$$z_0 = \mu_z, z_i = \mu_z + \sqrt{L+\kappa}\text{col}_i\mathbf{L}, z_{i+L} = \mu_z - \sqrt{L+\kappa}\text{col}_i\mathbf{L}, i = 1 \ldots L \quad (76)$$

where $\mathbf{L}$ is the lower-triangular and $\mathbf{LL}^\mathsf{T} = \Sigma_{zz}$. The $\kappa$ is a user-definable parameter which scales how far away the sigmapoints are from the mean and affects the accuracy of conversion. The dimension $L = \dim \mu_z$.

2) Unstack the sigmapoints representation and pass them through the nonlinear motion model

$$\check{x}_{k,i} = f(\hat{x}_{k-1,i}, v_k, w_{k,i}), z_i = \begin{bmatrix} \hat{x}_{k-1,i} \\ w_{k,i} \end{bmatrix} \quad (77)$$

3) Recombine the transformed sigmapoints to acquire the *predicted posterior* via

$$\check{x}_k = \sum_{i=0}^{2L} \alpha_i \check{x}_{k,i}, \check{P}_k = \sum_{i=0}^{2L} \alpha_i (\check{x}_{k,i} - \check{x}_k)(\check{x}_{k,i} - \check{x}_k)^\mathsf{T} \quad (78)$$

where $\alpha_i = \begin{cases} \frac{\kappa}{L+\kappa}, i = 0 \\ \frac{1}{2(L+\kappa)}, \text{otherwise} \end{cases}$.

4) The predicted posterior, observation noise and observation model replace the prior, motion noise, and motion model. Then the steps 1-2 repeat to compute the *desired moments*

$$\mu_{y,k} = \sum_{i=0}^{2L} \alpha_i \breve{y}_{k,i} \quad (79)$$

$$\Sigma_{yy,k} = \sum_{i=0}^{2L} \alpha_i (\breve{y}_{k,i} - \mu_{y,k})(\breve{y}_{k,i} - \mu_{y,k})^\top \quad (80)$$

$$\Sigma_{xy,k} = \sum_{i=0}^{2L} \alpha_i (\breve{x}_{k,i} - \breve{x}_k)(\breve{y}_{k,i} - \mu_{y,k})^\top \quad (81)$$

where $\breve{y}_{k,i}$ is the outcome of nonlinear observation model $g(\cdot)$.

In the *correction step*, the procedures are the same as that in Kalman filter where the predicted posterior is corrected by incorporating the observation $y_k$, and the posterior are

$$K_k = \Sigma_{xy,k} \Sigma_{y,k}^{-1} \quad (82)$$

$$\hat{x}_k = \breve{x}_k + K_k(y_k - \mu_{y,k}) \quad (83)$$

$$\hat{P}_k = \breve{P}_k - K_k \Sigma_{xy,k}^\top \quad (84)$$

The SPKF is improved by iterated SPKF (ISPKF) [34] by computing input sigmapoints around an operating point $x_{op,k}$ where $x_{op,k} = \begin{cases} \breve{x}_k, \text{first iteration} \\ \hat{x}_k, \text{rest iteration} \end{cases}$.

The *particle filter* is based on the Monte-Carlo sampling, and its steps are as follows:

1) Generate $M$ samples from the joint density by

$$\begin{bmatrix} \hat{x}_{k-1,m} \\ w_{k,m} \end{bmatrix} \leftarrow p(x_{k-1}|\breve{x}_0, v_{1:k-1}, y_{0:k-1}) p(w_k) \quad (85)$$

where $m$ is the particle index. The $p(x_{k-1}|\breve{x}_0, v_{1:k-1}, y_{0:k-1})$ is the prior density. The $p(w_k)$ is the motion noise density.

2) Generate the predicted posterior (*prediction step*) by passing samples and input $v_k$ through nonlinear motion model $f(\cdot)$

$$\breve{x}_{k,m} = f(\hat{x}_{k-1,m}, v_k, w_{k,m}) \quad (86)$$

3) Correct the predicted posterior (*correction step*) by incorporating observation $y_k$ in the *resampling* process (sampling the importance sampling)

$$\hat{x}_{k,m} \xleftarrow{resample} \{\breve{x}_{k,m}, W_{k,m}\} \quad (87)$$

where $W_{k,m}$ is the *importance weight* of predicted posterior (particle) based on the *divergence* between the desired posterior and predicted posterior

$$W_{k,m} = \frac{p(x_{k-1}|\breve{x}_0, v_{1:k-1}, y_{0:k})}{p(x_{k-1}|\breve{x}_0, v_{1:k-1}, y_{0:k-1})} = \eta p(y_k|\breve{x}_{k,m}) \quad (88)$$

where $\eta$ is a normalization constant. We assume $p(y_k|\breve{x}_{k,m})$ is the Gaussian $p(y_k|\breve{y}_{k,m})$, and $\breve{y}_{k,m}$ is acquired via simulating an expected sensor reading by passing $\breve{x}_{k,m}$ through nonlinear observation model $g(\cdot)$

$$\breve{y}_{k,m} = g(\breve{x}_{k,m}, 0) \quad (89)$$

To conclude, the SPKF and particle filter are classical examples where Monte-Carlo sampling is used to solve the Bayesian filter in nonlinear case. The samples in SPKF are generated deterministically because the sigmapoint transformation is deterministic, while the samples in particle filter are generated by sampling the importance sampling (resampling process). The particle filter is more accurate if there are as many samples as possible, but it is computationally expensive.

**MCMC sampling.** Instead of sampling the target probability density function $p(x)$ directly, we prefer to sample $q(x)$ which is an easy-to-sample proposal distribution and satisfies $p(x) = Mq(x), M < \infty$. Then, the $x^{(i)}$ sampled from the $q(x)$ are accepted if the acceptance condition is satisfied via the rejection sampling algorithm (Figure 2) [35]

```
Set i = 1
Repeat until i = N
    1. Sample x^(i) ~ q(x) and u ~ U_(0,1).
    2. If u < p(x^(i))/(Mq(x^(i))) then accept x^(i) and increment the counter i by
       1. Otherwise, reject.
```

Figure 2. The mechanism of the rejection sampling algorithm.

where $u \sim \mathcal{U}_{(0,1)}$ denotes sampling a uniform random variable on the interval $(0,1)$.

The stochastic process $x^{(i)} \in \mathcal{X} = \{x_1, x_1, \ldots, x_s\}$ is called a Markov chain if

$$p(x^{(i)}|x^{(i-1)}, \ldots, x^{(1)}) = T(x^{(i)}|x^{(i-1)}) \quad (90)$$

This means the evolution of the chain only depends on the *current state* and a *fixed transition matrix* $T$ in the space $\mathcal{X}$, and after $t$ iterations $\mu(x^{(1)})T^t$ converges to an invariant distribution $p(x)$

$$p(x) \leftarrow \mu(x^{(1)})T^t \quad (91)$$

where $\mu(x^{(1)})$ denotes the probability of the initial state $x^{(1)}$. The $T$ should have the *irreducibility* and *aperiodicity*, and invariant $p(x)$ should have the *reversibility*. The irreducibility denotes the $T$ cannot be reduced to separate smaller matrixes, while the aperiodicity denotes the chain cannot be trapped in cycles. The reversibility denotes

$$p(x^{(i)})T(x^{(i-1)}|x^{(i)}) = p(x^{(i-1)})T(x^{(i)}|x^{(i-1)}) \text{ or } p(x^{(i)}) = \sum_{x^{(i-1)}} p(x^{(i-1)})T(x^{(i)}|x^{(i-1)}) \quad (92)$$

The MCMC sampling algorithm can be seen as the combination of the rejection sampling algorithm, Markov chain, and Monte-Carlo sampling. The *Metropolis-Hastings* (MH) algorithm [36] and *Gibbs sampler* [37] are two representatives of the MCMC sampling.

```
1. Initialise x^(0).
2. For i = 0 to N − 1
    − Sample u ~ U_[0,1].
    − Sample x* ~ q(x*|x^(i)).
    − If u < A(x^(i), x*) = min{1, (p(x*)q(x^(i)|x*))/(p(x^(i))q(x*|x^(i)))}
         x^(i+1) = x*
      else
         x^(i+1) = x^(i)
```

Figure 3. The mechanism of Metropolis-Hastings algorithm.

The steps of Metropolis-Hastings is shown in Figure 3. First, the candidate sample $x^*$ is sampled from proposal distribution $q(x^*|x^{(i)})$ given the current sample $x^{(i)}$. Then, the Markov chain moves to the candidate sample which works as the current sample if acceptance condition $u < \mathcal{A}(x^{(i)}, x^*)$ is satisfied. The proposal distribution $q(x^*|x^{(i)})$ is self-designed, and it can be Gaussian distribution $\mathcal{N}(x^{(i)}, \sigma^2)$ for example. The choose of covariance $\sigma^2$ results in different width of proposal distribution, and therefore it should be balanced. Narrow proposal distribution means only one mode of $p(x)$ is visited, while wide proposal results in high rejection rate and subsequent high correlation of samples. Given the rejection sampling algorithm and Markov chain, the transition of Metropolis-Hastings can be constructed as

$$K_{HM}(x^{(i+1)}|x^{(i)}) = q(x^{(i+1)}|x^{(i)})\mathcal{A}(x^{(i)}, x^{(i+1)}) + \delta_{x^{(i)}}(x^{(i+1)})r(x^{(i)}) \quad (93)$$

where $r(x^{(i)}) = \int q(x^*|x^{(i)})(1 - \mathcal{A}(x^{(i)}, x^*))dx^*$ is a rejection-related term. The $\delta_{x^{(i)}}(x^{(i+1)})$ denotes the delta-Dirac mass located at $x^{(i)}$ and the proposal distribution $q(x^{(i+1)}|x^{(i)}) = \frac{1}{N}\sum_{i=1}^{N}\delta_{x^{(i)}}(x^{(i+1)})$. If the proposal distribution is independent of the current state $q(x^*|x^{(i)}) = q(x^*)$ and the proposal is assumed to be a symmetric random walk proposal $q(x^*|x^{(i)}) = q(x^{(i)}|x^*)$, the acceptance ratio $\mathcal{A}$ can be simplified further, and this results in the *independent sampler* and *Metropolis algorithm* respectively.

The Gibbs sampler is designed for the case of $n$-dimensional $x$ where $x = (x_1, x_2, \dots, x_n)$ and the probability density function $p(x)$ in Metropolis-Hastings algorithm is replaced by the *full conditions* expressed by

$$p(x) = p(x_j|x_1, \dots, x_{j-1}, x_{j+1}, \dots, x_n) = p(x_j|x_{-j}), j = 1, \dots, n \quad (94)$$

where $x_{-j} = (x_1, \dots, x_{j-1}, x_{j+1}, \dots, x_n)$ is the supplement of $x_j$ given the $n$-dimensional $x$ [38], and $p(x_j|x_{-j}), j = 1, \dots, n$ include $n$ joint distributions. If the full conditions are available and belong to the family of standard distributions like Gaussian, the new samples can be drawn *directly* by Figure 4.

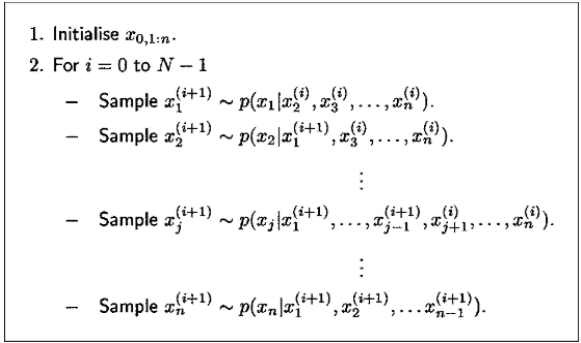

Figure 4. The mechanism of Gibbs sampler.

The Gibbs sampler can be seen as a special case of Metropolis-Hastings algorithm [35] where the proposal distribution of Gibbs sampler is

$$q(x^*|x^{(i)}) = \begin{cases} p(x_j^*|x_{-j}^{(i)}), \text{If } x_{-j}^* = x_{-j}^{(i)} \\ 0, \text{Otherwise} \end{cases} \quad (95)$$

and given $p(x)$ here refers to the full conditionals $p(x_j|x_{-j})$, the acceptance ratio $\mathcal{A}(x^{(i)}, x^*)$ is written as

$$\mathcal{A}(x^{(i)}, x^*) = \min\{1, \frac{p(x^*)q(x^{(i)}|x^*)}{p(x^{(i)})q(x^*|x^{(i)})}\} = 1 \quad (96)$$

This is the reason the new samples can be drawn directly without satisfying the acceptance condition. The high-dimensional joint distributions (conditions) $p(x_j|x_{-j}), j = 1, \dots, n$ can be factored into a directed graph, resulting in the directed acyclic graphs [39] which makes Gibbs sampling plausible for the approximation of Bayesian inference. In this case, the probability density function $p(x)$ and full conditions $p(x_j|x_{-j})$ are written as

$$p(x) = \prod_j p(x_j|x_{pa(j)}) \quad (97)$$

$$p(x_j|x_{-j}) = p(x_j|x_{pa(j)})\prod_{k\in ch(j)} p(x_k|x_{pa(k)}) \quad (98)$$

where $pa(j)$ and $ch(j)$ denote the parent and children's nodes of $x_j$ respectively. The parent $pa(j)$, children $ch(j)$, and children's parents $pa(k), k \in ch(j)$ are known as the *Markov basket*.

### 3.3.3 Approximation via variational inference

*Expectation-maximization (EM)*. The EM [40] algorithm is a classical iterative algorithm in maximum likelihood and MAP to estimate the local optimum of the parameters [41]. EM can be more efficient once it combines with the MCMC. EM can also be extended to the variational Bayes for cases with *large dataset* and *complex posterior models*. Given the variables or data $\{x, z\} \in \mathcal{X}$ where $x$ and $z$ denote the *observable* and *hidden (latent)* parts of the variables respectively, the local optimum of the likelihood $p(x|\theta)$ can be found by iterating the E step and M step respective to compute the expectation of complete likelihood or log likelihood $\log p(x, z|\theta)$ via

$$\text{E step: } Q(\theta) = \int [\log p(z, x|\theta)] p(z|x, \theta^{old}) \, dz \quad (99)$$

$$\text{M step: } \theta^{new} = \arg\max_\theta Q(\theta) \quad (100)$$

where $\theta$ is the parameter like Gaussian parameter, and $\theta^{old}$ the parameter value at previous time step. However, the computing of expectation in E step is intractable, and popular way to solve this is to approximate $p(z|x, \theta^{old})$ by sampling methods like MC and MCMC, resulting in the stochastic EM [42], MC-EM [43], and MCMC-EM [44]. For example, after the MCMC sampling like the HM, the E and M steps are simplified to

$$\text{E step: } \hat{Q}(\theta) = \frac{1}{N_i}\sum_{j=1}^{N_i} \log p(z^{(j)}, x|\theta) \quad (101)$$

$$\text{M step: } \theta^i = \arg\max_\theta \hat{Q}(\theta) \quad (102)$$

where $N_i$ denotes the number of samples, and $\hat{Q}(\theta)$ the estimated value of complete log likelihood.

*Variational inference (VI)*. The VI is an alternative approximation approach of MCMC sampling. Their key difference is that the MCMC approximates the posterior models via *sampling* with theoretical *guarantee* to the approximation accuracy, while the VI via the *optimization* without guarantee. The VI is *faster* than MCMC to handle the large dataset and complex posterior models like the Gaussian mixture model and conditionally conjugate model, but VI *underestimates* the variance of the posterior [14].

The optimization is achieved by positing a *family* of approximation density $Q$ over the hidden or latent variable $z$ where the latent variable help govern the data distribution, and finding the *best member* of this family $q^*(z)$ that minimizes the asymmetric and nonnegative Kullback-Leibler (KL) divergence to the posterior $p(z|x)$

$$q^*(z) = \arg\min_{q(z)\in Q} KL(q(z)||p(z|x)) \quad (103)$$

where $z = z_{1:n}$ and $x = x_{1:m}$ are the sets of variables. However, the $KL(q(z)||p(z|x))$ is not computable, because given the conditional density $p(z|x) = \frac{p(z,x)}{p(x)}$

$$KL(q(z)||p(z|x)) = \mathbb{E}[\log q(z)] - \mathbb{E}[\log q(z|x)] = \mathbb{E}[\log q(z)] - \mathbb{E}[\log p(z, x)] + \log p(x) \quad (104)$$

where the expectations are taken with respect to $q(z)$. The computing of $\log p(x)$ is intractable because the marginal density (evidence) $p(x) = \int p(z,x)dz$ cannot be computed in closed form. Instead of minimizing the $KL(q(z)||p(z|x))$ directly, we choose to maximize the $-KL(q(z)||p(z|x)) = \mathbb{E}[\log p(z,x)] - \mathbb{E}[\log q(z)] - \log p(x)$ which is still not computable because of $\log p(x)$. However, in the optimization, the $\log p(x)$ *is equivalent to a constant* with respect to the $q(z)$. Hence, if we can find the maximum of $\mathbb{E}[\log p(z,x)] - \mathbb{E}[\log q(z)]$, we can claim we find the maximum of $\mathbb{E}[\log p(z,x)] - \mathbb{E}[\log q(z)] - \log p(x)$ where $\mathbb{E}[\log p(z,x)] - \mathbb{E}[\log q(z)]$ is known as the evidence lower bound (ELBO)

$$ELBO(q) = \mathbb{E}[\log p(z,x)] - \mathbb{E}[\log q(z)] = \mathbb{E}[\log p(z)] + \mathbb{E}[\log p(x|z)] - \mathbb{E}[\log q(z)] = \mathbb{E}[\log p(x|z)] - KL(q(z)||p(z)) \quad (105)$$

The first part $\mathbb{E}[\log p(x|z)]$ is an expected likelihood, while the second part $-KL(q(z)||p(z))$ is the negative divergence of variational density $q(z)$ and prior density $p(z)$. Finally, the minimization of $KL(q(z)||p(z|x))$ turns to *find the balance* of $\mathbb{E}[\log p(x|z)]$ and $KL(q(z)||p(z))$ when maximizing $ELBO(q)$.

It is interesting to notice two facts: 1) $ELBO(q) = \mathbb{E}[\log p(z,x)] - \mathbb{E}[\log q(z)]$ where $\mathbb{E}[\log p(z,x)]$ is the objective of EM in E step. 2) Given $\log p(x) = KL(q(z)||p(z|x)) + ELBO(q)$, when $q(z) = p(z|x)$, $ELBO(q) = \log p(x)$. Given the above two facts that reveal the relation of EM and VI, the VI can be seen as the extension of EM when maximizing the likelihood over the latent variables, but they have the difference. The EM maximizes the expected complete likelihood $\mathbb{E}[\log p(z,x)]$, while VI tries to find the balance of $\mathbb{E}[\log p(z,x)]$ and $-\mathbb{E}[\log q(z)]$ in maximization, or maximizes the log marginal likelihood $\log p(x)$ when $q(z) = p(z|x)$.

The complexity of VI during the optimization relies on the complexity of the family. A classical family is the *mean-field variational family*. Its variational member is defined by

$$q(z) = \prod_{j=1}^{m} q_j(z_j) \quad (106)$$

where each latent variable $z_j$ is mutually independent and is governed by its own variational factor $q_j(z_j)$. The variational factors can be captured by Gaussian surrogate model or categorical model. A practical VI that relies on the mean-field variational family is the coordinate ascent variational inference (CAVI) [45]. As the Gibbs sampling, the *full conditional* of $z_j$ is defined by $p(z_j|\mathbf{z}_{-j}, x), j = 1, \dots, n$ where $\mathbf{z}_{-j} = (z_1, \dots, z_{j-1}, z_{j+1}, \dots, z_n)$ is the supplement of $z_j$ given $\mathbf{z}$. Other variational factors are written as $q_\ell(z_\ell), \ell \neq j$. Given the mean-field property (independent latent variables) and fixed $q_\ell(z_\ell)$, the optimal $q_j(z_j)$ is proportional to the exponentiated expected log of full conditional

$$q_j(z_j) \propto \exp\{\mathbb{E}_{-j}[\log p(z_j|\mathbf{z}_{-j}, x)]\} \propto \exp\{\mathbb{E}_{-j}[\log p(z_j, \mathbf{z}_{-j}, x)]\} \quad (107)$$

In this case, optimal $q_j(z_j)$ is equivalent to the proposal density in Gibbs sampling. Finally, the $ELBO(q) = \mathbb{E}[\log p(z,x)] - \mathbb{E}[\log q(z)]$ is maximized until the convergence to find the best $q(z)$. The CAVI can be extended to mixture models like the Gaussian mixture model where variational factors are mixed with Gaussians, and conditionally conjugate model where the global variables are the "parameters" and the local variables are per-data-point latent variables.

However, the CAVI is not so efficient in large dataset cases. In this case, the VI can combine with the natural gradients and stochastic optimization [46] to further improve its efficiency, resulting in the stochastic VI (SVI) [15].

### 3.4 Bayesian optimization

In the above section, the Bayesian posterior optimization is made under the premise that the (low-dimensional) samples are well-prepared or given, either in the batch sample case or in the case where the samples are collected recursively. We do not elaborate how to collect and use samples as little as possible to make the algorithms converge (evaluate the unknown function which is expensive to compute), therefore the maximum of the posterior estimation is found.

Here we review the Bayesian optimization (BO) which is to find the global maximum of an unknown function [47]. It models the objective function (unknown function captured by certain distributions like the Gaussian) as a random function which is used to determine informative sample *locations*. The GP-based Bayesian optimization with acquisition functions like the GP-Upper Confidence Bound (GP-UCB) [48] is a typical example which uses the posterior mean $\mu$ and variance $\sigma$ to compute the location of the next sample $a_n$ via maximizing the acquisition function

$$a_n = \arg\max_{a \in \mathcal{A}} \mu_{n-1}(a) + \beta_n^{\frac{1}{2}} \sigma_{n-1}(a) \quad (108)$$

where $\beta$ is an iteration-dependent scalar which reflects the confidence interval of the GP. Repeatedly evaluating the system at locations $\boldsymbol{a}$ improves the mean estimate of the underlying function and decreases the uncertainty at candidate locations for the maximum, such that the global maximum is provably found eventually. Other popular acquisition functions includes the probability of improvement, expected improvement, Bayesian expected losses, Thompson sampling and their hybrid variants [49]. BO makes the complex unknown function like high-dimensional unknown function easy to evaluate, but the trade-off of exploration and exploitation should be considered when exploring (collecting) data recursively.

## 4. model-based Bayesian RL

Model-based Bayesian estimation algorithms work well in simple and low-dimensional problems, but poor in small-scale and large-scale problems. The convergence of model-based Bayesian estimation approximation algorithms can be further improved by enjoying the advantages of RL in complex cases like the *large transition matrix case* and *partial observation scenarios*. Bayesian inference can combine with the model-based RL in small-scale problems, model-free RL in large-scale problems, and inverse RL in the searching of reward function, resulting in the *model-based Bayesian RL*, *model-free Bayesian RL*, and *Bayesian IRL*.

This section first gives the preliminaries of RL to ground the following algorithms. Second, classical model-based Bayesian RL is given. We focus on the Bayes-adaptive MDP (BA-MDP) and the methods to solve it.

Table 1. The definition and explanation of RL concepts.

| Name of the term | Mathematic form | Additional description |
|---|---|---|
| Expected value of the reward $R(s,a)$ | $\bar{r}(s,a) = \int r(s,a)q(dr|s,a)$ | - $r(s,a)$ is the realization of $R(s,a)$<br>- $R(s,a) \sim q(\cdot|s,a)$ is a random variable representing the reward obtained when action $a$ is taken in state $s$. |
| Transition density | $P^\mu(z'|z) = P(s'|s,a)\mu(a'|s')$ | - $P_0^\mu(z_0) = P_0(s_0)\mu(s_0|a_0)$<br>- $z = (s,a)$<br>- Policy $\mu: S \to A$ |
| Transition density on a trajectory $\xi$ | $Pr^\mu(\xi|\mu) = P_0^\mu(z_0)\prod_{t=1}^{T}P^\mu(z_t|z_{t-1})$ | --- |
| Reward discount (return) on a trajectory | $\bar{\rho}(\xi) = \sum_{t=0}^{T}\gamma^t\bar{r}(z_t)$ | - The discount $\gamma \in [0,1]$ |
| Expected return | $\mathbb{E}[\bar{\rho}(\xi)] = \int \bar{\rho}(\xi)Pr^\mu(\xi|\mu)d\xi$ | --- |
| Discounted return | $D^\mu(s) = \sum_{t=0}^{T}\gamma^t\bar{r}(z_t)|z_0 = (s,\mu(\cdot|s))$ | - $s_{t+1} = P^\mu(\cdot|s_t)$ |
| Value function | $V^\mu(s) = \mathbb{E}[D^\mu(s)]$ | --- |
| Action-value function | $Q^\mu(z) = \mathbb{E}[D^\mu(z)] = \sum_{t=0}^{T}\gamma^t\bar{r}(z_t)|z_0 = z$ | --- |
| Bellman equation for $V^\mu(s)$ | $V^\mu(s) = R^\mu(s) + \gamma\int P^\mu(s'|s)V^\mu(s')ds', s \in S$ | - $R^\mu(s)$ is the immediate reward |
| Bellman optimal equation | $V^*(s) = \max_{a \in A}\left[R^\mu(s) + \gamma\int P^\mu(s'|s)V^*(s')ds'\right], s \in S$ | - $V^*(s) = V^{\mu^*}(s)$ |
| Value iteration to find the optimal value function | $V_i^*(s) = \max_{a \in A}\left[R^\mu(s) + \gamma\int P^\mu(s'|s)V_{i-1}^*(s')ds'\right], s \in S$ | - Near all methods to find the optimal solution of an MDP are based on the *value iteration* and *policy iteration* of dynamic programming (DP) algorithms<br>- Policy iteration: The policy iteration consists of *policy evaluation* and *policy improvement* where value function is evaluated and improved until the convergence of value function |

## 4.1. Preliminaries of RL

The preliminaries of RL in this section include the basic definitions of RL, Markov decision process (MDP), and partial observable MDP (PO-MDP).

### 4.1.1 Definitions of RL

We list and elaborate the definitions of RL in Table 1 to ground the following algorithms.

### 4.1.2 Markov decision process (MDP)

MDP is based on the Markov chain and it has the Markov property: A stochastic process has the Markov property, if the conditional probability density functions (PDFs) of future states depend only upon current state, and not on any other past states. This means the future states only rely on the current state and have no relation to past states.

*MDP definition.* MDP is defined as a tuple $< S, A, P, P_0, q(\cdot|s,a) >$ where $S$ is the states, $A$ the actions, $P(\cdot|s,a)$ probability distribution over next states or transition density, $P_0$ the initial transitional probability distribution. $q(\cdot|s,a)$ a random variable which represents reward $R(s,a) \sim q(\cdot|s,a)$.

### 4.1.3 Partial observable MDP (POMDP)

*PO-MDP definition.* PO-MDP [16] is defined as a tuple $< S, A, O, P, \Omega, P_0, q(\cdot|s,a) >$. The difference of PO-MDP and MDP is $\Omega(\cdot|s,a) = P(O)$ where $\Omega$ is the transition density of the observation $O$ which is observed after executing action $a$.

The Bellman optimal equation of PO-MDP is written as

$$V^*(b_t) = \max_{a \in A}[\int R(s,a)b_t(s)ds + \gamma\int Pr(o|b_t,a)V^*(\tau(b_t,a,o))do] \quad (109)$$

where $b_t(s)$ is the information state (also called the *belief*, as the latent state in the Bayesian estimation) which represents the state under partial observation. The immediate reward cannot be obtained directly, but it can be computed indirectly by $\int R(s,a)b_t(s)ds$. Transition density also cannot be computed directly, therefore it is computed indirectly by $\int Pr(o|b_t,a)do$ conditioned on the historical observations and information state. The $\tau(b_t, a, o)$ represents the next information state, and it is defined and computed recursively by

$$\tau(b_t, a, o) = b_{t+1}(s') = \frac{\Omega(o_{t+1}|s', a_t)\int P(s'|s, a_t)b_t(s)ds}{\int \Omega(o_{t+1}|s'', a_t)\int P(s''|s, a_t)b_t(s)dsds''}$$ (110)

which describes the relationship of the information state, observation, and state. Bellman optimal equation of PO-MDP is proved to be piecewise-linear and convex, therefore it can be computed by

$$V^*(b_t) = \max_{\alpha \in \Gamma_t} \int \alpha(s) b_t(s) ds$$ (111)

where the linear segment $\Gamma_t = \{\alpha_0, \alpha_1.., \alpha_m\}$. The $\alpha$ is the $\alpha$-function, and $\alpha_i(b_t) = \int \alpha_i(s) b_t(s) ds$.

### 4.2. model-based Bayesian RL

The basic idea of model-based Bayesian RL is that the agent uses the collected data to first build a model of the domain's dynamics and then uses this model to optimize its policy. The advantage of model-based Bayesian RL is data-efficient in complex real-world small-scale problems. Its challenge is the exploration-exploitation trade-off. That is the balance between the need to explore the space of all possible policies, and the desire to focus data collection on trajectories that yield better outcomes.

This section focuses on the classical Bayes-adaptive MDP (BA-MDP) and the methods to solve it. These methods include the *approximation methods* and *exploration bonus methods*.

### 4.2.1 Bayes-adaptive MDP

In the PO-MDP, the belief (information state) after $t$ step trajectory is written as $b_{t+1}(s') = \frac{\Omega(o_{t+1}|s', a_t)\int P(s'|s, a_t)b_t(s)ds}{\int \Omega(o_{t+1}|s'', a_t)\int P(s''|s, a_t)b_t(s)dsds''}$. Recall that if the unknown variables $x$ is captured by two independent distributions $p_1(x)$ and $p_2(x)$, its probability density $p(x)$ is written as $p(x) = \eta p_1(x) p_2(x), \eta = (\int p_1(x) p_2(x) dx)^{-1}$ or $p(x|y_1, y_2) = \eta p(x|y_1) p(x|y_2), \eta = \frac{p(y_1)p(y_2)}{p(y_1, y_2)p(x)}$ in Bayesian context. Similarly, in the case of *belief-MDP* where the PO-MDP can be seen as one of its representatives, we assume the unknown transition $\theta$ is captured by two independent distributions $\theta_{s,a,s'}$ and $\theta_{s,a,r}$, and the belief is computed by $b_t(\theta) = \prod_{s,a} b_t(\theta_{s,a,r}) b_t(\theta_{s,a,s'})$ where $\theta_{s,a,s'}$ is the unknown probability transition from state $s$ to state $s'$ after the action $a$, and $\theta_{s,a,r}$ is the unknown probability to obtain the reward $r$. Hence, the belief $b_{t+1}(\theta')$ in the next time step after learning from the samples can be written as

$$b_{t+1}(\theta') = \eta \Pr(s', r|s, a, \theta') \int \Pr(s', \theta'|s, a, \theta) b_t(\theta) ds d\theta$$ (112)

where $\eta$ is the normalization factor, $\theta$ the unknown transition probability. The $\Pr(s', r|s, a, \theta')$ corresponds to the *updated reward probability*, while $\int \Pr(s', \theta'|s, a, \theta) b_t(\theta) ds d\theta$ corresponds to the *updated state transition probability*. The reward probability $\Pr(s', r|s, a, \theta')$ is computable, but the state transition probability $\int \Pr(s', \theta'|s, a, \theta) b_t(\theta) ds d\theta$ is intractable because it requires the computation over all observable states and possible belief states. This process is computationally expensive.

The computation of the intractable integral in the belief-MDP can be simplified by the BA-MDP where the probability densities are captured by the Dirichlet distribution.

*BA-MDP definition.* The BA-MDP [16][50][51] is defined by $<S', A, P', P'_0, R'>$ where $S'$ is the hyper-states (information states) $S \times \Phi$. $P'(\cdot|s, \phi, a)$ the transition distribution between hyper-states. Reward $R'(s, \phi, a) = R(s, a)$. Transition function $\Phi$ which is the posterior parameter. $\phi$ a realization of transition function $\Phi$ and $\Phi = \{\phi_{s,a}, \forall s, a \in S \times A\}$.

The BA-MDP assumes the structure form of parameter uncertainty to simplify the computation. That is, the posterior is assumed to be captured by the Dirichlet distribution. The *hyper-state transition density* $\Pr(s', \phi'|s, a, \phi)$ is factorized by

$$\Pr(s', \phi'|s, a, \phi) = \Pr(s'|s, a, \phi) \Pr(\phi'|s, a, s', \phi)$$ (113)

where the first part $\Pr(s'|s, a, \phi) = \frac{\phi'_{s,a,s'}}{\sum_{s'' \in S} \phi'_{s,a,s''}}$ by taking the expectation over all possible transition functions, and the second part $\Pr(\phi'|s, a, s', \phi) = \begin{cases} 1 \text{ if } \phi'_{s,a,s'} = \phi_{s,a,s'} + 1 \\ 0, otherwise \end{cases}$ since the update of $\phi$ to $\phi'$ is deterministic. Hence, the hyper-state transition density is written as

$$P'(s', \phi'|s, a, \phi) = \frac{\phi'_{s,a,s'}}{\sum_{s'' \in S} \phi'_{s,a,s''}} \Pr(\phi'|s, a, s', \phi) = \frac{\phi'_{s,a,s'}}{\sum_{s'' \in S} \phi'_{s,a,s''}} \mathbb{I}(\phi'_{s,a,s'} = \phi_{s,a,s'} + 1)$$ (114)

Then, we use the hyper-state transition density to compute the policy where the action is selected via the value function. Given the Bellman optimal equation, the optimal value function at a time step is written as

$$V^*(s, \phi) = \max_{a \in A}[R'(s, \phi, a) + \gamma \sum_{(s', \phi') \in S'} P'(s', \phi'|s, a, \phi) V^*(s', \phi')] = \max_{a \in A}[R(s, a) + \gamma \sum_{s' \in S'} P'(s', \phi'|s, a, \phi) V^*(s', \phi')]$$ (115)

To ensure the convergence of the policy, the value iteration is applied to compute the optimal policy of BA-MDP via

$$V_t^*(s, \phi) = \max_{a \in A}[R^\mu(s) + \gamma \int P^\mu(s'|s) V_{t-1}^*(s') ds'] = \max_{a \in A}[R(s, a) + \gamma \int P'(s', \phi'|s, a, \phi) V_{t-1}^*(s', \phi') ds' d\phi']$$ (116)

The advantages of BA-MDP are two folds: 1) It better handles the complex small-scale real-world problems than the Bayesian estimations. 2) The policy of BA-MDP is expressed over the information state which includes model uncertainty. This makes BA-MDP more explainable than other RL algorithms like the model-free RL.

However, the computation of integrals in the value iteration is still expensive. Another critical problem of the BA-MDP is the *exploration and exploitation dilemma*. That is, the balance of short-term information (instant reward) and long-term information which results in higher future reward. This problem is not only for BA-MDP, but also for all RL algorithms during the policy exploration. This problem can be mitigated via the *approximation methods* and *exploration bonus methods* in the next sections.

### 4.2.2 Value approximation methods

Here the value approximation methods includes the *offline value approximation*, *online near-myopic value approximation*, and *online tree search approximation*.

The *finite-state controllers* and *Bayesian exploration exploitation tradeoff in learning* (BEETLE) [16] are two representatives of the offline value approximation. The finite-state controllers use the graph to define BA-MDP where the nodes denote the memory states and the edges denote the observations. Then, the expected value is computed in closed form by recursively applying Bellman equations. The difference between BEETLE and PO-MDP is that the hyper-states in BEETLE are randomly sampled from interactions. However, this method is still computationally expensive.

The online near-myopic value approximation just accesses fewer hyper-states to find the policy. Therefore there is less requirement for computing resources. However, it causes suboptimal convergence of value estimation. The *Bayesian dynamic programming* and *the value of information heuristic (1-step estimation)* [16] are two representatives of the online near-myopic value approximation methods. The Bayesian dynamic programming samples a model from the posterior distribution over parameters, based on dynamic programming via the simulation. This method ignores the posterior uncertainty and causes slow convergence of the posterior. The convergence can be improved by keeping maximum likelihood estimations of the value function. The value of information heuristic considers the expected return and expected value. It uses the Dirichlet distribution to compute the distribution over the action value $Q^*(s, a)$ which is used to estimate the improvement of policy (1-step estimation). However, the 1-step estimation is myopia and results in suboptimal convergence.

The online tree search approximation methods are typically based on the classical *forward search tree* or the *Monte-Carlo tree search* (MCTS) [52], resulting in the Bayesian adaptive Monte-Carlo planning (BA-MCP) [53][54]. The forward search tree builds a fixed depth forward search tree which includes the hyper-states with limited steps (depth $d$) in a trajectory. The leaf nodes of the tree uses the immediate reward as the default value function which is necessary but naive. The BA-MCP incorporates two policies to traverse and grow the forward search tree. Two policies are the upper confidence bounds (UCT) applied to trees and the rollout. The action is selected by the UCT and value function via

$$a^* = \arg\max_a Q(s, h, a) + c\sqrt{\frac{\log n(s, h)}{n(s, h, a)}} \quad (117)$$

where $Q(s, h, a)$ is the value function which can be the immediate reward, $c$ the constant, $n(s, h)$ the number of times the node corresponding to state $s$ and history $h$ has been visited in the tree, $n(s, h, a)$ the number of times action $a$ was chosen in this node. An untried action is selected to start the rollout until the terminal node. The node visited will not be added to the tree, and $n(s, h)$ and $n(s, h, a)$ are saved in rollout process. Just one model is sampled from the posterior at the tree root to reduce the computation. The rollout is navigated to accelerate the search by a model-free estimation defined by

$$\hat{Q}(s_t, a_t) \leftarrow \hat{Q}(s_t, a_t) + \alpha(r_t + \gamma \max_a \hat{Q}(s_{t+1}, a) - \hat{Q}(s_t, a_t)) \quad (118)$$

### 4.2.3 Exploration Bonus Methods

The convergence of BA-MDP can be improved by the Bayesian exploration bonus (BEB) [55] which integrates the exploration bonus to the value function by

$$\tilde{V}_t^*(s, \phi) = \max_{a \in A}\left[R(s, a) + \frac{\beta}{1 + \sum_{s' \in S} \phi_{s,a,s'}} + \sum_{s \in S} P(s', \phi'|s, a, \phi)\tilde{V}_{t-1}^*(s', \phi')\right] \quad (119)$$

where $\frac{\beta}{1+\sum_{s'\in S}\phi_{s,a,s'}}$ is the exploration bonus. The $\beta$ is the constant that decides the magnitude of the bonus. The exploration bonus decays with $\frac{1}{n}$ where $n \sim \sum_{s' \in S} \phi_{s,a,s'}$. This makes BEB approach the optimal Bayesian solution $V^*(s, \phi)$ faster.

The exploration bonus can be the variance-based reward bonus (VBRB) [56] where the exploration bonus is decided by the posterior variance. Hence the modified value function is

$$\tilde{V}_t^*(s, \phi) = \max_{a \in A}[R(s, \phi, a) + \hat{R}_{s,\phi,a} + \sum_{s \in S} P(s', \phi'|s, a, \phi)\tilde{V}_{t-1}^*(s', \phi')] \quad (120)$$

where the bonus is $\hat{R}_{s,\phi,a}$ which is defined as

$$\begin{cases} \hat{R}_{s,\phi,a} = \beta_R \sigma_{R(s,\phi,a)} + \beta_P\sqrt{\sum_{s' \in S}\sigma^2_{P(s',\phi'|s,a,\phi)}} \\ \sigma^2_{R(s,\phi,a)} = \int R(s,\theta,a)^2 b(\theta)d\theta - R(s,\phi,a)^2 \\ \sigma^2_{P(s',\phi'|s,a,\phi)} = \int P(s'|s,\theta,a)^2 b(\theta)d\theta - P(s',\phi'|s,a,\phi)^2 \end{cases} \quad (121)$$

where $\beta_R$ and $\beta_P$ are constants to decide the magnitude of bonus. The variance-based exploration bonus bounds the exploration complexity to approach optimal Bayesian solution $V^*(s, \phi)$ faster.

## 5. Model-free Bayesian reinforcement learning

The agent of model-free Bayesian RL directly learns an optimal (or good) action-selection strategy from the collected data, as other RL algorithms. The advantage of model-free Bayesian RL over model-based Bayesian RL is its efficiency in cases where the solution space (e.g., policy space) exhibits more regularity than the underlying dynamics. The model-free RL is better suitable for the large-scale real-world problems than the model-based Bayesian RL, but it should manage the exploration-exploitation trade-off as the model-free Bayesian RL.

The classical model-free RL like the temporal difference learning (TD) [57], policy gradient (PG) [58] and actor-critic (AC) [59] can be integrated to the Bayesian framework, resulting in the value function Bayesian RL, Bayesian PG, and Bayesian AC.

### 5.1 Value function Bayesian RL

A typical representative of value function Bayesian RL is the Gaussian process temporal difference learning (GPTD) [60] which can be written as the Gaussian process regression (GPR) we elaborated in the Bayesian estimation section. Recall that the value function is defined to be the expected value of discounted return $V^\mu(s) = \mathbb{E}[D^\mu(s)]$, therefore the discounted return is decomposed to

$$D(s) = \mathbb{E}[D(s)] + D(s) - \mathbb{E}[D(s)] = V(s) + \Delta V(s), \Delta V(s) = D(s) - V(s) \quad (122)$$

Also recall that discounted return is defined by $D^\mu(s) = \sum_{t=0}^T \gamma^t \bar{r}(z_t)| z_0 = (s, \mu(\cdot|s)), s_{t+1} = P^\mu(\cdot|s_t)$. Hence, the discounted return of current state $s$ and next state $s'$ is defined by

$$D(s) = R(s) + \gamma D(s') \quad (123)$$

where $R(s)$ is the reword received in current state. Hence, it is easy to see

$$R(s) = D(s) - \gamma D(s') = V(s) + \Delta V(s) - \gamma(V(s') + \Delta V(s')) = V(s) - \gamma V(s') + \Delta V(s) - \gamma \Delta V(s') = V(s) - \gamma V(s') + N(s, s'), N(s, s') = \Delta V(s) - \gamma \Delta V(s') \quad (124)$$

Above equation can be rewritten as

$$\boldsymbol{R}_{T-1} = \boldsymbol{H}\boldsymbol{V}_T + \boldsymbol{N}_{T-1} \quad (125)$$

where $\boldsymbol{R}_T = (R(s_0), \ldots R(s_{T-1}))^\top$, $\boldsymbol{V}_{T+1} = (V(s_0), \ldots V(s_T))^\top$, $\boldsymbol{N}_T = (N(s_0, s_1), \ldots N(s_{T-1}, s_T))^\top$, and $\boldsymbol{H}$ is the $T \times T$ matrix

$$\boldsymbol{H} = \begin{bmatrix} 1 & -\gamma & 0 & \cdots & 0 & 0 \\ 0 & 1 & -\gamma & & 0 & 0 \\ \vdots & & & \ddots & & \vdots \\ 0 & 0 & 0 & \cdots & 1 & -\gamma \\ 0 & 0 & 0 & & 0 & 1 \end{bmatrix} \quad (126)$$

Recall the definition of GP in the Bayesian estimation section

$$\boldsymbol{Y}_T = \boldsymbol{H}\boldsymbol{F}_T + \boldsymbol{N}_T \quad (127)$$

$\boldsymbol{Y}_T = (y_1, y_2 \ldots y_T)^\top$, $\boldsymbol{F}_T = (F(x_1), F(x_2) \ldots F(x_T))^\top$, and $\boldsymbol{N}_T \sim \mathcal{N}(0, \boldsymbol{\Sigma})$ where $\boldsymbol{\Sigma}$ is the covariance of measurement noise. It is easy to model $\boldsymbol{R}_{T-1} = \boldsymbol{H}\boldsymbol{V}_T + \boldsymbol{N}_{T-1}$ to be a GP where the reward $\boldsymbol{R}_{T-1}$ is the observable measurement, $\boldsymbol{H}$ the linear transformation, and $\boldsymbol{V}_T$ the unknown function which we are going to use Bayesian inference to infer its posterior, and $\boldsymbol{N}_{T-1}$ the noise which can be modeled as Gaussian distribution $\boldsymbol{N}_{T-1} \sim \mathcal{N}(0, \boldsymbol{\Sigma})$. This method is called Monte-Carlo GPTD (MC-GPTD) [16]. Hence, by applying GP prior to $\boldsymbol{V}_T$ where $\boldsymbol{V}_T \sim \mathcal{N}(0, \boldsymbol{K}), \boldsymbol{K}_{i,j} = k(x_i, x_j)$. It is easy to obtain the posterior of $\boldsymbol{V}_T$

$$\begin{cases} \mathbb{E}[V(s)|\boldsymbol{R}_T] = \boldsymbol{k}(x)^\top \boldsymbol{\alpha} \\ Cov[V(s), V(s')|\boldsymbol{R}_T] = k(s, s') - \boldsymbol{k}(s)^\top \boldsymbol{C}\boldsymbol{k}(s') \end{cases} \quad (128)$$

where $\boldsymbol{\alpha}$ and $\boldsymbol{C}$ denote *sufficient statistics of the posterior moments* and

$$\boldsymbol{\alpha} = \boldsymbol{H}^\top (\boldsymbol{H}\boldsymbol{K}\boldsymbol{H}^\top + \boldsymbol{\Sigma})^\top (\boldsymbol{R}_T) \quad (129)$$

$$\boldsymbol{C} = \boldsymbol{H}^\top (\boldsymbol{H}\boldsymbol{K}\boldsymbol{H}^\top + \boldsymbol{\Sigma})^\top \boldsymbol{H} \quad (130)$$

**5.2 Bayesian policy gradient**

The Bayesian PG requires the expensive computation of the integral. This problem can be solved by the Bayesian quadrature (BQ) [61] which is a Bayesian approach to evaluate the integral via the samples of its integrand. We first elaborate the BQ, and then give the Bayesian PG.

*Bayesian quadrature.* Recall that the performance of RL policy in a trajectory is measured by the expected return $\eta(\mu) = \mathbb{E}[\bar{\rho}(\xi)] = \int \bar{\rho}(\xi) Pr^\mu(\xi|\mu) d\xi$ where the reward discount (return) is defined by $\bar{\rho}(\xi) = \sum_{t=0}^T \gamma^t \bar{r}(z_t), \gamma \in [0,1]$. The transition density on a trajectory is defined by $Pr^\mu(\xi|\mu) = P_0^\mu(z_0) \prod_{t=1}^T P^\mu(z_t|z_{t-1})$. The policy is updated by the gradient of expected return

$$\nabla \eta(\theta) = \int \bar{\rho}(\xi) \frac{\nabla \Pr(\xi;\theta)}{\Pr(\xi;\theta)} \Pr(\xi;\theta) d\xi = \int \bar{\rho}(\xi) \boldsymbol{u}(\xi;\boldsymbol{\theta}) \Pr(\xi;\theta) d\xi \quad (131)$$

where $\theta$ is the approximation of policy. The $\frac{\nabla \Pr(\xi;\theta)}{\Pr(\xi;\theta)}$ is the likelihood ratio of trajectory (Fisher score function), and it can be rewritten as

$$\boldsymbol{u}(\xi;\boldsymbol{\theta}) = \frac{\nabla \Pr(\xi;\theta)}{\Pr(\xi;\theta)} = \nabla \log \Pr(\xi;\theta) = \sum_{t=0}^{T-1} \nabla \log \mu(a_t|s_t;\theta) \quad (132)$$

It is hard to compute the $\nabla \eta(\theta)$ directly when $\nabla \eta(\theta)$ is in the form of integral $\int \bar{\rho}(\xi) \boldsymbol{u}(\xi;\boldsymbol{\theta}) \Pr(\xi;\theta) d\xi$. The BQ solves the computing of integral by simplifying the integral to the form

$$\zeta = \nabla \eta(\theta) = \int f(x) g(x) dx \quad (133)$$

where $f(x)$ is unknown and it is going to be solved by applying Bayesian inference to infer its posterior. The $g(x)$ is a known function. It is possible to model $f(x)$ as the GP by specifying the Normal prior distribution to $f(x)$. Therefore, $f(\cdot) \sim \mathcal{N}(\bar{f}(\cdot), k(\cdot,\cdot))$, and $\mathbb{E}[f(x)] = \bar{f}(\cdot), Cov[f(x), f(x')] = k(x, x')$.

Let samples $D_M = \{(x_i, y(x_i))\}_{i=1}^M$ where $y(x)$ is the noise-corrupted samples of $f(x)$. To infer the posterior of $f(x)$, it is possible to apply GP prior to $f(x)$ where $f(x) \sim \mathcal{N}(\bar{f}, \boldsymbol{K}), \boldsymbol{K}_{i,j} = k(x_i, x_j)$, therefore the mean and covariance (posterior) of $f(x)$ are

$$\begin{cases} \mathbb{E}[f(x)|D_M] = \bar{f}(x) + \boldsymbol{k}(x)^\top \boldsymbol{\alpha} \\ Cov[f(x), f(x')|D_M] = k(s, s') - \boldsymbol{k}(x)^\top \boldsymbol{C}\boldsymbol{k}(x') \end{cases} \quad (133)$$

where $\boldsymbol{\alpha}$ and $\boldsymbol{C}$ denote *sufficient statistics of the posterior moments* and

$$\boldsymbol{\alpha} = \boldsymbol{H}^\top (\boldsymbol{H}\boldsymbol{K}\boldsymbol{H}^\top + \boldsymbol{\Sigma})^\top (\boldsymbol{y} - \bar{\boldsymbol{f}}) \quad (134)$$

$$\boldsymbol{C} = \boldsymbol{H}^\top (\boldsymbol{H}\boldsymbol{K}\boldsymbol{H}^\top + \boldsymbol{\Sigma})^\top \boldsymbol{H} \quad (135)$$

where $\boldsymbol{y} = (y(x_1) \ldots y(x_M))^\top$, $\bar{\boldsymbol{f}} = (\bar{f}(x_1) \ldots \bar{f}(x_M))^\top$, $\boldsymbol{K}_{i,j} = k(x_i, x_j)$, and $\boldsymbol{k}(x) = (k(x_1, x), k(x_2, x) \ldots k(x_M, x))^\top$. Assuming that the integral $\zeta = \nabla \eta(\theta) = \int f(x) g(x) dx$ is linear, the posterior moments of $\zeta$ are given by

$$\begin{cases} \mathbb{E}[\zeta|D_M] = \int \mathbb{E}[f(x)|D_M] g(x) dx = \zeta_0 + \boldsymbol{b}^\top \boldsymbol{C}(\boldsymbol{y} - \bar{\boldsymbol{f}}) \\ Var[\zeta|D_M] = \iint Cov[f(x), f(x')|D_M] g(x) g(x') dx dx' \\ \qquad = b_0 - \boldsymbol{b}^\top \boldsymbol{C}\boldsymbol{b} \end{cases} \quad (136)$$

where $Var[\zeta|D_M]$ is the gradient variance, $\zeta_0 = \int \bar{f}(x) g(x) dx$, $\boldsymbol{b} = \int \boldsymbol{k}(x) g(x) dx$, and $b_0 = \iint k(x, x') g(x) g(x') dx dx'$.

*Bayesian policy gradient algorithm (BPG).* According to BQ, the gradient of BPG $\nabla \eta(\theta) = \int \bar{\rho}(\xi) \nabla \log \Pr(\xi;\theta) \Pr(\xi;\theta) d\xi$ can be simplified to the form $\int f(x) g(x) dx$ where

$$\begin{cases} f(x) = \bar{\rho}(\xi) \nabla \log \Pr(\xi;\theta) \\ g(x) = \Pr(\xi;\theta) \end{cases} \text{or} \begin{cases} f(x) = \bar{\rho}(\xi) \\ g(x) = \nabla \log \Pr(\xi;\theta) \Pr(\xi;\theta) \end{cases} \quad (137)$$

The $f(x)$ is an unknow function and it can be modeled as the GP. The $g(x)$ should be a known function, but in BPG, the trajectory transition function $\Pr(\xi;\theta)$ is unknown. However, it is sufficient to assign $\Pr(\xi;\theta)$ to $g(x)$ via a Fisher kernel, instead of using $\Pr(\xi;\theta)$ directly. When $f(x) = \bar{\rho}(\xi)\nabla\log\Pr(\xi;\theta)$ and $g(x) = \Pr(\xi;\theta)$, the Fisher kernel

$$k(\xi,\xi') = (1 + \boldsymbol{u}(\xi)^\intercal \boldsymbol{G}^{-1}\boldsymbol{u}(\xi'))^2, \boldsymbol{G}(\theta) = \mathbb{E}[\boldsymbol{u}(\xi)\boldsymbol{u}(\xi)^\intercal] \quad (138)$$

is selected to solve the integral, and $\boldsymbol{G}(\theta)$ is the Fisher information matrix. Finally, the moments of gradient are obtained via

$$\begin{cases} \mathbb{E}[\nabla\eta(\theta)|D_M] = \boldsymbol{YCb} \\ \boldsymbol{Var}[\nabla\eta(\theta)|D_M] = (b_0 - \boldsymbol{b}^\intercal \boldsymbol{Cb})\boldsymbol{I} \\ (\boldsymbol{b})_i = 1 + \boldsymbol{u}(\xi_i)^\intercal \boldsymbol{G}^{-1}\boldsymbol{u}(\xi_i) \\ b_0 = 1 + n \end{cases} \quad (139)$$

The computing of moments of gradient when $f(x) = \bar{\rho}(\xi)$ and $g(x) = \nabla\log\Pr(\xi;\theta)\Pr(\xi;\theta)$ is the same as the method above.

### 5.3 Bayesian actor-critic

The observation unit of policy gradient is the trajectory, while in actor-critic the observation is one-step transition. Hence, the expected return of the policy $\mu$ in the actor-critic may be written as

$$\eta(\mu) = \int w^\mu(z)\bar{r}(z)dz \quad (140)$$

where $z = (s,a)$, and $w^\mu(z) = \sum_{t=0}^\infty \gamma^t P_t^\mu(z)$ denotes the *discounting weight* of $z$. The discounting weight can be rewritten as $w^\mu(s) = \int w^\mu(s,a)da$ over the state $s$. The $P_t^\mu$ is the $t$-step state-action occupancy density defined by $P_t^\mu(z_t) = \int \prod_{i=1}^t P^\mu(z_i|z_{i-1}) dz_0 \ldots dz_{t-1}$. Therefore, the gradient of expected return in actor-critic is

$$\nabla\eta(\theta) = \int w^\mu(s,a;\theta)\nabla\mu(a|s;\theta)Q(s,a;\theta)\,dsda \quad (141)$$

In actor-critic, the actor is updated according to $\nabla\eta(\theta)$ to obtain new actor for generating new action and action value, while the critic provides action value $Q(s,a;\theta)$ for the update of the actor. The actor and critic contribute to the convergence of each other.

In Bayesian AC [62], the computing of integral $\int w(s;\theta)\nabla\mu(a|s;\theta)Q(s,a;\theta)\,dsda$ is simplified to the form $\int f(x)g(x)\,dx$ as that of Bayesian PG. Let

$$\begin{cases} f(x) = Q(z;\theta), z = (s,a) \\ g(x) = g(z;\theta) = w^\mu(z;\theta)\nabla\mu(a|s;\theta) \end{cases} \quad (142)$$

According to GPTD, it is easy to obtain the posterior moments of $Q(z;\theta)$ after $t$ time-steps

$$\begin{cases} \mathbb{E}[Q(z)|D_t] = \boldsymbol{k}_t(z)^\intercal \boldsymbol{\alpha}_t \\ Cov[Q(z),Q(z')|D_t] = k(z,z') - \boldsymbol{k}_t(z)^\intercal \boldsymbol{C}_t \boldsymbol{k}_t(z') \end{cases} \quad (143)$$

According to the linearization of the integral $\int w(s;\theta)\nabla\mu(a|s;\theta)Q(s,a;\theta)dsda$, it is easy to obtain the moments of policy gradient $\nabla\eta(\theta)$

$$\begin{cases} \mathbb{E}[\nabla\eta(\theta)|D_t] = \int g(z;\theta)\mathbb{E}[Q(z)|D_t]dz \\ \boldsymbol{Cov}[\nabla\eta(\theta)|D_t] = \int g(z;\theta)\boldsymbol{Cov}[Q(z),Q(z')|D_t]g(z';\theta)^\intercal dzdz' \end{cases} \quad (144)$$

This is the general form of $\nabla\eta(\theta)$ posterior, and the following computing of gradient posterior is almost the same as that of Bayesian PG. The Bayesian AC makes the best of Markov property of trajectory by step-based expected return, instead of the trajectory-based expected return in Bayesian PG. This reduces the variance of policy update, therefore Bayesian AC takes less samples to converge when comparing with Bayesian PG.

## 6. Bayesian RL in complex cases

The Bayesian RL can be extended to small-scale but complex cases. These cases typically include the unknown reward case, partial observable case, multi-agent case, multi-task case, non-linear nonlinear case, and the case with complex structured MDPs like the hierarchical MDPs. The focuses of this section are the unknown reward case, partial observable case, multi-agent case and multi-task case.

### 6.1 Unknown reward case

model-based Bayesian RL with unknown reward [56][16] is achieved by extending the hyper-states $\phi$ to unknown parameters $\theta$

$$\theta = (\phi,\vartheta), S' \in S \times \Theta \quad (145)$$

where $\vartheta$ is the prior over unknown reward function. Hence, modeling the reward function by suitable distribution is the key to forward. In simple cases, the reward function can be captured by the Beta distribution or Dirichlet distribution. In complex reward case, it can be captured by Gaussian distribution $\vartheta = \{\mu,\sigma\}$. The prior can be defined on the variance by

$$f(\psi) \propto \psi \exp(-\psi^2 \frac{\sigma_0^2}{2}) \quad (146)$$

where $f(\psi)$ is the prior density, and the precision $\psi = \frac{1}{\sigma}$. The prior can be also defined on the mean by

$$f(\mu) \propto \mathcal{N}(\mu_0,\sigma^2) \quad (147)$$

where $f(\mu)$ is the prior density. With the $n$ sampled rewards (observations), $f(\psi)$ and $f(\mu)$ switch into

$$f(\psi) \propto \psi^{n-1}\exp(-\psi^2 \frac{(n\hat{\sigma}+\sigma_0^2)}{2}) \quad (148)$$

$$f(\mu) \propto \mathcal{N}(\hat{\mu},\sigma^2/n) \quad (149)$$

where $\hat{\mu}$ and $\hat{\sigma}$ are the mean and variance in $n$ samples.

Hence, when sampling the posterior of parameters to compute the policy like the value function of model-based Bayesian RL, the posterior of the reward function should be sampled simultaneously.

### 6.2 Bayesian adaptive partial observable MDPs (BA-PO-MDPs)

The difference of BA-PO-MDP [63] and PO-MDP is that the BA-PO-MDP's hyper-states $S' \in S \times \Phi \times \Psi$ where $\Phi$ and $\Psi$ capture the space of Dirichlet distribution for the conjugate prior over *unknown transition distribution* and *observation* *distribution* respectively. In BA-PO-MDP, the hyper-state transition is factored by

$$P(s', \phi', \psi', o|s, \phi, \psi, a) = P(s'|s, \phi, a)P(o|s', \psi, a)P(\phi'|\phi, s, a, s')P(\psi'|\psi, o, a, s') \quad (150)$$

where

$$\begin{cases} P(\phi'|s, a, s', \phi) = \begin{cases} 1 \text{ if } \phi'_{s,a,s'} = \phi_{s,a,s'} + 1 \\ 0, \text{otherwise} \end{cases} \\ P(\psi'|\psi, o, a, s') = \begin{cases} 1 \text{ if } \phi'_{s,a,s'} = \phi_{s,a,s'} + 1 \\ 0, \text{otherwise} \end{cases} \\ P(s'|s, \phi, a) = \frac{\phi'_{s,a,s'}}{\sum_{s'' \in S} \phi'_{s,a,s''}} \\ P(o|s', \psi, a) = \frac{\phi'_{o,a,s'}}{\sum_{o \in O} \phi'_{o,a,s'}} \end{cases} \quad (151)$$

The optimal value function of BA-PO-MDP is defined by

$$V^*(b_t(s, \phi, \psi)) = \max_{a \in A}[R'(s, \phi, \psi, a)b_t(s, \phi, \psi) + \gamma \sum_{o \in O} P(o|b_{t-1}, a)V^*(\tau(b_t, a, o))] \quad (152)$$

which includes many beliefs because the hyper-states count is $\Phi \times \Psi$ in BA-PO-MDP, instead of $\Phi$ in POMDP. The value function estimation requires estimating the Bellman equation over all possible hyper-states for every belief. This means all models should be sampled from Dirichlet posterior. Then, models are solved and action is sampled from solved models. However, this estimation method is intractable. One possible way to solve this problem is selecting actions via the Bayes risk which is smallest expected loss defined by

$$BR(a) = \sum_{s,\phi,\psi} Q(b_t(s, \phi, \psi, a))b_t(s, \phi, \psi) - \sum_{s,\phi,\psi} Q(b_t(s, \phi, \psi, a^*))b_t(s, \phi, \psi) \quad (153)$$

This method generates a bound of the sample number, but provides a myopic view of uncertainty.

**6.3 Multi-agent case**

The single-agent RL focuses on finding the maximum reward of one agent from the environment, while the multi-agent RL focuses on finding the *global* maximum reward. In this process, each agent tries to maximum their own *interest* for higher reward. However, the interest of one agent may contradict the interests of other agents. It matters to coordinate the interest of each agent to find better balanced policies of agents for maximum global reward. The multi-agent RL faces four challenges basically: *1) Partial observable state.* Each agent just knowns its own state and the states of its neighbors. *2) Unknown system dynamics.* It's hard to search and find the best prior distribution for the system model, and the observations always mix with noise, bias, and error. *3) Exponential joint policy space.* The joint policy space expands exponentially with the increase of the agent number. This means exponential computations. *4) The coordination of the agent's policies.*

In the first challenge, the multi-agent RL can be modeled as a PO-MDP or BA-PO-MDP basically. In the second challenge, the overall dynamics can be decoupled into several local dynamics corresponding to each agent to solve PO-MDP, resulting in transition decoupled PO-MDP (TD-PO-MDP) [64]. In the third challenge, the MCTS can be applied to the PO-MDP for online planning, resulting in partially observable Monte Carlo planning (PO-MCP) [65]. The joint policy space can be reduced by the factored-value POMCP (FV-POMCP) [66] which decomposes the value function and global look-ahead tree into overlapping factors and multiple local look-ahead trees respectively. Moreover, the above tricks can be combined together, resulting in the transition-decoupled factored value based Monte-Carlo online planning (TD-FV-MCP) [67] to further improve the performance of multi-agent motion planning. However, the efficiency of searching and expanding the local look-ahead tree for each agent is poor, and there is a lack of efficient policy coordination among agents, resulting in suboptimal convergence in training. The work [68] attempts to solve these problem by applying the decentralized greedy search. Here we elaborate [68] as an example to deepen reader's understanding in multi-agent motion planning.

The graph like the undirected graph better models the MDP, and is becoming popular to describe the dependencies of the agent's states in different time steps. The works of multi-agent motion planning prefer to build on the undirected graph and realize the multi-agent training by combining the model-based Bayesian RL with Dirichlet distribution for online learning and MCTS for online planning. For instance, the [68] casts multi-robot patrolling problem as the BA-TD-POMDP formulation which is a tuple $< M, U, A, O, D, R, B, \gamma, H >$ where the planning process is modeled by the graph.

- $M = \{m_1, m_2..m_{|M|}\}$ is a set of robots.
- $U = \{U_1, U_2..U_{|M|}\}$ is the set of hyper-states $S$ and Dirichlet parameter $\varphi$ written as $U = < S, \varphi >$. The hyper-states $S = < S^V, S^I >$ where $S^V$ denotes the position state and $S^I$ the information state. The Dirichlet parameter $\varphi$ counts for the state transition.
- $A = \{A_1, A_2..A_{|M|}\}$ is the set of joint actions.
- $O = \{O_1, O_2..O_{|M|}\}$ is the set of observations, and $O = < O^V, O^I >$ where $S^V = O^V$ because position state is fully observed in this model.
- $D = \{D_1, D_2..D_{|M|}\}$ is joint expected system dynamics function (the objective). The $D_k$ is the dynamics function of the robot $m_k$, and

$$D_k = \begin{pmatrix} d_{11} & d_{12} & \cdots & d_{1K} \\ d_{21} & d_{22} & & d_{2K} \\ \vdots & & \ddots & \vdots \\ d_{K1} & d_{K2} & \cdots & d_{KK} \end{pmatrix} \quad (154)$$

$$d_{ij} = \sum_{I_n \in I} T_{\bar{\varphi}_k}^{I_i I_n} O_k^{I_n I_j}, T_{\bar{\varphi}_k}^{I_i I_n} = \mathbb{E}[T_k^{I_i I_n}|\bar{\varphi}_k] = \frac{\bar{\varphi}_k^{I_i I_n}}{\sum_{I_m \in I} \bar{\varphi}_k^{I_i I_m}}, \bar{\varphi}_k^{I_i I_n} = \varphi_k^{I_i I_n} + \delta^{I_i I_n} \quad (155)$$

where $T_{\bar{\varphi}_k}^{I_i I_n}$ is the expected state transition probability from information state $I_i$ to $I_n$. The $\delta^{I_i I_n}$ is the number of times the new state pair $(I_i, I_n)$ appears in $\bar{s}_k(t) = \{s_k(0), s_k(1)...s_k(t)\}$ which is the local state history visited by robot $m_k$. The $O_k^{I_n I_j}$ is the state observation probability from $I_n$ to $I_j$ and it is known and defined by

$$P^O = \begin{pmatrix} P_{11}^O & P_{12}^O & \cdots & P_{1K}^O \\ P_{21}^O & P_{22}^O & & P_{2K}^O \\ \vdots & & \ddots & \vdots \\ P_{K1}^O & P_{K2}^O & \cdots & P_{KK}^O \end{pmatrix} = \begin{pmatrix} P_1^O \\ P_2^O \\ \vdots \\ P_K^O \end{pmatrix} \quad (156)$$

The real system dynamics function $P^{IO}$ is defined by

$$P^{IO} = P^I P^O \cong D \quad (157)$$

where $P^I$ denotes the unknown state transition matrix and it is defined by

$$P^I = \begin{pmatrix} P^I_{11} & P^I_{12} & \cdots & P^I_{1K} \\ P^I_{21} & P^I_{22} & & P^I_{2K} \\ \vdots & & \ddots & \vdots \\ P^I_{K1} & P^I_{K2} & \cdots & P^I_{KK} \end{pmatrix} = \begin{pmatrix} P^I_1 \\ P^I_2 \\ \vdots \\ P^I_K \end{pmatrix} \quad (158)$$

- $R$ is the global immediate reward of all robots by taking joint action $a$ in the hyper state $u \in S$. It is defined by

$$R(u, a) = \sum_{m_k \in M} R_k(s_k, a_k) \quad (159)$$

- $B = \{B_1, B_2..B_{|M|}\}$ is the global belief state which is defined by $B = \{B_s, B_\varphi\}$. It represents a distribution over hyper states. The $B_\varphi$ is the belief state over count vector $\varphi$. The belief state $B_s$ is defined by $B_s = <B^V, B^I>$ where the state belief $B^V$ and information belief $B^I$ are belief states over position state and information state respectively. The information belief $B^I$ can be factored by

$$B^I_k = \{b^I_1, b^I_2, \ldots, b^I_{|V_k|}\}, b^I_n = \{p^I_1, p^I_2, \ldots, p^I_K\} \quad (160)$$

and information state at vertex $v$ of the graph is updated by

$$b^I_k(t+1) = \delta\big(b^I_k(t)\big) = \begin{cases} \Lambda D_k, v = v^{cur} \\ b^I_k(t)D_k, v \neq v^{cur} \end{cases} \quad (161)$$

where $\Lambda$ is a unit vector where first element is 1 and 0 otherwise. The $v^{cur}$ is the visited vertex by any robot.

Robot motion planning is achieved by *online learning* and *online planning* processes. In online learning process, Bayesian inference has two utilities:

1) Bayesian learning is used to infer and update the expected information state transition matrix $T$ by $T^{I_i I_n}_{\overline{\varphi}_k} = \mathbb{E}[T^{I_i I_n}_k | \overline{\varphi}_k] = \frac{\overline{\varphi}^{I_i I_n}_k}{\sum_{I_m \in I} \overline{\varphi}^{I_i I_m}_k}$ where transitional uncertainty is modeled by Dirichlet distribution. Hence, the system dynamics function $D$ works as the policy to select optimal actions for motion planning.

2) Bayesian learning is also used to infer the unknown information state $s_k$ based on the observation $o_k$ using posterior belief. First, the current prior belief $b(s_k)$ is computed by $b^I_k(t+1) = \delta\big(b^I_k(t)\big)$. Second, the posterior belief $b(s_k|o_k)$ is computed by

$$b(s_k|o_k) \leftarrow P^O_{s_k, o_k} b(s_k) \quad (162)$$

Third, the estimated information state $s'_k$ is computed by

$$s'_k \leftarrow \arg\max_{s_k} b(s_k|o_k) \quad (163)$$

Then the $s'_k$ is added to the local state history $\bar{s}_k(t)$.

In online planning process, first, the *Search* procedure is used to acquire optimal policy $\pi_k$ and value function $V_k$ by

$$(\pi_k, V_k) \leftarrow Search(h_k, \pi_{C_k}) \quad (164)$$

where $h_k$ denotes the local observation-action history. The $\pi_{C_k}$ is the policy of the prioritized robot set of robot $m_k$ which is defined as a set of robots allocated before robot $m_k$. Second, the $\pi_k$ and $V_k$ are transmitted to other robots to receive the policy and value of other robots $(\pi_{(k)}, V_{(k)})$ by the coordination graph

$$(\pi_{(k)}, V_{(k)}) \leftarrow Communication(\pi_k, V_k) \quad (165)$$

Third, the robot and policy with the largest value are stored after comparing the $V_k$ and $V_{(k)}$ by

$$m^*, \pi_{m^*} \leftarrow \arg\max_{m_k}(V_k, V_{(k)}) \quad (166)$$

If $m^*$ is the $m_k$, the $m_k$ keeps $\pi_k$ as its policy. Otherwise, $m^*$ and $\pi_{m^*}$ are added to prioritized robot set $C_k$ and $\pi_{C_k}$, and belief $B_k(h_k)$ is updated based on the $\pi_{C_k}$ by

$$C_k \leftarrow C_k \cup m^* \quad (167)$$

$$\pi_{C_k} \leftarrow \pi_{C_k} \cup \pi_{m^*} \quad (168)$$

$$B_k(h_k) \xleftarrow{\pi_{C_k}} B_k(h_k) \quad (169)$$

To acquire the optimal policy and value $(\pi_k, V_k)$ in the Search procedure, first, necessary states are sampled by

$$(s^V_k, \varphi_k) \sim <B^V_k(h_k), B^\varphi_k(h_k)> \quad (170)$$

$$s^I_k \sim B^I_k(h_k) \quad (171)$$

$$s_k \leftarrow <s^V_k, s^I_k> \quad (172)$$

Then, the *simulation* is used to expand the lookahead tree, while the estimated immediate marginal reward $\hat{R}_k$ is computed by the *Simulator G* of the simulation via

$$(s'_k, o^*_k, \hat{R}_k) \leftarrow G(s_k, \dot{D}, a_k, \pi_{C_k}, depth) \quad (173)$$

where $\dot{D}$ is sampled by $\dot{D} \leftarrow Dir(\varphi_k)$. Simulator $G$ is used to execute the action and get the observation and Marginal reward of each robot in *real* system via

$$(s_k, o_k) \leftarrow \dot{D}(\cdot,\cdot \, | s_k, \pi_{C_k}(t), a_k) \quad (174)$$

Marginal reward is defined by

$$\hat{R}_k = R_k - pe \quad (175)$$

where $pe$ is the penalty to prevent the asynchronous double counting problem when the robot $m_k$ visits a vertex $v$ before other prioritized robot set $m_{(k)}$. Optimal policy and value are obtained by

$$\pi^*_k(depth) \leftarrow (a^*_k, V^*_k) \leftarrow \arg\max_{a_k} V_k(h_k a_k o_k) \quad (176)$$

where $V_k$ is computed based on marginal reward. Meanwhile, the $h_k$ is updated by $h_k \leftarrow h_k a^*_k o^*_k$.

The tree expansion in the Simulation procedure requires the computing of the next observation-action history $h'_k$ to decide the expansion via

$$h'_k \leftarrow h_k a^*_k o^*_k \quad (177)$$

where $a^*_k$ here is computed by upper confidence bound

$$a^*_k \leftarrow \arg\max_{a_k}[V_k(h_k a_k o_k) + c\sqrt{\frac{\log(N(h_k)+1)}{N(h_k a_k o_k)}}] \quad (178)$$

where $N(h_k a_k o_k)$ denotes the number of times action $h_k a_k o_k$ was chosen in this node. The $o^*_k$ is computed by Simulator $G$. If $h'_k$ doesn't belong to the tree, the belief $B(h'_k)$ is computed via

$$B(h'_k) \leftarrow \delta(B(h_k)) \quad (179)$$

and the tree is expanded by initializing the leaf node via

$$Tree_k(h'_k) \leftarrow <N_{init}(h'_k), V_{init}(h'_k), B(h'_k)> \quad (180)$$

Meanwhile, the marginal reward of the next step $\hat{R}'_k$ is computed by the Rollout procedure which computes a simulated value for the node $h_k$ by executing the random policy via

$$\hat{R}'_k \leftarrow \hat{R}_k + \gamma Rollout(s'_k, h'_k, \pi_{c_k}, depth) \quad (181)$$

If $h'_k$ belongs to the tree, $\hat{R}_k$ is computed by searching the tree through the Simulation procedure until the depth is no less than $H$ via

$$\hat{R}'_k \leftarrow \hat{R}_k + \gamma Simulation(s'_k, \dot{D}, h'_k, \pi_{c_k}, depth + 1) \quad (182)$$

The $\hat{R}'_k$ is used to compute and update the value function $V(h'_k)$ via

$$V(h'_k) = \mathbb{E}_{\pi_k}[\sum_{d=0}^{H-1} \gamma^{depth} \hat{R}'_k(depth)] \quad (183)$$

$$V(h'_k) \leftarrow V(h'_k) + \frac{\hat{R}'_k - V(h'_k)}{N(h'_k)}, N(h'_k) \leftarrow N(h'_k) + 1 \quad (184)$$

The $N(h_k)$ also needs to be updated via

$$N(h_k) \leftarrow N(h_k) + 1 \quad (185)$$

The $V(h'_k)$ and $N(h_k)$ will be used in the simulation of the next round until the depth is no less than $H$.

### 6.4 Multi-task case for the generalization

The multi-task RL (MRL) aims to learn multiple tasks in parallel with the share representation, and what is learnt in each tasks contributes to the learning of other tasks. The MRL is a kind of transfer learning which improves model generalization by training with the domain knowledge of related tasks. For instance, the model trained with car images can be used to recognize the truck in the image classification. The transfer learning includes multiple methods, such as the *instance transfer*, *representation transfer*, or *parameter transfer*. There are various challenges the MRL faces. The challenges include the *scalability*, *distraction dilemma*, *partial observability*, *exploration/exploitation dilemma, catastrophic forgetting*, and *negative knowledge transfer* [69].

There are basically two types of MRL tasks: 1) RL tasks. 2) Inverse RL tasks. The difference between these two tasks is that the agent is fed with expert knowledge in inverse RL tasks, while the agent learns the knowledge explored by itself from the environment in RL tasks. The Bayesian inference can be applied to the above two MRL tasks. As a consequence, the knowledge explored by the agent itself or the expert knowledge which contains multiple tasks can be recognized or classified, and a policy which is adaptive to multiple tasks is acquired by learning from this classified knowledge.

For example, in RL tasks, [70] considers the dynamics environment as a sequential of stationary tasks. These multiple tasks is divided into different groups by clustering. The groups or environments parameterized by $\vartheta$ are modeled as the Chinese restaurant process (CRP). The posterior of each environment parameter $\vartheta$ is computed by Bayesian rules based on the CRP prior and the likelihood of the environment parameter. Once encountering a new environment, the posterior of the new environment will be computed to decide which group the new environment belongs to. Then, the stored knowledge in matched group will be used to learn policy $\theta$ in new environment. The stored knowledge is represented as the mixture of environment models (modeled by CPR) which is updated by EM algorithm [40] in an incremental manner where E-step in EM computes the posterior of environment parameter in new or updated environment, whereas the M-step updates environment parameters incrementally for future learning. In the inverse RL case, [71] preserves the adaptation functionalities of imitation learning but also propagates state variables and their uncertainties during real-time coordinated implementation. By leveraging the binary Gaussian process classification in the Bayesian framework, additional functionality, such as multiple task recognition is proposed to enhance the generalization capability.

## 7. Bayesian IRL

IRL [72][73] aims to find the reward function $r(s, a)$ given the expert demonstrations $Z = \{\zeta_1, \zeta_2, ..\}$ where $\zeta_i = \{(s_t, a_t)\}_{t=0}^{T}$ generated by optimal policy $\pi^*$. The reward function can be formulated by $r(s, a; f)$ parameterized by feature vector $f$

$$f \coloneqq [f_1, f_2, \ldots, f_N]^\top \quad (186)$$

The problem of IRL under MDP is to predict the reward function $f$ with dataset $\mathcal{D} \coloneqq (y, \mathcal{X})$ where $\mathcal{X}$ is the observed input and $y$ the noisy reward feature

$$y \coloneqq [f_1 + e_1, f_2 + e_2, \ldots, f_N + e_N]^\top \quad (187)$$

Assuming noise $e_i$ is independent and identically distributed like the Gaussian white noise $e_i \sim \mathcal{N}(0, \sigma^2)$. The reward feature can be treated as the realization of Gaussian random field with input $x$

$$f(x) \sim GP(\mu(x), k(x, x')) \quad (188)$$

Hence, the objective of Bayesian IRL with Gaussian distribution [74] is to acquire the estimated reward function (mean of posterior) $\hat{f}(x)$. This is achieved via the expectation of prior $f(x)$ given the dataset $\mathcal{D}$

$$\hat{f}(x) = \mathbb{E}(f(x)|\mathcal{D}) \quad (189)$$

Due to the assumption that $f(x)$ and $e_i$ are the Gaussian distributions, the noisy reward function

$$y \sim \mathcal{N}(\mu, K_y) \quad (190)$$

where $\mu = \mu(x_i)$, and $(K_y)_{ij} \coloneqq k(x_i, x_j) + \sigma^2 \delta_{ij}$. Hence, the posterior distribution of the reward function can be written as

$$f(x)|\mathcal{D} \sim \mathcal{N}(\mu_p, \sigma_p^2) \quad (191)$$

where the mean and variance of posterior are written as

$$\begin{cases} \mu_p = \mu + k(x)^\top K_y^{-1}(y - \mu) \\ \sigma_p^2 = k(x, x) - k(x)^\top K_y^{-1} k(x) \end{cases} \quad (192)$$

where $k(x) = (k(x_1, x), k(x_2, x)..k(x_T, x))^\top$.

The kernel $k(x, x')$ of the reward function can be parameterized as $k(x, x'; \theta)$ where $\theta$ is the hyperparameter. The $\theta$ can be acquired by maximizing the log likelihood (objective)

$$\theta = \arg\max_\theta \mathcal{L}(\theta|\mathcal{D}) \quad (193)$$

and the hyperparameter is learnt by the gradient ascent

$$\theta \leftarrow \theta + B\nabla_\theta \mathcal{L}(\theta|\mathcal{D}) \quad (194)$$

where $B$ is the recursive approximation of the *inverse Hessian*. The convergence in learning the hyperparameter can be further improved by $l_1$-regularizer, therefore the learning objective is written as

$$\text{minimize} - \mathcal{L}(\theta|\mathcal{D}) + \lambda\|w\|_1, \text{subject to} - \theta_i \leq 0, i = 1, \dots, P+2 \quad (195)$$

where $w = [w_1, w_1, \dots, w_P]^\top$ is the regression coefficient vector.

In complex non-linear non-Gaussian case, the approximation methods like the MCMC play an important role to approximate the posterior of reward function [75][76]. When the parameter space is high-dimensional, Bayesian optimization is applied to simplify the parameter space via projecting the parameters to a single point in a latent space. This ensures nearby points in the latent space correspond to reward functions yielding similar likelihoods.

## 8. Hybridization of Bayesian inference and RL

Bayesian inference can combine with model-based RL and model-free RL to explore better motion planning policy, resulting in model-based Bayesian RL and model-free Bayesian RL. The Bayesian inference can also hybridize with RL from other perspectives to improve the convergence of RL. The popular ways are the *Bayesian optimization for RL parameter tunning and parameter space reduction, variational inference in RL, and Bayesian-based RL objective*.

**8.1 Bayesian optimization for RL parameter tunning and parameter space reduction**

Here we first give the example of Bayesian optimization for RL parameter tuning, and then the BO for parameter space reduction.

*Parameter tunning.* In parameter space tunning case, [77] aims to find the optimal $d$-dimensional hyper-parameters in RL via $x^* = \arg\max_{x \in X} f(x, T_{max})$ where the unknown function $f$ is modeled as the GP. The training cost $\sum_{i=1}^N c(x_i, t_i)$ of evaluated settings $[x_i, t_i]$ is kept as low as possible. This is achieved by learning from the training curves via Bayesian optimization to find the optimal point

$$z_n = [x_n, t_n] = \arg\max_{x \in X, t \in [T_{min}, T_{max}]} \alpha(x,t)/\mu_c(x,t) \quad (196)$$

where the acquisition function $\alpha(x,t)$ is the expected improvement (EI). The $\mu_c(x,t) = \beta^T[x,t]$ is the mean of cost where $\beta = (Z^TZ)^{-1}Zc, Z = [x_i, t_i], c = [c_i]$. The optimization process is further improved by compressing the training curves by the Logistic function to the utility score which assigns higher values to the same performance if it is being maintained over more episodes. Finally, the optimal point $z_n$ is found by maximizing the GP log marginal likelihood with the sampled data from training curves via active learning [78].

*Parameter space reduction.* In parameter space reduction case, Bayesian optimization with GP-UCB is the popular but low-efficient way to reduce the parameter space and find the optimal parameters, as in [79]. [80] constrains the policy search space of RL to a sublevel-set of the Bayesian surrogate model's predictive uncertainty. The goal is to locally improve on the initial policy, such that

$$\theta_{n+1} = \arg\max_{\theta \in C_n} \alpha(\theta), C_n = \{\theta|\sigma_n(\theta) \leq \gamma\sigma_f\} \cap \Theta \quad (197)$$

where $C_n$ is the confidence region defined by a ball centered around $\theta_0$ with radius $r_0 = \|\theta - \theta_0\|$ determined by the condition $\sigma_0(\theta; r_0) = \gamma\sigma_f$. The $\theta_{n+1}$ is the next policy parameter. The $\alpha(\theta)$ is the acquisition function. The $\sigma_f$ is the signal standard deviation of the GP's kernel. The tunable parameter $\gamma \in (0,1]$ which governs the effective size of $C_n$. The $\Theta$ is the parameter space. The confidence region $C_n$ can grow adaptively after each rollout of the system in a cautious manner via

$$\|J(\theta_i) - J(\theta_j)\| \leq L_J\|\theta_i - \theta_j\|, \theta_i, \theta_j \in C_n \quad (198)$$

where $L_J$ denotes the Lipschitz continuity. Once the algorithm finds a local optimum, it explores the surrounding region until the global optimum.

The search space can be better reduced by a two-stage Bayesian optimization where the first stage uses GP surrogate model to create a reduced space (hyperplane) and the second stage takes another Bayesian optimization (knowledge gradient policy) with the particle filter to further reduce the search space and find the optimal parameters [81].

**8.2 Variational inference in RL**

Variational inference can derive the evidence lower bound which can constrain the learning process of RL, therefore improving the convergence. Variational inference can also approximate the posterior via the optimal variational family to improve the RL convergence directly. After that, the convergence can be further improved by penalizing the data with poor distribution.

*The evidence lower bound to constrain the RL learning.* As [9], [82] models the system as the PGM and computes the optimal parameters of PGM by maximizing the likelihood of dataset $\log \prod_{(\vec{x},\vec{m},\vec{a},\vec{r}) \in \mathcal{D}} p(\vec{x},\vec{m},\vec{O}^p|\vec{a}) = \sum_{(\vec{x},\vec{m},\vec{a},\vec{r}) \in \mathcal{D}} \log p(\vec{x},\vec{m},\vec{O}^p|\vec{a})$ where the Bayesian variational inference is used to maximize the log likelihood of the dataset (sensor inputs $\vec{x}$, mask $\vec{m}$, action $\vec{a}$, and the optimality $\vec{O}^p$ which is a reward-related binary random parameter). The $p$ denotes "post", and $\mathcal{D}$ denotes the dataset. The $\log p(\vec{x},\vec{m},\vec{O}^p|\vec{a})$ can be represented with latent state by $\log p(\vec{x},\vec{m},\vec{O}^p|\vec{a}) = \log \iint p(\vec{x},\vec{m},\vec{O}^p,\vec{z}^w,\vec{a}^p|\vec{a})d\vec{z}^wd\vec{a}^p$ where $\vec{z}^w$ represents the latent state and $w$ denotes "whole trajectory $\tau$ with length $H$". The $\log p(\vec{x},\vec{m},\vec{O}^p|\vec{a})$ can be further represented with *variational distribution* $q(\vec{z}^w, \vec{a}^p|\vec{x}, \vec{a})$ by $\log p(\vec{x},\vec{m},\vec{O}^p|\vec{a}) = \log \iint p(\vec{x},\vec{m},\vec{O}^p,\vec{z}^w,\vec{a}^p|\vec{a}) \frac{q(\vec{z}^w,\vec{a}^p|\vec{x},\vec{a})}{q(\vec{z}^w,\vec{a}^p|\vec{x},\vec{a})} d\vec{z}^wd\vec{a}^p$ where variational distribution $q(\vec{z}^w, \vec{a}^p|\vec{x}, \vec{a})$ is defined as $q(\vec{z}^w, \vec{a}^p|\vec{x}, \vec{a}) = q(\vec{z}|\vec{x},\vec{a})\pi(a_{\tau+H}|z_{\tau+H}) \prod_{t=\tau+1}^{\tau+H-1} p(z_{t+1}|z_t, a_t)\pi(a_t|z_t)$. The $\pi(a_{\tau+H}|z_{\tau+H}) \prod_{t=\tau+1}^{\tau+H-1} p(z_{t+1}|z_t, a_t)\pi(a_t|z_t)$ is the distribution of trajectory $\tau$ and it is acquired by executing policy $\pi$ with latent state transition $p$. The $q(\vec{z}|\vec{x},\vec{a})$ denotes the posterior of latent states. Finally, the log likelihood $\log p(\vec{x},\vec{m},\vec{O}^p|\vec{a})$ can be approximated by the $ELBO$

$$ELBO = \mathbb{E}_{q(\vec{z}^w, \vec{a}^p|\vec{x}, \vec{a})}[\log p(\vec{x}, \vec{m}, \vec{O}^p, \vec{z}^w, \vec{a}^p|\vec{a}) \\ - \log q(\vec{z}^w, \vec{a}^p|\vec{x}, \vec{a})] \leq \log p(\vec{x}, \vec{m}, \vec{O}^p|\vec{a})$$
(199)

The first part of $ELBO$ $\mathbb{E}_{q(\vec{z}^w, \vec{a}^p|\vec{x}, \vec{a})}[\log p(\vec{x}, \vec{m}, \vec{O}^p, \vec{z}^w, \vec{a}^p|\vec{a})]$ corresponds to the learning of generative models. The second part of $ELBO$ $\mathbb{E}_{q(\vec{z}^w, \vec{a}^p|\vec{x}, \vec{a})}[-\log q(\vec{z}^w, \vec{a}^p|\vec{x}, \vec{a})]$ corresponds to the learning of RL model with latent state $\vec{z}^w$. These two parts of $ELBO$ can be maximized jointly to improve the overall convergence of the algorithm.

*Variational posterior approximation to improve RL convergence.* [83] incorporates the model and action uncertainties to the model-based RL to improve the convergence. The model uncertainty is acquired by computing the posterior of the graph model parameterized by $\theta$ via

$$p(\theta|\mathcal{D}) \cong \frac{1}{E}\sum_i^E \delta(\theta - \theta_i) \quad (200)$$

where $E$ is the number of networks. The $\delta$ is the Dirac delta function. Each particle $\theta_i$ is independently trained by stochastic gradient descent to make $p(\theta|\mathcal{D}) \propto p(\mathcal{D}|\theta)p(\theta)$. The action uncertainty (policy with uncertainty) is described as the variational distribution $q(\boldsymbol{a}_{\geq t}; \phi)$ defined by

$$q(\boldsymbol{a}_{\geq t}; \phi^{(j)}) = \sum_{m=1}^M \pi_m^{(j)} \mathcal{N}(\boldsymbol{a}_{\geq t}; \boldsymbol{\mu}_m^{(j)}, \boldsymbol{\sigma}_m^{(j)}) \quad (201)$$

where $q$ is parameterized by neural networks $\phi^{(j)} := \{\pi_m^{(j)}, \boldsymbol{\mu}_m^{(j)}, \boldsymbol{\sigma}_m^{(j)}\}$. The $\pi$ is the policy of RL. The $\mu$ and $\sigma$ are the mean and variance. The $q(\boldsymbol{a}_{\geq t}; \phi^{(j)})$ approximates the intractable posterior (objective) $p(\boldsymbol{a}_{\geq t}, \boldsymbol{z}, \theta|\mathcal{O}_{\geq t}, \boldsymbol{x}_{\leq t})$ by minimizing $D_{KL}(q(\cdot)||p(\boldsymbol{a}_{\geq t}, \boldsymbol{z}, \theta|\mathcal{O}_{\geq t}, \boldsymbol{x}_{\leq t}))$ where $\boldsymbol{x}$ and $\boldsymbol{z}$ are observed and latent states. The $\mathcal{O} \in \{0,1\}$ describes the optimality of the latent state. Hence, the convergence is improved by maximizing the approximated objective $q(\boldsymbol{a}_{\geq t}; \phi)$ which is also used to select actions.

*Penalizing the data with poor distribution.* [84] incorporates uncertainty estimation to actor-critic algorithm to detect state-action pairs with the out-of-distribution (OOD) and down-weight their distribution in the training objectives. This results in better convergence of algorithms. The objective $p(\theta|X,Y) = p(Y|X,\theta)p(\theta)/p(Y|X)$ is computed by the dropout Bayesian variational inference where Q function is parameterized by $\theta$. $X$ capture all the state-action pairs in the training set, while $Y$ capture the true Q value of the states. Dropout Bayesian variational inference trains RL with dropout before every weight layer, and also performs dropout at test time, resulting in the variance defined by

$$Var[Q(s,a)] \approx \sigma^2 + \frac{1}{T}\hat{Q}_t(s,a)^\intercal \hat{Q}_t(s,a) - \mathbb{E}[\hat{Q}(s,a)]^\intercal [\hat{Q}(s,a)] \quad (202)$$

where $\sigma$ is the data noise. The $\frac{1}{T}\hat{Q}_t(s,a)^\intercal \hat{Q}_t(s,a)$ represents how much the model is uncertain about its predictions. The $\mathbb{E}[\hat{Q}(s,a)]$ is the mean. The uncertainty-weighted policy distribution $\pi'(a|s)$ is defined by inserting the variance $Var[Q(s,a)]$ via

$$\pi'(a|s) = \frac{\frac{\beta}{Var[Q_\theta^{\pi'}(s,a)]}\pi(a|s)}{Z(s)}, Z(s) = \int \frac{\beta}{Var[Q_\theta^{\pi'}(s,a)]}\pi(a|s)da$$
(203)

where $\pi(a|s)$ is the old policy distribution. Accordingly, the loss functions of actor and critic are inserted with the variance to decrease the probability of maximizing the Q function on OOD samples, and down-weigh the Bellman loss for the Q function respectively.

### 8.3 Bayesian-based RL objective

The RL convergence is expected to be improved by designing the Bayesian-based objective. We review and conclude two types of Bayesian-based RL objective: 1) Bayesian surprise/curiosity as the intrinsic reward (objective) for RL policy exploration. 2) The probably approximately correct (PAC) as the loss objective to bound the value approximation error.

*Bayesian surprise/curiosity.* The optimization of RL parameters can be achieved by maximizing the variational lower bound on future states log-likelihood

$$J = \mathbb{E}_{z_{t+1} \sim q(z)}[\log p_\theta(s_{t+1}|z_{t+1})] - \\ \beta D_{KL}[q_\theta(z_{t+1}|s_t, a_t, s_{t+1})||p_\theta(z_{t+1}|s_t, a_t)] \quad (204)$$

where $\log p_\theta(s_{t+1}|z_{t+1})$ is the reconstruction model which reconstructs the future state $s_{t+1}$ from the latent future state $z_{t+1}$. The $\beta$ is to control disentanglement in the latent representation. The variational distribution $q_\theta(z_{t+1}|s_t, a_t, s_{t+1})$ approximates the true posterior of the latent variable. The $p_\theta(z_{t+1}|s_t, a_t)$ is the latent prior. [19] takes the information gain $\mathcal{I}(z_{t+1}; s_{t+1}|s_t, a_t)$ approximated by the Bayesian surprise $D_{KL}[q_\theta(z_{t+1}|s_t, a_t, s_{t+1})||p_\theta(z_{t+1}|s_t, a_t)]$ as the intrinsic reward via

$$r_t^i = \mathcal{I}(z_{t+1}; s_{t+1}|s_t, a_t) \approx \\ D_{KL}[q_\theta(z_{t+1}|s_t, a_t, s_{t+1})||p_\theta(z_{t+1}|s_t, a_t)] = \\ \mathbb{E}_{q_\theta(z_{t+1}|\cdot)}[\log q_\theta(z_{t+1}|\cdot) - \log p_\theta(z_{t+1}|\cdot)] = \\ -H[q_\theta(z_{t+1}|\cdot)] + H[q_\theta(z_{t+1}|\cdot), p_\theta(z_{t+1}|\cdot)] \quad (205)$$

This means the maximization of parameters can be acquired by searching for states with minimal entropy of the posterior $-H[q_\theta(z_{t+1}|\cdot)]$ and a high cross-entropy value between the posterior and the prior $H[q_\theta(z_{t+1}|\cdot), p_\theta(z_{t+1}|\cdot)]$.

The work [20] takes the Bayesian curiosity to form the combined reward $r_t$ for RL via

$$r_t = e_t + \eta \cdot c_t, c_t = \log(\sigma^2(o_t)) \quad (206)$$

where $e_t$ is the immediate reward from the environment, and $c_t$ is the curiosity reward from Bayesian curiosity. The $\eta$ is a hyperparameter that controls the weight of the curiosity reward. the $\sigma^2(o_t)$ is the variance of the input $o_t$. The posterior moments (mean $\mu_N(\boldsymbol{o}_i)$ and variance $\sigma_N^2(\boldsymbol{o}_i)$ of the input) are computed by

$$\mu_N(\boldsymbol{o}_i) = \mathbf{m}_N^\intercal \phi_\psi(\boldsymbol{o}_i), \sigma_N^2(\boldsymbol{o}_i) = \beta^{-1} + \phi_\psi(\boldsymbol{o}_i)^\intercal \mathbf{S}_N \phi_\psi(\boldsymbol{o}_i) \quad (207)$$

where the network $\phi_\psi(\boldsymbol{o}_i)$ with weight $\psi$ captures nonlinearity and reduce dimensionality of the input. The $\beta$ is a hyperparameter representing noise precision in the data. The $\mathbf{m}_N$ and $\mathbf{S}_N$ are the posterior moments (mean and variance) after the training with $N$ samples where the prior is defined by Bayesian linear regression with Gaussian distribution. The network $\phi_\psi(\boldsymbol{o}_i)$ is trained by minimizing the negative log-likelihood loss.

*PAC loss objective.* The variational inference can derive the objective for RL training, but the approximation error is unknown in this process. The work [85] upper bounds the value approximation error (Bellman error) by the probably approximately correct (PAC) [86] via

$$\left\|Q_\theta^{\pi_\psi}(s,a) - Q^{\pi_\psi}(s,a)\right\|_{p_\infty^\psi,\pi_\psi}^2 \leq$$
$$\frac{1}{(1-\gamma)^2}\left\|\mathbb{E}_{a\sim\pi_\psi(\cdot|a)}[\mathcal{T}_\psi Q_\theta^{\pi_\psi}(s,a) - Q_\theta^{\pi_\psi}(s,a)]\right\|_{p_\infty^\psi,\pi_\psi}^2 \quad (208)$$

where $Q_\theta^{\pi_\psi}(s,a)$ with the network $\theta$ is the approximation of true value $Q^{\pi_\psi}(s,a)$. The $p_\infty^\psi$ is the stationary distribution of the state transition $p(\cdot|s,a)$ under policy $\pi_\psi$. The $\gamma$ is a discounted factor. The $\mathcal{T}_\psi$ is the Bellman backup operator. The upper bound of approximation error results in the PAC Bayesian bound which is used as the loss function of the SAC $L_{PAC}(\phi)$ simplified as

$$L_{PAC}(\phi) \coloneqq R_N(q;\phi) + \sqrt{\frac{D_{KL}(q(\theta;\phi)||p_0(\theta))}{N}} - \mathbb{E}_{\theta,s,a}\left[r + \gamma\mathbb{V}_{s'}[\mathbb{E}_{a'}[Q_\theta^{\pi_\psi}(s',a')]]\right] \quad (209)$$

where the first part $R_N(q;\phi)$ is the empirical risk defined as the soft Bellman error. The second part $\sqrt{\frac{D_{KL}(q(\theta;\phi)||p_0(\theta))}{N}}$ encourages the predictors (posteriors) $q(\theta;\phi)$ that are similar to the prior $p_0(\theta)$ where $N$ is the number of observations. The third part $\mathbb{E}_{\theta,s,a}\left[r + \gamma\mathbb{V}_{s'}[\mathbb{E}_{a'}[Q_\theta^{\pi_\psi}(s',a')]]\right]$ corrects a potential overestimation of the value of the current state.

**8.4 Other hybridizations of Bayesian inference and RL**

Apart from the above popular hybridizations of Bayesian inference and RL, the RL convergence can be also improved by the mix of environment dynamics and explored data, and the mix of control prior (deterministic mapping function) and RL.

*Mix of environment dynamics and explored data.* The work [87] presents a dual representation of environmental dynamics that combines the designer's knowledge and explored data to improve the convergence of RL. The empirical stochastic model can be written as

$$x' = f(x,u) + \xi^M \quad (210)$$

where $x'$ is the next state, and $u$ is the action. The $f$ is the deterministic part of dynamics. The $\xi^M$ is the additive stochastic uncertainty with unknown mean and covariance, but it follows the Gaussian distribution. However, the additive stochastic uncertainty exists in collected data set and it can be written as

$$\xi_j^\mathcal{D} = x_{j+1}^\mathcal{D} - f(x_j^\mathcal{D}, u_j^\mathcal{D}) \quad (211)$$

where $\xi_j^\mathcal{D}$ is the uncertainty in dataset $\mathcal{D} = \{(x_j^\mathcal{D}, u_j^\mathcal{D}, x_{j+1}^\mathcal{D})\}$. It is possible to incorporate $\xi_j^\mathcal{D}$ from the dataset to approximate the mean and covariance of unknown $\xi^M$ in an iterative Bayesian way when computing the mean and covariance of environmental dynamics via

$$\begin{bmatrix}\mu_k \\ \mathcal{K}_k\end{bmatrix} = IBE(\mu_{k-1}, \mathcal{K}_{k-1}, \xi_k^\mathcal{D}) \quad (212)$$

where $\mu_k$ and $\mathcal{K}_k$ are the mean and covariance of dynamics. The $IBE$ denotes the iterative Bayesian estimator. This results in the mixed stochastic model

$$x' = f(x,u) + \hat{\xi} \quad (213)$$

where $\hat{\xi}$ is the approximation of additive stochastic uncertainty due to the incorporation of uncertainty in dataset $\xi_j^\mathcal{D}$. Finally, the estimated next state $x'$ is used in policy iteration process of RL to improve the convergence.

*Mix of control prior (deterministic mapping function) and RL.* The work [88] fuses the model-free RL policy $\pi(\cdot|s_t)\sim\mathcal{N}(\mu_\pi,\sigma_\pi^2)$ with the control prior action distribution $\psi(\cdot|s_t)\sim\mathcal{N}(\mu_\psi,\sigma_\psi^2)$ to form a composite distribution $\phi(\cdot|s_t)\sim\mathcal{N}(\mu_\phi,\sigma_\phi^2)$ to improve the convergence of RL. The control priors are the deterministic mapping functions from state to action $a_t = \psi(s_t)$. The fusion of policies is achieved by factoring the fused distribution in the Bayesian context via

$$p(a|\theta_\pi,\theta_\psi) = \frac{p(\theta_\pi,\theta_\psi|a)p(a)}{p(\theta_\pi,\theta_\psi)} = \eta p(a|\theta_\pi)p(a|\theta_\psi) \quad (214)$$

where $\eta = \frac{p(\theta_\pi)p(\theta_\psi)}{p(\theta_\pi,\theta_\psi)p(a)}$. The $p(a|\theta_\pi,\theta_\psi)$ forms the hybrid policy $\phi$ where its mean and variance are computed by

$$\mu_\phi = \frac{\mu_\pi\sigma_\psi^2 + \mu_\psi\sigma_\pi^2}{\sigma_\pi^2 + \sigma_\psi^2}, \sigma_\phi^2 = \frac{\sigma_\pi^2\sigma_\psi^2}{\sigma_\psi^2 + \sigma_\pi^2} \quad (215)$$

## 9. Bayesian inference for better interpretability

The interpretability is widely accepted as the post-training explanations which elaborate how the system is understanding the environment or capture the dependencies between the time steps within an episode and that across different episodes. Therefore, the action selected by the policy is justified. Here we reviewed two interpretable motion planning algorithms based on the Bayesian inference and RL. The first algorithm provides the interpretability via decoding latent state to *sematic mask* (the bird-view semantics of the road map, routing, detected objects, and ego state of the agent), while the second algorithm derives the distribution of final rewards and deliver *time step importance* eventually.

*The interpretability via the sematic masks.* The work [9] builds on probabilistic graphical model (PGM) to represent conditional dependence between random variables, and formulate the sequential latent driving environment and maximum entropy RL process. Therefore, four types of model are acquired: 1) Generative models $p(x_t|z_t)$ and $p(m_t|z_t)$. The $p(x_t|z_t)$ decodes the latent state to observed state, while $p(m_t|z_t)$ decodes latent state to *sematic mask (the bird-view semantics of the road map, routing, detected objects, and ego state of the agent)* which provides interpretability showing how the system is understanding the environment semantically. 2) Latent dynamic model $p(z_{t+1}|z_t,a_t)$ which predicts the next latent state based on current latent state. 3) Inference model $p(z_{t+1}|x_{t+1},a_t)$ which infers the latent state based on observed state. 4) Policy model $\pi(a_t|z_t)$ which selects optimal actions based on latent state.

The optimal parameters of PGM can be acquired by maximizing the likelihood of dataset. Hence, here the Bayesian variational inference is used to maximize the log likelihood of the dataset (sensor inputs $\vec{x}$, mask $\vec{m}$, action $\vec{a}$, and the optimality $\vec{O}^p$ which is a reward-related binary random parameter) by

$$\log \prod_{(\vec{x},\vec{m},\vec{a},\vec{r})\in\mathcal{D}} p(\vec{x},\vec{m},\vec{O}^p|\vec{a}) = \sum_{(\vec{x},\vec{m},\vec{a},\vec{r})\in\mathcal{D}} \log p(\vec{x},\vec{m},\vec{O}^p|\vec{a})$$

(216)

where $p$ denotes "post", and $\mathcal{D}$ denotes the dataset. The $\log p(\vec{x}, \vec{m}, \vec{O}^p | \vec{a})$ can be represented with latent state by

$$\log p(\vec{x}, \vec{m}, \vec{O}^p | \vec{a}) = \log \iint p(\vec{x}, \vec{m}, \vec{O}^p, \vec{z}^w, \vec{a}^p | \vec{a}) d\vec{z}^w d\vec{a}^p \quad (217)$$

where $\vec{z}^w$ represents the latent state where $w$ denotes "whole trajectory $\tau$ with length $H$". The $\log p(\vec{x}, \vec{m}, \vec{O}^p | \vec{a})$ can be further represented with *variational distribution* $q(\vec{z}^w, \vec{a}^p | \vec{x}, \vec{a})$ by

$$\log p(\vec{x}, \vec{m}, \vec{O}^p | \vec{a}) = \log \iint p(\vec{x}, \vec{m}, \vec{O}^p, \vec{z}^w, \vec{a}^p | \vec{a}) \frac{q(\vec{z}^w, \vec{a}^p | \vec{x}, \vec{a})}{q(\vec{z}^w, \vec{a}^p | \vec{x}, \vec{a})} d\vec{z}^w d\vec{a}^p \quad (218)$$

where variational distribution $q(\vec{z}^w, \vec{a}^p | \vec{x}, \vec{a})$ is defined as

$$q(\vec{z}^w, \vec{a}^p | \vec{x}, \vec{a}) = q(\vec{z}|\vec{x},\vec{a})\pi(a_{\tau+H}|z_{\tau+H}) \prod_{t=\tau+1}^{\tau+H-1} p(z_{t+1}|z_t, a_t)\pi(a_t|z_t) \quad (219)$$

where $\pi(a_{\tau+H}|z_{\tau+H}) \prod_{t=\tau+1}^{\tau+H-1} p(z_{t+1}|z_t, a_t)\pi(a_t|z_t)$ the distribution of trajectory $\tau$ is acquired by executing policy $\pi$ with latent state transition $p$. The $q(\vec{z}|\vec{x},\vec{a})$ denotes the posterior of latent states. Finally, the log likelihood $\log p(\vec{x}, \vec{m}, \vec{O}^p | \vec{a})$ can be represented as

$$\log p(\vec{x}, \vec{m}, \vec{O}^p | \vec{a}) = \log \mathbb{E}_{q(\vec{z}^w, \vec{a}^p|\vec{x},\vec{a})}\left[\frac{p(\vec{x},\vec{m},\vec{O}^p,\vec{z}^w,\vec{a}^p|\vec{a})}{q(\vec{z}^w,\vec{a}^p|\vec{x},\vec{a})}\right] \geq$$
$$\mathbb{E}_{q(\vec{z}^w, \vec{a}^p|\vec{x},\vec{a})}[\log p(\vec{x},\vec{m},\vec{O}^p,\vec{z}^w,\vec{a}^p|\vec{a}) - \log q(\vec{z}^w,\vec{a}^p|\vec{x},\vec{a})] = ELBO \quad (220)$$

where $ELBO$ denotes the *evidence lower bound*, and "$\geq$" is obtained by *Jensen's inequality*.

In the first part of $ELBO$, the $p(\vec{x}, \vec{m}, \vec{O}^p, \vec{z}^w, \vec{a}^p | \vec{a})$ can be factorized as

$$p(\vec{x}, \vec{m}, \vec{O}^p, \vec{z}^w, \vec{a}^p | \vec{a}) =$$
$$p(\vec{x}, \vec{m}, \vec{O}^p, z_{\tau+2:\tau+H}, \vec{a}^p | \vec{z}, \vec{a}) p(\vec{z}|\vec{a}) =$$
$$p(\vec{x}|\vec{z})p(\vec{m}|\vec{z})p(\vec{O}^p, z_{\tau+2:\tau+H}, \vec{a}^p | z_{\tau+1}) p(\vec{z}|\vec{a}) =$$
$$p(\vec{x}|\vec{z})p(\vec{m}|\vec{z})\frac{p(\vec{O}^p, \vec{z}^p, \vec{a}^p)}{p(z_{\tau+1})} p(\vec{z}|\vec{a}) \quad (221)$$

Due to the definition of optimality $p(O_t|z_t, a_t) = \exp(r(z_t, a_t))$ and the soft optimality assumption, the $p(\vec{O}^p, \vec{z}^p, \vec{a}^p)$ can be factorized as

$$p(\vec{O}^p, \vec{z}^p, \vec{a}^p) = p(\vec{z}^p, \vec{a}^p)p(\vec{O}^p|\vec{z}^p, \vec{a}^p) =$$
$$p(\vec{a}^p)p(\vec{z}^p)p(\vec{O}^p|\vec{z}^p, \vec{a}^p) =$$
$$p(\vec{a}^p)[p(z_{\tau+1})\prod_{t=\tau+1}^{\tau+H-1} p(z_{t+1}|z_t, a_t)]\exp(\sum_{t=\tau+1}^{\tau+H} r(z_t, a_t)) \quad (222)$$

Hence, the first part of $ELBO$ can be represented by
$$p(\vec{x}, \vec{m}, \vec{O}^p, \vec{z}^w, \vec{a}^p | \vec{a}) = p(\vec{x}|\vec{z})p(\vec{m}|\vec{z})p(\vec{a}^p)$$
$$[\prod_{t=\tau+1}^{\tau+H-1} p(z_{t+1}|z_t, a_t)]\exp(\sum_{t=\tau+1}^{\tau+H} r(z_t, a_t))p(\vec{z}|\vec{a}) \quad (223)$$

Finally, the $ELBO$ can be represented by

$$ELBO = \mathbb{E}_{q(\vec{z}^w, \vec{a}^p|\vec{x},\vec{a})}[\log p(\vec{x}|\vec{z}) + \log p(\vec{m}|\vec{z}) +$$
$$\log p(\vec{a}^p) + \log p(\vec{z}|\vec{a}) + \log \prod_{t=\tau+1}^{\tau+H-1} p(z_{t+1}|z_t, a_t) +$$
$$\sum_{t=\tau+1}^{\tau+H} r(z_t, a_t) - \log q(\vec{z}|\vec{x},\vec{a}) - \log \pi(a_{\tau+H}|z_{\tau+H}) -$$
$$\log \prod_{t=\tau+1}^{\tau+H-1} p(z_{t+1}|z_t, a_t) - \log \prod_{t=\tau+1}^{\tau+H-1} \pi(a_t|z_t)] =$$
$$\mathbb{E}_{q(\vec{z}^w, \vec{a}^p|\vec{x},\vec{a})}[\log p(\vec{x}|\vec{z}) + \log p(\vec{m}|\vec{z}) + \log p(\vec{z}|\vec{a}) -$$
$$\log q(\vec{z}|\vec{x},\vec{a})] + \mathbb{E}_{q(\vec{z}^w, \vec{a}^p|\vec{x},\vec{a})}[\sum_{t=\tau+1}^{\tau+H}[r(z_t, a_t) -$$
$$\log \pi(a_t|z_t) + \log p(a_t)]] \quad (224)$$

The maximum likelihood can be acquired by maximizing the $ELBO$. The maximization of first part of $ELBO$ $\mathbb{E}_{q(\vec{z}^w, \vec{a}^p|\vec{x},\vec{a})}[\log p(\vec{x}|\vec{z}) + \log p(\vec{m}|\vec{z}) + \log p(\vec{z}|\vec{a}) - \log q(\vec{z}|\vec{x},\vec{a})]$ results in the learning of latent environment model, while the maximization of second part of $ELBO$ $\mathbb{E}_{q(\vec{z}^w, \vec{a}^p|\vec{x},\vec{a})}[\sum_{t=\tau+1}^{\tau+H}[r(z_t, a_t) - \log \pi(a_t|z_t) + \log p(a_t)]]$ results in the learning of RL model. This means the latent environment model and RL model can be learned jointly.

In the first part of $ELBO$, due to the Markov property, the $\log p(\vec{x}|\vec{z})$, $\log p(\vec{m}|\vec{z})$, $\log p(\vec{z}|\vec{a})$ and $\log q(\vec{z}|\vec{x},\vec{a})$ are unfolded as

$$\log p(\vec{x}|\vec{z}) = \log \prod_{t=1}^{\tau+1} p(x_t|z_t) = \sum_{t=1}^{\tau+1} \log p(x_t|z_t) \quad (225)$$

$$\log p(\vec{m}|\vec{z}) = \log \prod_{t=1}^{\tau+1} p(m_t|z_t) = \sum_{t=1}^{\tau+1} \log p(m_t|z_t) \quad (226)$$

$$\log p(\vec{z}|\vec{a}) = \log[p(z_1)\prod_{t=1}^{\tau} p(z_{t+1}|z_t, a_t)] = \log p(z_1) +$$
$$\sum_{t=1}^{\tau} \log p(z_{t+1}|z_t, a_t) \quad (227)$$

$$\log q(\vec{z}|\vec{x},\vec{a}) = \log\left[q(z_1|\vec{x},\vec{a})\prod_{t=1}^{\tau} q(z_{t+1}|z_t, \vec{x},\vec{a})\right]$$
$$= \log\left[q(z_1|x_1)\prod_{t=1}^{\tau} q(z_{t+1}|z_t, x_{t+1}, a_t)\right]$$
$$= \log q(z_1|x_1) + \sum_{t=1}^{\tau} \log q(z_{t+1}|z_t, x_{t+1}, a_t)$$
(228)

Finally, the $q(\vec{z}|\vec{x},\vec{a})$ replaces $q(\vec{z}^w, \vec{a}^p|\vec{x},\vec{a})$ because the first of $ELBO$ only relates to the latent states, and the $\mathbb{E}_{q(\vec{z}^w, \vec{a}^p|\vec{x},\vec{a})}[\cdot]$ can be rewritten as

$$\mathbb{E}_{q(\vec{z}|\vec{x},\vec{a})}[\log p(\vec{x}|\vec{z}) + \log p(\vec{m}|\vec{z}) + \log p(\vec{z}|\vec{a}) -$$
$$\log q(\vec{z}|\vec{x},\vec{a})] \approx \mathbb{E}_{q(\vec{z}|\vec{x},\vec{a})}[\sum_{t=1}^{\tau+1} \log p(x_t|z_t) +$$
$$\sum_{t=1}^{\tau+1} \log p(m_t|z_t) - D_{KL}(q(z_1|x_1)||p(z_1) -$$
$$\sum_{t=1}^{\tau+1} D_{KL}[q(z_{t+1}|z_t, x_{t+1}, a_t)||p(z_{t+1}|z_t, a_t)]] \quad (229)$$

In the second part of $ELBO$, according the definition of variational distribution $q(\cdot)$, $\mathbb{E}_{q(\vec{z}^w, \vec{a}^p|\vec{x},\vec{a})}[\sum_{t=\tau+1}^{\tau+H}[r(z_t, a_t) - \log \pi(a_t|z_t) + \log p(a_t)]]$ can be rewritten as

$$\mathbb{E}_{q(\vec{z}^w, \vec{a}^p|\vec{x},\vec{a})}[\sum_{t=\tau+1}^{\tau+H}[r(z_t, a_t) - \log \pi(a_t|z_t) +$$
$$\log p(a_t)]] = \mathbb{E}_{\substack{z_{\tau+1} \sim p(z_{\tau+1}|\vec{x},\vec{a}) \\ z_{t+1} \sim p(z_{t+1}|\vec{x},\vec{a}) \\ a_t \sim \pi(a_t|z_t)}}[\sum_{t=\tau+1}^{\tau+H}[r(z_t, a_t) -$$
$$\log \pi(a_t|z_t)]] \quad (230)$$

where $\log p(a_t)$ can be ignored. Obviously, the second part of $ELBO$ is the standard objective of maximum entropy RL like SAC. Therefore, the policy of the second part of $ELBO$ is maximized by

$$\pi_{new} = \arg\min_{\pi'} D_{KL}(\pi'(\cdot|z_t) || \frac{\exp Q^{\pi_{old}}(z_t, \cdot)}{Z^{\pi_{old}}(z_t)}) \quad (231)$$

where $Z^{\pi_{old}}(z_t)$ is the partition function for distribution normalization. It can be ignored because it does not contribute to the gradient of new policy. The $Q^{\pi_{old}}(z_t, \cdot)$ guides the policy update to ensure an improved new policy. New policy $\pi_{new}$ is constrained to a parameterized family of distribution $\pi'(\cdot|z_t) \in \Pi$ like Gaussians to ensure the tractable and optimal new policy.

*The interpretability via the time-step importance.* The work [10] makes the policy explainable via the model $y = g(f(x))$ where $f(\cdot)$ is modeled as GP to extract features which capture the correlations between time steps and those across different episodes. The $g(\cdot)$ is an explainable Bayesian prediction model which infers the *distribution of final rewards y and deliver time step importance*. This is achieved by the explanation model

$$f|X \sim \mathcal{N}(0, k = \alpha_t^2 k_{\gamma t} + \alpha_e^2 k_{\gamma e}),$$

$$y_i|\mathbf{F}^{(i)} \sim \begin{cases} \text{Cal}\left(\text{softmax}(\mathbf{F}^{(i)}\mathbf{W}^T)\right), \text{if conducting classification} \\ \mathcal{N}(\mathbf{F}^{(i)}\mathbf{w}^T, \sigma^2), \text{otherwise} \end{cases}$$
(232)

where $\alpha_t^2$ and $\alpha_e^2$ are component weights. The $k_{\gamma t}(h_t^{(i)}, h_k^{(j)})$ and $k_{\gamma e}(e^{(i)}, e^{(j)})$ are the correlations between time-steps (the $t$-th steps in episode $i$ and the $k$-th steps in episode $j$) and that across the episodes (episodes $i$ and $j$) respectively. The latent representation of one episode $\{h_t^{(i)}\}_{t=1:T}$ is acquired by feeding the state-action pairs $x_t^{(i)} = [s_t^{(i)}, a_t^{(i)}]$ to an RNN $\phi$. The episodic embedding $e^{(i)}$ is acquired by feeding the episodic representation $h_T^{(i)}$ to the MLP $\phi_1$. The matrix **F** is from the conversion of the flattened response $f$ where $\mathbf{F}^{(i)}$ is the $i$-th episode's encoding given by the GP mode. The $y_i|\mathbf{F}^{(i)}$ follows the conditional likelihood distribution $\mathcal{N}(\mathbf{F}^{(i)}\mathbf{w}^T, \sigma^2)$ or a categorical distribution $\frac{\exp((\mathbf{F}^{(i)}\mathbf{W}^T)_k)}{\sum_k \exp((\mathbf{F}^{(i)}\mathbf{W}^T)_k)}$ because given the GP regression model, the reward is captured by GP and defined by $y_i = \mathbf{F}^{(i)}\mathbf{W}^T + \epsilon_1, \epsilon_1 \sim \mathcal{N}(0, \sigma^2)$. The $\mathbf{W}^T$ or $\mathbf{w}^T$ is the mixing weight which is the constant and represents the time step importance.

The direct inference of the above model is computationally expensive. This can be solved by the *inducing points method* which simplifies the posterior computation by reducing the effective number of samples in $X$ from $NT$ to $M$, and also by the variational inference which simplifies the maximizing of the log marginal likelihood $\log p(y|X, Z, \Theta_n, \Theta_k, \Theta_p)$ to the maximizing of the evidence lower bound (ELBO) $\mathbb{E}_{q(f)}[\log p(y|f)] - \text{KL}[q(\mathbf{u})||p(\mathbf{u})]$ obtained by the Jensen's inequality via

$$\log p(y|X, Z, \Theta_n, \Theta_k, \Theta_p) \geq \mathbb{E}_{q(f)}[\log p(y|f)] - \text{KL}[q(\mathbf{u})||p(\mathbf{u})] \quad (233)$$

where $Z$ and $\mathbf{u}$ are the inducing points and the GP outputs of $Z$ respectively. The neural encoder parameters $\Theta_n = \{\phi, \phi_1\}$. The GP parameters $\Theta_k = \{\alpha_t, \alpha_t, k_{\gamma t}, k_{\gamma e}\}$. The prediction parameter $\Theta_p = \{\mathbf{w}/\mathbf{W}, \sigma^2\}$. The variational posterior over the inducing variable $q(\mathbf{u}) \sim \mathcal{N}(\mu, \Sigma)$. The marginal variational posterior distribution $q(f) \sim \mathcal{N}(\mu_f, \Sigma_f)$. The second part of ELBO is easy to compute via minimizing the KL divergence of $q(\mathbf{u})$ and $p(\mathbf{u})$. The first part of ELBO is computed by the reparameterization trick [89] and approximation with the Monte-Carlo Method.

# 10. Bayesian safe robotic motion planning

The safety of robotic motion planning can be better secured by considering additional factors in the Bayesian context. Here we review and conclude that safety is expected to be further improved via five perspectives: 1) Mix of GP, Lyapunov and Barrier functions. The Bayesian GP contributes to the learning of environment dynamics with uncertainty quantification, where the Lyapunov and Barrier functions further improve the stability and safety of motion planning system respectively. 2) Safe sets. Safety can be improved by building safe sets which include the *safe weight sets*, *safe data sets*, and *safe (trustable) parameter sets*. 3) Risk-averse or risk-aware objective. Designing the risk-averse or risk-aware objectives contributes to the learning of safe policies of RL which make the robot get rid of the risks or the dangers. 4) Better uncertainty quantification. The Bayesian GP takes the first step to the uncertainty quantification of the estimations via the covariance of posterior moments. The uncertainty is expected to be further quantified from many perspectives like the bounded posterior and the minimization of epistemic uncertainty and aleatoric uncertainty. 5) Robustness. The safety of motion planning is heavily correlated with the robustness against the approximation error, time-varying disturbances, and mismatch between the simulation and reality.

## 10.1 Mix of Gaussian process, Lyapunov and Barrier functions

The work [90] uses Bayesian Gaussian process (GP) models with polynomial kernel functions which is used to approximate the region of attractions (ROA) to learn the unmodeled dynamics of a nonlinear system. Safety is ensured if the system state stays within ROA of a stabilizing control policy when collecting data where the Lyapunov function is used to maintain the stability of nonlinear system with the control policy. An initial control policy is constructed by a GP model which is updated with data. GP model enlarges the ROA and increases the range of safe exploitation.

The safety in [91] is guaranteed via the Bayesian GP together with Lyapunov function and Barrier function. Bayesian GP provides an avenue for maintaining appropriate degrees of caution in the face of the unknown. Lyapunov function guarantees stability during adaptation and convergence of tracking errors, while the safety is guaranteed via the Barrier function.

## 10.2 Safe sets

*Safe weight set.* In Bayesian neural network (BNN), weights are random variables and their values are sampled via the same distribution. The work [92] seeks training the safe weight sets for which every action or trajectory is safe as long as the BNN samples its weights from the safe weight sets. Safety certificate is searched in the form of a safe positive invariant which proves the safety of the safe weight sets. Finally, the BNN policy is re-calibrated by rejecting the unsafe sampled weight to guarantee safety.

The work [93] acquires the safe reward function in imitation learning tasks via $R(\xi) = \sum_{s_i \in \xi} \widehat{w}^T \phi(s_i)$ where $\widehat{w}$ is the estimated reward weight and $\phi$ the feature of state. So, the objective is to obtain better posterior of the weight $P(\widehat{w}|\mathcal{D})$ where the safe weight set is required to compute the likelihood and prior. The $P(\widehat{w}|\mathcal{D})$ is computed via Bayesian inference where the prior is captured by Dirichlet distribution. The safe weight set is built via quantifying the uncertainty by normalized Shannon entropy $H(w_i)$ via

$$H(w_i) = -\frac{\sum_{k=1}^{|\mathcal{R}|} p_{w_i}(w_k) \log_2 p_{w_i}(w_k)}{\log_2 |\mathcal{R}|} \quad (234)$$

where the $w_i$ is the reward weight. The $\mathcal{R}$ is the vectors of reward function. $H(w_i) \in [0,1]$. $H(w_i) = 0$ indicates the certainty, while $H(w_i) = 1$ indicates the maximum uncertainty. The value of $H(w_i)$ decides whether choose of the weight or not. If $H(w_i) \geq 1 - \epsilon$ where $\epsilon$ is the level of acceptable risk, the lowest reward weight is chosen. If $H(w_i) \leq 1 - \epsilon$, expected value $\mathbb{E}[w_i]$ of the marginal distribution is chosen as the reward weight. The marginal distribution $p_{w_i}$ is computed by holding the weight being marginalized constant and summing over all possible values of the other $n - 1$ variates via $p_{w_i}(k) = \sum_{\forall \widehat{w}_j \in \widehat{w} | \widehat{w}_j \neq \widehat{w}_i} P(w_1, \dots, w_i = k, \dots, w_n)$.

*Safe data set.* The contextual Bayesian optimization [94], the safe Bayesian optimization (SafeOpt), and the SafeOpt with multiple constraint (SafeOpt-MC) [95] are the extension of BO. Contextual BO enables optimization of functions that depend on additional, external variables, which are called contexts. The context $z \in \mathcal{Z}$ is included in the kernel by

$$k((a, i, z), (a', i', z')) = k_a((a, i), (a', i')) \cdot k_z(z, z') \quad (235)$$

Safety constraint is a special case of the context. SafeOpt aims to find global maximum within the currently known safe set (exploitation), and expand the safe set (exploration). SafeOpt better balances the trade-off between the exploitation and exploitation. SafeOpt-MC further extends the SafeOpt by incorporating multiple constraints separated from the objective to compute the location of the next sample via

$$a_n \leftarrow \arg\max_{a \in G_n \cup M_n} \max_{i \in \mathcal{I}} w_n(a, i) \quad (236)$$

where $w$ is defined by $w_n(a, i) = u_n^i(a) - l_n^i(a)$. The upper bound of the contained set $C_n(a, i)$ at iteration $n$ is defined by $u_n^i(a) \coloneqq \max C_n(a, i)$. The lower bound of $C_n(a, i)$ is defined by $l_n^i(a) \coloneqq \min C_n(a, i)$ where the contained set is defined by $C_n(a, i) = C_{n-1}(a, i) \cap Q_n(a, i)$. The $Q_n(a, i)$ is the GP's confidence intervals for the surrogate function and $Q_n(a, i) \coloneqq [\mu_{n-1}(a, i) \pm \beta_n^{\frac{1}{2}} \sigma_{n-1}(a, i)]$ where $\beta$ is a scalar that determines the desired confidence level. The $G_n$ is an optimistic set of parameters that could potentially enlarge the safe set $S_n$ which is defined using *Lipschitz continuity property*, while $M_n$ is the subset of $S_n$ that could either improve the estimate of the maximum or expand the safe set. The convergence to the global optimum can be further improved by exploring outside the initial safe area to create new safe area while still guaranteeing safety with high probability [96].

*Safe (trustable) parameter set.* The work [97] describes the safety to be the trust which follows a Beta distribution $t_i \sim Beta(\alpha_i, \beta_i)$. The parameter set is defined by $\theta = \{\alpha_0, \beta_0, w^s, w^f\}$ where $w^s$ and $w^f$ are the gains due to the human agent's positive and negative experiences in each subtask. The optimal parameter $\theta^*$ is estimated by the maximum a posteriori estimation (MAP). The posterior is defined as $p(\theta|P_m, T_m, r)$ where the $P_m$, $T_m$, and $r$ are the robotic agent's performance, trust history and robot reliability respectively. The unknown prior $p(\theta)$ is estimated via the maximum likelihood.

### 10.3 Risk-averse or risk-aware objective

The work [98] ensures the safety via defining a risk-averse or risk-aware objective. First, the risk level is defined as $\tau \in \Omega \stackrel{\text{def}}{=} \{2,1,0\}$. Then, the safety metrics based distribution is defined as $P(d|\tau)$ where $d$ is the relative position of cars. Therefore, the posterior distribution of the risk given the relative position is computed via Bayesian inference by $P(\tau|d) = \frac{P(d|\tau)P(\tau)}{\sum_{\tau \in \Omega}(P(\tau)P(d|\tau))}$. Instead of acquiring the optimal policy by maximizing the accumulated reward in standard RL like DQN, a risk-aware objective is defined as $\pi^* = \arg\min_\pi \mathbb{E}_\pi \{\sum_{i=0}^\infty \gamma^i \varepsilon_{t+i} | s_t\}$ where $\gamma$ is a discounted factor, and $\varepsilon$ is the risk coefficient defined by $\varepsilon = \mathbb{E}(\tau) = \sum_{\tau \in \Omega} \tau P(\tau|d)$.

The work [99] seeks the high-reward and safe policy by solving the constrained MDP (CMDP) objective

$$\max_{\pi \in \Pi} \max_{p_\theta \in \mathcal{P}} J(\pi, p_\theta), \text{s.t.} \max_{p_{\theta^i} \in \mathcal{P}} J^i(\pi, p_{\theta^i}) \leq d^i, \forall i \in \{1, \dots, C\} \quad (237)$$

where $J(\pi, p_\theta)$ is the reward objective defined by $J(\pi, p_\theta) = \mathbb{E}[\sum_{t=0}^T r_t | s_0]$. Actions are selected by $a_t \sim \pi(\cdot | s_t)$. Next state $s_{t+1}$, reward $r_t$, and cost of unsafe behavior $c_t^i$ are inferred by the unknown transition distribution $p_\theta$ via $s_{t+1}, r_t, c_t^i \sim p_\theta(\cdot | s_t, a_t)$ where its posterior (Bayesian predictive distribution $p(\theta|\mathcal{D})$) is approximated by the *stochastic weight averaging-Gaussian* (SWAG). The $J^i(\pi, p_{\theta^i})$ is the constraint of the unsafe behavior defined by $J^i(\pi, p_{\theta^i}) = \mathbb{E}[\sum_{t=0}^T c_t^i | s_0] \leq d^i$. The $\mathcal{P}$ is plausible transition distributions acquired via the *optimism in the face of uncertainty* and *upper confidence reinforcement learning* (UCRL). The $d^i$ are human-defined thresholds, and $i = 1, \dots, C$ denote the distinct unsafe behaviors. Finally, the CMDP is solved via the augmented *Lagrangian with proximal relaxation method* [99].

Unlike the CMDP that attaches constraints to the objective, [100] constrains the objective to yield safe policy by attaching the penalty to the accumulative reward when approximating the value function via

$$V_{\tilde{\theta},t}^\theta \coloneqq \mathbb{E}_E \left[ \sum_{j=t}^{T-1} r(j, x_j, u_j; \theta_R) - I(\rho_j) \middle| \begin{array}{l} s_t = s \\ a_j = \bar{a}^*(j, s_j; \tilde{\theta}) \\ s_{j+1} = f(s_j, a_j; \theta_F) + \epsilon_{F,j} \\ \rho_j = \min_{\rho \geq 0} \rho \text{ s.t. } s_j \in \mathcal{S}_\delta(\rho) \end{array} \right] \quad (238)$$

where $\tilde{\theta} = (\tilde{\theta}_R, \tilde{\theta}_F)$ is the imperfect estimate of true parameter $\theta = (\theta_R, \theta_F)$. The $\epsilon_{F,j}$ is the external turbulence. The $I(\rho_j)$ is the penalty where $\rho \geq 0$ is the slack variable. The $\mathcal{S}_\delta(\rho)$ is the cautious soft-constraint formulation with parameter $\delta$ which defines the degree of cautiousness. The penalty $I(\rho_j)$ is acquired by optimizing state trajectories subject to a tightened state constraint set. This bounds the expected number of unsafe learning episodes based on the Lipschitz continuity property, like the number of episodes in which the tight state constraints are violated, and results in $\mathcal{S}_\delta(\rho)$. This method combines model-based RL using Bayesian posterior sampling with a safe policy approximated by $V_{\tilde{\theta},t}^\theta$, and yields a new class of model-based RL policies with safety guarantees.

### 10.4 Better uncertainty quantification

Safety can be improved indirectly by better uncertainty quantification. One approach to acquire better quantified uncertainty is to use *probabilistic neural networks* to leverage large amounts of data, instead of standard Bayesian GP which is the point estimate based on small amount of data. The neural weights are modeled as distributions like the Gaussian. The likelihood of target $y$ given the weights $W$, features $X$ and noise precision $\gamma$ is defined by

$$p(y|W,X,\gamma) = \prod_{n=1}^{N} \mathcal{N}(y_n|f(x_n;W),\gamma^{-1}) \quad (239)$$

The Gaussian prior can be defined as

$$p(W|\lambda_p) = \prod_{l=1}^{L} \prod_{i=1}^{V_l} \prod_{j=1}^{V_{l-1}+1} \mathcal{N}(w_{i,j,l}|0,\lambda_p^{-1}) \quad (240)$$

where $l$ is the layer, $V_l$ the number of neurons at layer $l$, and $\lambda_p$ the precision parameter. Given the dataset $\mathcal{D} = (X,y)$, the posterior is acquired by applying the Bayes rule

$$p(W,\gamma,\lambda_p|\mathcal{D}) = \frac{p(y|W,X,\gamma)p(W|\lambda_p)p(\gamma)p(\lambda_p)}{p(y|X)} \quad (241)$$

The prediction of output is acquired via

$$p(y_*|x_*,\mathcal{D}) = \int p(y_*|x_*,W,\gamma)p(W,\gamma,\lambda_p|\mathcal{D})d\gamma d\lambda_p dW \quad (242)$$

This intractable integral can be approximated by the *probabilistic backpropagation* (PBP) which is more efficient than the MC dropout ensemble (MDE) [101]. In the PBP, the weight $w_{i,j,l}$ is defined by one-dimensional Gaussian. In the forward propagation process, the marginal log-likelihood $\log Z$ is acquired. In the backpropagation process, $\log Z$ is used to update the mean and variance of the posterior via

$$\begin{cases} m^{new} = m + v \frac{\partial \log Z}{\partial m} \\ v^{new} = v - v^2 \left[ \frac{\partial \log Z^2}{\partial m} - 2 \frac{\partial \log Z}{\partial v} \right] \end{cases} \quad (243)$$

where $m^{new}$ and $v^{new}$ are the new mean and variance. Finally, the new Gaussian belief $q^{new}(w) = \mathcal{N}(w|m^{new},v^{new})$ is acquired. In recurrent neural networks case like the LSTM [101], each layer consists of two weights. Therefore

$$h_t = h_t = f(W_h h_{t-1} + W_x x_t) \quad (244)$$

where $f(\cdot)$ is the activation function. The $h_t$ is the hidden state. Hence, additional new Gaussian beliefs should be computed for the weight of hidden state $W_h$. As a result, the computing resources are reduced and safety is improved simultaneously.

Better uncertainty quantification can be obtained via the bounded posterior of Bayesian estimator. [102] takes two Bayesian estimators to ensure the imprecise but safe motion estimation: 1) Upper bounded posterior of conservative Bayesian estimator based on partial prior knowledge before the tasks. 2) Upper and lower bounded posterior of Bayesian estimator with Dirichlet prior during the task.

The work [103] guarantees the safety by minimizing the epistemic uncertainty due to the prior distribution over MDP and aleatoric uncertainty due to the inherent stochasticity of MDP. This is achieved by maximizing the conditional value at risk (CVaR) objective via

$$\max_{\pi \in \Pi^{M^+}} CVaR_\alpha(G_\pi^{M^+}) = \max_{\pi \in \mathcal{G}^+} \min_{\sigma \in \Sigma^{\mathcal{G}^+}} \mathbb{E}[G_{(\pi,\sigma)}^{\mathcal{G}^+}] \quad (245)$$

where $G_\pi^{M^+}$ and $G_{(\pi,\sigma)}^{\mathcal{G}^+}$ are the distribution over total returns, and they are defined by $G_\pi^M = \sum_{t=0}^{H} R(s_t, a_t)$. The $M^+$ denotes a Bayes-Adaptive MDP (BAMDP). $\mathcal{G}^+$ is the Bayes-Adaptive CVaR Stochastic Game where "+" denotes the augmented parameter in BAMDP. $\Pi^{\mathcal{G}^+}$ denotes the set of Markovian agent policies mapping agent states to agent actions. $\Sigma^{\mathcal{G}^+}$ denotes the set of Markovian adversary policies. The computing of CVaR is intractable. Therefore, the MCTS approximation approach is used to handle the large state space by focusing search in promising areas, while the progressive widening [103] paired with Bayesian optimization is used to handle the continuous action space for the adversary.

### 10.5 Robustness

Safety is also interpreted as the robustness against the approximation error, time-varying disturbances, and mismatch between the simulation and reality. [50] builds on the BAMDP and modifies the online Bayesian estimation to be robust against approximation errors of the parametric model to a real plant. [104] incorporates GP to the $\mathcal{L}_1$ adaptive controller in the system defined by

$$\dot{x}(t) = A_m x(t) + B_m(u(t) + f(x(t))), y(t) = C_m x(t) \quad (246)$$

where $x(t)$, $y(t)$ and $u(t)$ are the system state, regulated output, and control input respectively. The $A_m$, $B_m$ and $C_m$ are known matrixes. The $f$ is an unknown uncertainty modeled as a GP. This results in the safe and efficient learning of uncertainty dynamics due to the GP, and the robustness and tracking performance under the time-varying disturbances due to the $\mathcal{L}_1$ adaptive controller. [105] takes the Bayesian optimization with Gaussian process to random the simulator during training according to a distribution over domain parameters. This results in a policy which maximizes the real-word objective and shortens the mismatch between the simulation and reality.

## 11. Conclusion and future directions

### 11.1 Analytical summary and conclusion

This section summarizes different methods reviewed to solve robotic motion planning problems under Bayesian context for better data-efficiency, interpretability, and safety. First, an overview is shown via the knowledge graph (Figure 5), and the latest progress of Bayesian inference for robotic motion planning is also given. Second, some open questions for incorporating Bayesian inference into robotic motion planning are discussed, and future robotic motion planning under Bayesian context is also analyzed.

*Overview of Bayesian inference for data-efficient, explainable, and safe robotic motion planning.* Bayesian inference is a statistical or probabilistic approach where the Bayes' theorem is applied to update the probability (density) of the hypothesis like the future robotic states or actions. The update of the probability relies on learning from collected information or evidence. The probability theories ground the Bayesian inference. Currently, data-efficiency, interpretability, and safety are becoming critical in robotic motion planning for practical applications. Bayesian inference is a promising direction to achieve this goal. The objective of robotic motion planning is to acquire a probabilistic model or policy by which the future states or actions can be generated. Bayesian estimation is a classical but still a hot research area to realize this via maintaining and updating the environmental dynamics or transitions, as more information or evidence is becoming available. The convergence of environmental transition in Bayesian estimation can be further improved by enjoying the advantage of RL in convergence, due to the architecture of RL which secures an efficient convergence of the algorithm. This is achieved by incorporating Bayesian inference to RL, resulting in the Bayesian RL which includes the model-based Bayesian RL, model-free Bayesian RL, and hybridization of Bayesian inference and RL. The Bayesian RL is competent to solve the motion planning tasks in complex scenarios cases and inverse RL tasks.

The direct consequence of applying Bayesian inference to robotic motion planning tasks is the improvement of data efficiency. This means that the environmental transition or policy can be converged with less data in training. Bayesian inference also improves the interpretability and safety of robotic motion planning by providing uncertainty quantification to the prediction or hypothesis via like the covariance of the posterior. The interpretability can be further improved by deriving more post-training explanations of the policy, such as the time-step importance and the sematic mask. The safety of robotic motion planning can be further secured by considering more additional safety factors. These factors includes the mix of Gaussian process, Lyapunov and Barrier functions, finding safe sets (weight set, data set and parameter set), designing risk-averse or risk-aware objectives, better uncertainty quantification of the policy, and the robustness against the approximation error, time-varying disturbances, and mismatch between the simulation and reality.

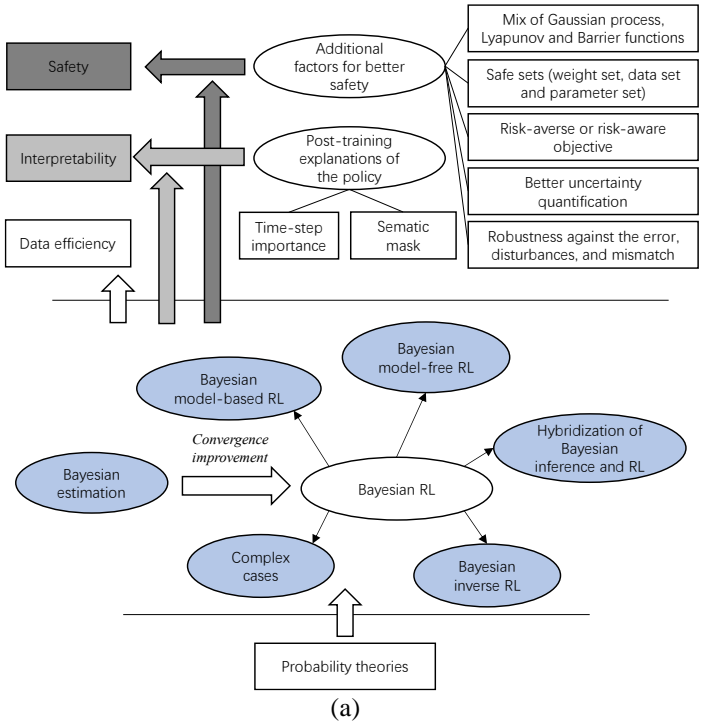

(a)

Figure 5. An overview of knowledge graph of Bayesian inference for data-efficient, explainable, and safe robotic motion planning.

*Bayesian estimation.* The Bayesian learning grounds the Bayesian posterior estimation where the state transition (parameter) is captured by surrogate models like Gaussian. Bayesian estimation is easy to apply to linear Gaussian case. If the collected data is processed in a batch, the batch Bayesian inference is applicable to compute the posterior moments, while the recursive Bayesian inference like Kalman filter is appropriate if the data is processed recursively. The posterior computing of Bayesian estimation is complex where the state transformation model is non-linear and the parameter is captured by complex distributions like the Gaussian mix distribution and non-Gaussian distribution. In batch case, the batch Bayesian inference is still applicable to compute the posterior, but additional process is required to simplify the posterior computing, such as the linearization. In recursive case, the recursive Bayesian inference known as the Bayes filter is possible to compute the posterior theoretically, but the posterior computing is intractable or computationally expensive, due to the infinite-dimension space of probability density function which requires infinite memory to represent the posterior belief and the infinite computing resources required to compute the integral.

The posterior approximation methods solve these problems well. The linearization is applied to the Bayes filter, resulting in the EKF and IEKF. The approximation via the sampling methods like Monte-Carlo sampling results in the Sigmapoint Kalman filter (SPKF) and iterated SPKF (ISPKF). The approximation via *sampling the importance sampling method* based on Monte-Carlo sampling results in the particle filter (PF) and its variants. Among the posterior approximation methods, the linearization brings errors and biases, therefore this method is inaccurate. The sampling methods are accurate, but the accuracy is heavily dependent on the number of samples. However, more samples require much computation, therefore the sampling methods face the dilemma between the accuracy and sample numbers. The variational inference which is seen as the extension of the EM is an alternative of MCMC and it is accurate, but the approximation process of variational inference is unknown. To be specific, the sampling methods like the MCMC approximate the posterior models via sampling with theoretical guarantee to the approximation accuracy, while the variational inference via the optimization without guarantee. The variational inference is *faster* than MCMC to handle the large dataset and complex posterior models like the Gaussian mixture model and conditionally conjugate model, but variational inference underestimates the posterior covariance.

The above methods of Bayesian estimation are available given the data is well-collected, therefore the state transition is updated by learning from these data. They do not consider how to collect the data and the optimum of the posterior, therefore the optimal transitions or dynamics are acquired with as little data as possible. The Bayesian optimization like the GP-UCB considers this problem by repeatedly evaluating the system at locations $a$. This improves the mean estimate of the underlying function and decreases the uncertainty at candidate locations for the maximum, such that the global maximum is provably found eventually. Sometimes, it is unnecessary to directly compute the posterior to acquire the state transitions or policies for generating future states or actions. The MAP, maximum likelihood, and SWF are alternatives of Bayesian estimation to acquire the state transitions or policies under Bayesian contest. The knowledge graph of Bayesian estimation is shown in Figure 6.

Bayesian estimation is still a hot topic nowadays. Some works directly apply the Bayesian theorem [106] [107] [108], Bayesian learning with Gaussian distribution [109], and Bayesian optimization [110] to simple motion prediction tasks. Recursive Bayesian inference (recursive Bayesian filter) better suits the complex non-linear non-Gaussian cases, and widely used in autonomous driving tasks for robotic motion planning or trajectory prediction [9] [111] [112] [113] [114]. Recursive Bayesian filter [115] [116] [117] [118] and particle filter [119] are also used to predict the intention of humans to facilitate the motion planning of the robot in the human-robot interaction tasks. The approximation methods play critical rules to approximate the posterior in these tasks, especially the sampling methods [120], EM [117] and variational inference [121]. The neural network also combines with Bayesian estimation to handle the complex tasks, resulting in Bayesian neural network (BNN) where model uncertainty is represented by a distribution over the weights of the neural network [120].

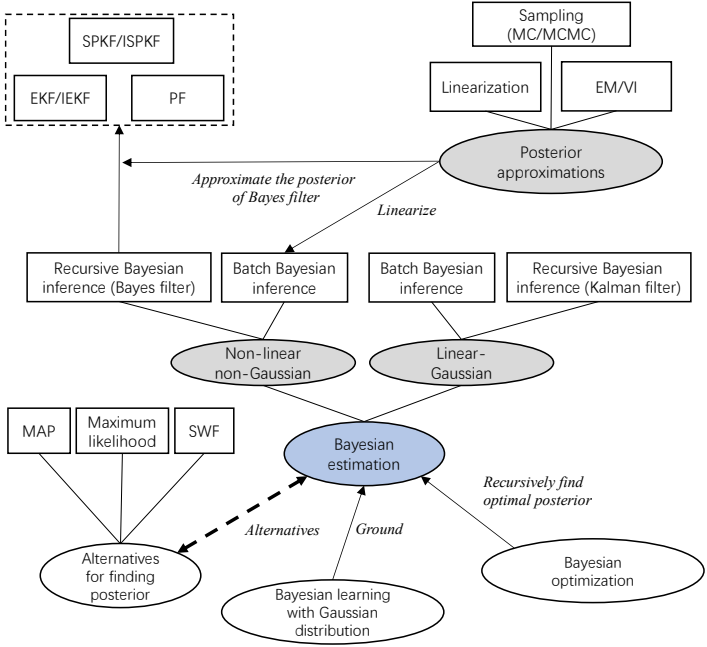

Figure 6. The knowledge graph of Bayesian estimation.

*model-based Bayesian RL.* The Bayesian estimation methods have difficulties in computation and convergence where the parameter space is high-dimensional. These problems can be alleviated by incorporating Bayesian estimation methods to the RL framework, resulting in model-based Bayesian RL and model-free Bayesian RL. The difference of model-based Bayesian RL and model-free Bayesian RL is that model-based Bayesian RL acquires the policy by training and maintaining the state transitions or environmental dynamics, while the state transitions are unknown in model-free Bayesian RL. This makes model-based Bayesian RL more explainable than model-free Bayesian RL. Once the solution space (e.g., policy space) exhibits less regularity than the underlying dynamics, model-based Bayesian RL is better than model-free Bayesian RL for motion planning tasks.

The RL problem can be basically described as a MDP or PO-MDP where the solutions are always intractable. Model-based Bayesian RL describes the problem as the BA-MDP. However, the solution (the computing of value function via the environmental dynamics) in BA-MDP is still computationally expensive. Moreover, model-based Bayesian RL has the exploration-exploitation dilemma which demotivates the convergence. The problems in the computing of value function is expected to be alleviated via the value approximation methods which include the offline value approximation, online near-myopic value approximation, and online tree search approximation. The offline value approximation computes the value function in closed form which is computationally expensive. The online near-myopic value approximation is based on one-step prediction which results in sub-optimal convergence. The online tree search approximation like BA-MCP attracts much attention nowadays. The sampling method and search tree in the online tree search approximations reduce the posterior computation and consider the history to generate policy, instead of one-step prediction in the online near-myopic value approximation. The exploration-exploitation dilemma in BA-MDP can be alleviated via the exploration bonus methods which add exploration bonus to the computing of value function. The knowledge graph of model-based Bayesian RL is shown in Figure 7.

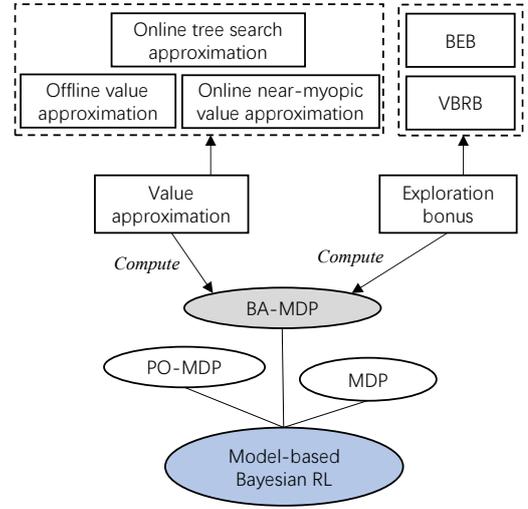

Figure 7. The knowledge graph of model-based Bayesian RL.

*Model-free Bayesian RL.* Model-based Bayesian RL assumes the parameters like the prior is captured by the surrogate models like the Gaussian and Dirichlet distributions. This simplifies the computation in model-based Bayesian RL, but it brings error and bias because the surrogate models cannot fully represent the true distribution of parameters especially in non-linear non-Gaussian cases. Hence, model-free Bayesian RL is more efficient in convergence than model-based Bayesian RL in cases where the solution space (e.g., policy space) exhibits more regularity than the underlying dynamics. The model-free Bayesian RL is better suitable for the large-scale real-world problems than the model-based Bayesian RL.

The model-free RL is basically categorized to optimal value RL like DQN, policy gradient RL like policy gradient algorithm, and actor-critic RL like actor-critic algorithm. In Bayesian context, these three types of RL correspond to the value function Bayesian RL, Bayesian policy gradient, and Bayesian actor-critic. The GPTD is a representative of the value function Bayesian RL where the value function is unknown and it is modeled as the Bayesian GP. Therefore, the posterior of value function (policy) is acquired as that in Bayesian learning. The integral (the gradient to update the policy) derived from classical policy gradient is intractable and Monte-Carlo sampling is used to compute the integral. This method is inaccurate and brings high variance. Bayesian policy gradient alleviates this problem by Bayesian quadrature which computes the integral by modeling its integrand as the Bayesian GP. Bayesian actor-critic further improves the convergence of Bayesian policy gradient by using step-based gradient, instead of trajectory-based gradient to update the policy. In Bayesian actor-critic, Bayesian quadrature is used to solve the integral. The integral is assumed to be linear. The unknown integrand (value function) in the integral is modeled as the Bayesian GP, and it is computed using the same steps as that in GPTD. The knowledge graph of model-free Bayesian RL is shown in Figure 8.

The value function Bayesian RL and Bayesian actor-critic are active in robotic motion planning tasks. For instance, the work [122] employs temporal difference learning in a Bayesian framework to learn vehicle control signals from sensor data.

The work [123] performs Bayesian analysis on the value function for robotic continuous control. The work [124] proposes actor-critic with teacher ensembles (AC-Teach) that guides the learning of a Bayesian DDPG agent via an ensemble of suboptimal teachers.

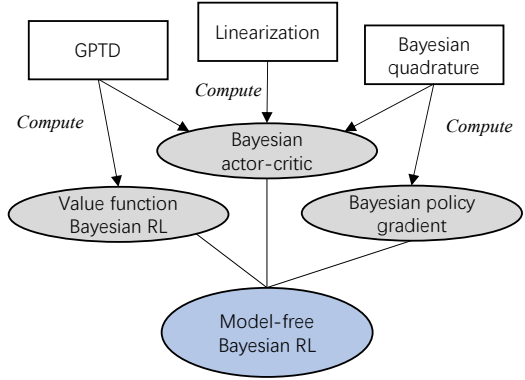

Figure 8. The knowledge graph of model-free Bayesian RL.

*Bayesian RL in complex cases.* Model-free Bayesian RL is more efficient in convergence than model-based Bayesian RL in cases where the solution space (e.g., policy space) exhibits more regularity than the underlying dynamics. In practice, model-free Bayesian RL is more efficient than model-based Bayesian RL in complex large-scale tasks. Hence, Bayesian RL in complex cases mainly focuses on model-based Bayesian RL in cases where there are unknown reward functions, partial observable states, multiple agents, and multiple tasks in the environment.

In unknown reward cases, the reward function can be modeled as the Gaussian GP to compute its posterior. When sampling the posterior of state transitions to compute the policy like the value function, the posterior of the reward function should be sampled simultaneously.

In partial observable case (BA-PO-MDPs), the computing of value function requires estimating the Bellman equation over all possible hyper-states for every belief. This means all models should be sampled from the posterior. Then, models are solved and action is sampled from solved models. This estimation method is intractable. Bayes risk is expected to solve this problem, but it provides a myopic view of uncertainty. How to sample the posterior in BA-PO-MDPs in an efficient way is a problem remaining to be solved.

In multi-agent case, multi-agent RL faces four challenges: Partial observable state, unknown system dynamics, exponential joint policy space, and the coordination of the agent's policies. The solutions TD-PO-MDP, POMCP, FV-PO-MCP, TD-FV-MCP, and BA-TD-POMDP are expected to handle these challenges. The MCTS is applied to the PO-MDP for online planning, resulting in partially observable Monte-Carlo planning (POMCP). The joint policy space can be reduced by the factored-value POMCP (FV-PO-MCP) which decomposes the value function and global look-ahead tree into overlapping factors and multiple local look-ahead trees respectively. Moreover, the above tricks can be combined together, resulting in the transition-decoupled factored value-based Monte-Carlo online planning (TD-FV-MCP) to further improve the performance of multi-agent motion planning. However, the efficiency of searching and expanding the local look-ahead tree for each agent is poor, and there is a lack of efficient policy coordination or communication among agents, resulting in suboptimal convergence in training. The BA-TD-PO-MDP attempts to solve these problem by applying the decentralized greedy search, but this still results in suboptimal convergence. The computation of tree search and the efficiency of policy coordination among agents are problems that remain to be solved. The centralized training with decentralized execution (CTDE) where all agents can access (partial observable) data from all other agents [125], parameter or policy sharing [126][127], experience sharing where the agents share experience and maintain separate policy and value networks [128] are possible solutions to improve the convergence via reducing the computation and improving the coordination efficiency of agents.

In multi-task RL (MRL), MRL should learn multiple tasks in parallel with the share representation, and what is learnt in each tasks contributes to the learning of other tasks. This means the MRL is a kind of transfer learning where the *instance*, *representation*, or *parameter can be transferred to other tasks*. MRL can learn the knowledge explored by itself, while it can also learn from the expert knowledges. They are two types of MRL problems, and how to identify different tasks from knowledge in these two problems matters in the realization of MRL. The knowledge graph of Bayesian RL in complex cases is shown in Figure 9.

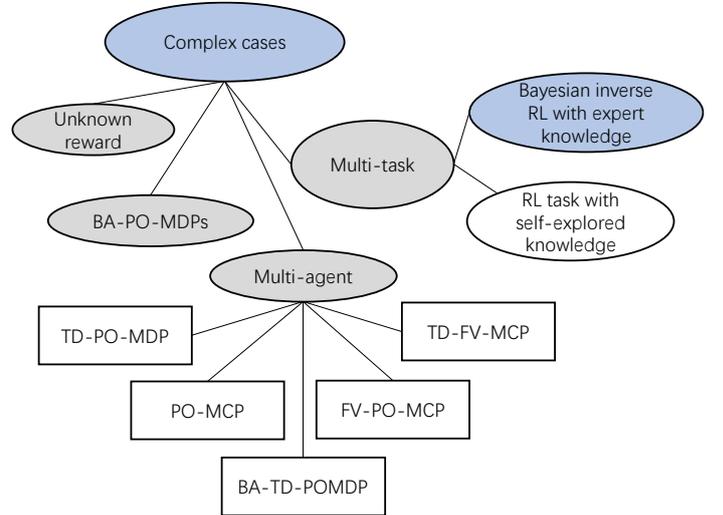

Figure 9. The knowledge graph of Bayesian RL in complex cases.

*Hybridization of Bayesian inference and RL.* Instead of incorporating Bayesian inference to RL directly to acquire model-based Bayesian RL and model-free Bayesian RL, Bayesian inference is possible to work as a function module of RL to improve the convergence of RL, resulting in hybridization of Bayesian inference and RL. The Bayesian optimization, variational inference, and Bayesian-based objective are popular ways to contribute to the convergence of RL.

Bayesian optimization can help in the training parameter/hyper-parameter tunning by modeling the training curves as the Bayesian GP. Bayesian optimization can also reduce the search space of parameters (policy) via constraining the policy search space of RL to a sublevel-set of the Bayesian surrogate model's predictive uncertainty, or projecting the high-dimensional space or low-dimensional hyperplane.

The evidence lower bound can be derived from variational inference to constrain or guide the convergence of model-free RL like the SAC. Variational inference can approximate the posterior of policy in model-based RL to acquire an accurate policy, therefore improving the convergence. Variational inference is also used to quantify the uncertainty of value function of RL to detect the low-quality data which is penalized in the training. This improves the data quality in training, therefore improving convergence.

The convergence of RL can be improved by designing the Bayesian-based objectives where the Bayesian curiosity/surprise and PAC loss are used as the objectives to guide the convergence process of RL. Other hybridizations of Bayesian inference and RL also help. They are the mix of dynamics and explored data, and the mix of control prior and RL. The knowledge graph of the hybridization of Bayesian inference and RL is shown in Figure 10.

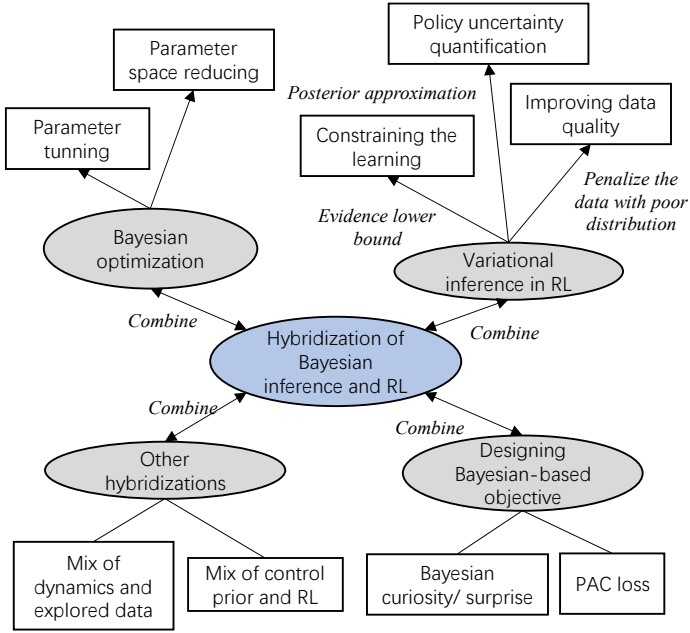

Figure 10. The knowledge graph of the hybridization of Bayesian inference and RL.

To conclude, we investigated possible methods related to Bayesian inference for robotic motion planning in this review. These methods include the classical Bayesian estimation, model-based Bayesian RL, model-free Bayesian RL, Bayesian inverse RL, Bayesian RL in complex cases, and hybridization of Bayesian inference and RL. Bayesian estimation is competent to solve the simple robotic motion planning tasks where the parameter space is low-dimensional. Model-based Bayesian RL better suits the small-scale complex robotic motion planning by training and maintaining explainable state transitions, model-free Bayesian is well-performed in large-scale complex motion planning but the state transitions are unknown. Bayesian inference can also work as a part of RL to improve the convergence of RL in motion planning tasks. The direct consequence of incorporating Bayesian inference to RL is the improvement of data-efficiency. The Bayesian inference methods, either Bayesian estimation or Bayesian RL, provide the uncertainty quantification to the posterior of the policy. This improves the interpretability and safety of robotic motion planning to some degree. The interpretability of robotic motion planning can be further improved by deriving more post-training explanations like the time-step importance and semantic masks to justify the policy. The safety of robotic motion planning is also expected to be further improved by considering more safe-related factors like the safe sets, risk-aware objectives, and robustness. Finally, data-efficient, explainable, and safe robotic motion planning policies are expected to be acquired to contribute to the real-world applications under the Bayesian context.

## 11.2 Open questions and future directions

The improvement of data-efficient, explainable, and safe robotic motion planning under Bayesian context relies on the improvement of Bayesian inference from every aspects, including the improvement of Bayesian inference itself, improvement of RL, and how to combine the Bayesian inference and RL. Bayesian inference assumes the unknown function (state transitions or policy) is captured by certain distributions like the Gaussian. This assumption provides the uncertainty quantification or interpretability to the predictions, but how much the assumption reflects the true distribution of unknown function, especially in complex non-linear non-Gaussian cases.

*Question 1: How to select and justify the assumptions (prior distributions), therefore reducing the difference of our assumptions and the true distribution of unknown functions in complex cases?*

The MCMC sampling works slowly in large-scale scenarios, but it provides theoretical guarantee to its approximation process. The VI is faster in large-scale scenarios where its approximation process is realized by minimizing the KL divergence. However, this process is unknown. It is interesting to design a new approximation method in VI to replace KL divergence to make VI explainable. The combination of MCMC and VI is also a good solution to improve the interpretability of VI.

*Question 2: How to improve the interpretability of VI?*

BO better finds the optimum of unknown functions in the complex cases like the high-dimensional parameter space by recursively finding and evaluating the next sample. However, as the RL, the BO faces the exploration-exploitation dilemma. How to balance the trade-off of exploration-exploitation matters to further improve the data-efficiency.

*Question 3: How to better balance the trade-off of exploration-exploitation of BO?*

In model-based Bayesian RL where the hyper-states is high-dimensional, such as the BA-PO-MDP with unknown reward function, the computing of value function requires estimating the Bellman equation over all possible hyper-states for every belief. This means all models should be sampled from the posterior. Additionally, the posterior of reward function should also be sampled simultaneously. This results in various possible posteriors which should be evaluated one by one to acquire maximum posterior. This process is computationally expensive, and there lacks efficient posterior sampling and evaluation methods to acquire the maximum posterior as fast as possible. Hence,

*Question 4: How to design efficient posterior sampling and evaluation methods to acquire the maximum posterior as fast as possible in high-dimensional hyper-states case with unknown reward function?*

In multi-agent RL, the coordination and cooperation of agent's policies decide the overall convergence of multi-agent RL. The optimal policy of each agent does not mean the optimal performance of multi-agent RL sometimes. Hence,

*Question 5: How to coordinate and cooperate agents' policies to find a better trade-off of each agent's policy, therefore leading to optimal convergence or performance of multi-agent RL?*

In the hybridization of Bayesian inference and RL, how to use Bayesian inference to model some (intermediate) outcomes of RL like the training curves as the unknown function to regulate or guide the training of RL is a promising direction to explore.

*Question 6: How to use Bayesian inference to model some (intermediate) outcomes of RL like the training curves as the unknown function to regulate or guide the training of RL?*

Moreover, how to fast detect and classify the different tasks is helpful to improve the convergence of multi-task RL. More post-training explanations and additional safety measures remain to explore. Future data-efficient, explainable, and safe robotic motion planning may focus on these directions to make the robot practical to use in the real-world.

Sigma Point Kalman Filter with Applications to Long Range Stereo," *Proc. Robot. Sci. Syst.*, pp. 1–8, 2006.